\PassOptionsToPackage{table,dvipsnames}{xcolor}

\documentclass{applemlr}

\usepackage{amsmath}
\usepackage{enumerate}
\usepackage{algorithm}
\usepackage{algpseudocode}
\usepackage{amsfonts}
\usepackage{amsthm}
\usepackage{cleveref}
\usepackage{diagbox}
\usepackage{colortbl}
\usepackage{amssymb}
\usepackage{xspace}
\usepackage{wrapfig}
\usepackage{adjustbox}
\usepackage{tabularx}
\usepackage{booktabs}
\usepackage{mathtools}
\usepackage{tikz}
\usepackage{enumitem}
\usepackage{silence}
\usepackage{dsfont}
\usepackage[table]{xcolor}
\usepackage[dvipsnames]{xcolor}
\usepackage{multirow}
\usepackage{makecell}
\usepackage{xfakebold}

\usepackage{amsmath,amsfonts,bm}

\def\eqref#1{equation~\ref{#1}}

\def\1{\bm{1}}

\DeclareMathAlphabet{\mathsfit}{\encodingdefault}{\sfdefault}{m}{sl}
\SetMathAlphabet{\mathsfit}{bold}{\encodingdefault}{\sfdefault}{bx}{n}

\newcommand{\E}{\mathbb{E}}

\newcommand{\R}{\mathbb{R}}

\newcommand{\softmax}{\mathrm{softmax}}

\definecolor{textgray}{HTML}{6E6E73}
\usetikzlibrary{positioning, calc}
\usetikzlibrary{decorations.pathmorphing}

\makeatletter
\patchcmd{\wrong@fontshape}{\@gobbletwo}{}{}{}
\makeatother
\numberwithin{equation}{section}
\makeatletter
\AtBeginDocument{
  \urlstyle{sf}
  
}
\makeatother

\definecolor{light}{RGB}{125, 125, 125}
\crefname{tcb@cnt@pbox}{code}{code}
\Crefname{tcb@cnt@pbox}{Code}{Code}
\crefname{assumption}{assumption}{assumption}
\Crefname{assumption}{Assumption}{Assumptions}

\newtcolorbox[auto counter]{pbox}[2][]{
  colback=white,
  title=Code~\thetcbcounter: #2,
  #1,fonttitle=\sffamily,
  fontupper=\sffamily,
  arc=2pt,
  colframe=bgcolor,
  coltitle=fgcolor,
  colbacktitle=bgcolor,
  toptitle=0.25cm,
  bottomtitle=0.125cm
}

\makeatletter
\newcommand\applefootnote[1]{%
  \begingroup
  \renewcommand\thefootnote{}%
  \renewcommand\@makefntext[1]{\noindent##1}%
  \footnote{#1}%
  \addtocounter{footnote}{-1}%
  \endgroup
}
\makeatother

\definecolor{cverbbg}{gray}{0.90}

\usepackage{url}
\usepackage{listings}
\usepackage[textsize=tiny]{todonotes}

\usetikzlibrary{positioning, decorations.pathreplacing, calc}

\theoremstyle{plain}

\theoremstyle{definition}

\theoremstyle{remark}

\crefname{figure}{Figure}{Figures}
\Crefname{figure}{Figure}{Figures}
\crefname{table}{Table}{Tables}
\Crefname{table}{Table}{Tables}
\crefname{section}{Section}{Sections}
\Crefname{section}{Section}{Sections}
\crefname{equation}{Equation}{Equations}
\Crefname{equation}{Equation}{Equations}
\crefname{appendix}{Appendix}{Appendices}
\Crefname{appendix}{Appendix}{Appendices}
\crefname{theorem}{Theorem}{Theorems}
\Crefname{theorem}{Theorem}{Theorems}
\crefname{proposition}{Proposition}{Propositions}
\Crefname{proposition}{Proposition}{Propositions}
\crefname{lemma}{Lemma}{Lemmas}
\Crefname{lemma}{Lemma}{Lemmas}
\crefname{corollary}{Corollary}{Corollaries}
\Crefname{corollary}{Corollary}{Corollaries}
\crefname{definition}{Definition}{Definitions}
\Crefname{definition}{Definition}{Definitions}
\crefname{assumption}{Assumption}{Assumptions}
\Crefname{assumption}{Assumption}{Assumptions}
\crefname{remark}{Remark}{Remarks}
\Crefname{remark}{Remark}{Remarks}

\newcommand{\RegAtt}{\textsc{Nectar}}

\newcommand{\bx}{\mathbf{x}}
\newcommand{\by}{\mathbf{y}}
\newcommand{\bq}{\mathbf{q}}
\newcommand{\bk}{\mathbf{k}}
\newcommand{\bv}{\mathbf{v}}

\newcommand{\bY}{\mathbf{Y}}
\newcommand{\bR}{\mathbf{R}}

\newcommand{\Rot}[1]{\bR_{#1}}

\newcommand{\ip}[2]{\langle #1, #2 \rangle}

\title{{\sc{Nectar}}: Neural Estimation of Cached-Token Attention via Regression}

\author{Joao Monteiro}
\author{Michal Klein}
\author{Pierre Ablin}
\author{Marco Cuturi}

\affiliation{Apple}

\abstract{%
Evaluating softmax attention over a fixed long context requires reading every cached key-value pair for each new query token. Yet, for a given static context (a book, a manual, a legal corpus) the attention is a deterministic function of the input query. As a result, that function can be regressed and studied with a more efficient functional form, e.g. a neural network. We propose \RegAtt{}, which fits a compact neural network to this function for queries drawn from a task-relevant distribution. \RegAtt{} fits two networks per KV-head, at each layer: a \emph{target} network that predicts the attention output and a \emph{score} network that predicts the log-normalizer. The pair plugs into the standard masked self-attention at inference time, replacing the $O(n)$ attention over the cache with a forward pass whose cost does not depend on $n$. Each module carries $|\theta|$ parameters per layer and KV-head, typically much smaller than the $2nd$ KV-cache footprint at the same granularity. We report experiments on models from $1.7$B to $8$B parameters across five long-context datasets. The approximation error tracks the next-token accuracy gap to full attention, and allocating capacity non-uniformly across layers reduces that gap in our ablation. Beyond this analysis of metrics, and most importantly for practical use, we check that the text generation following a question prompt of a model equipped with a \RegAtt{} module matches in semantic content those obtained by giving the same model access to the full cache.%
}

\metadata[Correspondence]{\sffamily \{jmonteiro2, michalk, p\_ablin, cuturi\}@apple.com}
\date{\sffamily May 12, 2026}

\begin{document}

\maketitle

\section{Introduction}
\label{sec:intro}

Large language models (LLMs) now handle a wide range of tasks, from creative writing to complex reasoning \citep{brown2020gpt3,touvron2023llama,yang2025qwen3}. However, their ability to leverage extensive contextual information, such as entire books, technical manuals, or legal corpora, remains constrained by the cost of attention and the memory footprint of key-value (KV) caches \citep{shazeer2019fast,pope2022efficiently}.

\paragraph{The long-context challenge.}
Many practical applications require an LLM to answer detailed queries against a large, static source of knowledge---a regulatory handbook, a novel, a patient record---that is consulted repeatedly over time. Serving such a context requires storing keys and values for every token in a KV-cache. For a context of $n$ tokens with head dimension $d$, $L$ layers, and $H$ KV heads, this cache requires $O(2nLHd)$ memory. To make this concrete: loading \emph{The Great Gatsby} novel (${\sim}61$K tokens) into \texttt{Qwen3-4B} ($L{=}36$, $H{=}8$, $d{=}128$) produces a KV-cache of ${\sim}9$\,GB in \texttt{bfloat16}, already exceeding the model's own ${\sim}8$\,GB \texttt{bfloat16} footprint. At inference, every new query token must attend to the entire cache.

\paragraph{The finetuning bottleneck.}
One approach to incorporating long-term knowledge is to fine-tune models on specific documents \citep{hu2022lora}. However, this compresses document knowledge into model weights---a fundamentally different operation from attending to the document at inference time. Recent comparisons confirm that in-context retrieval outperforms finetuning for detailed factual recall \citep{ovadia2023finetuning}. Fine-tuning cannot preserve the query-dependent nature of attention, which selectively retrieves different information depending on the query. This raises the question underlying this work: can one represent, for a static context, the mapping from queries to attention outputs as a learned object?

\paragraph{Existing approaches.}
Several lines of work have tackled long-context efficiency. \emph{Sparse attention} methods \citep{kitaev2020reformer,wang2020linformer,beltagy2020longformer} attend to a subset of tokens, reducing complexity but potentially missing important context. \emph{KV-cache compression} approaches reduce the cache at inference time by evicting less-important entries \citep{zhang2023h2o,li2024snapkv} or compressing them into fewer tokens \citep{mu2023learning,chevalier2023adapting}. \emph{Subquadratic architectures} such as state-space models \citep{gu2023mamba} and linear attention \citep{katharopoulos2020transformers} avoid the quadratic bottleneck entirely but sacrifice some of the expressiveness of softmax attention. \emph{Retrieval-augmented generation} (RAG) \citep{lewis2020retrieval} retrieves relevant passages from an external store but requires separate retrieval infrastructure. More recently, a family of \emph{context distillation} methods has emerged: \emph{cartridges} \citep{eyuboglu2026cartridges} replace attention to the full long context with attention to a small set of learned ``soft'' tokens, selected adaptively through distillation on the model's next-token predictions, not by mimicking the attention mechanism itself. Follow-up work reduces the trainable parameters for faster compaction \citep{zweiger2026fast}. Related approaches train LoRA-style adapters to emulate the model's behavior when attending to the full context \citep{caccia2025training}, with hypernetwork variants that amortize adapter generation \citep{charakorn2025text}. A unifying thread in these distillation approaches is that they train on the model's \emph{input--output} behavior (e.g.\ next-token prediction), without explicitly targeting the internal attention computation over the long context.

\paragraph{Our approach: regressing attention to static context as a function.}
For a static context, the attention output at every layer is a deterministic function of the query. We regress this function with a compact neural network trained on queries drawn from a task-relevant distribution, i.e.\ queries expected at inference time. Unlike distillation approaches that optimize only end-to-end next-token behavior, \RegAtt{} targets the internal attention computation itself; the loss can also include a next-token-distillation term, and we report the effect of both signals in our experiments. \Cref{fig:overview} illustrates the approach. Our main contributions are:

\begin{figure}[t]
    \centering
    \includegraphics[width=\linewidth]{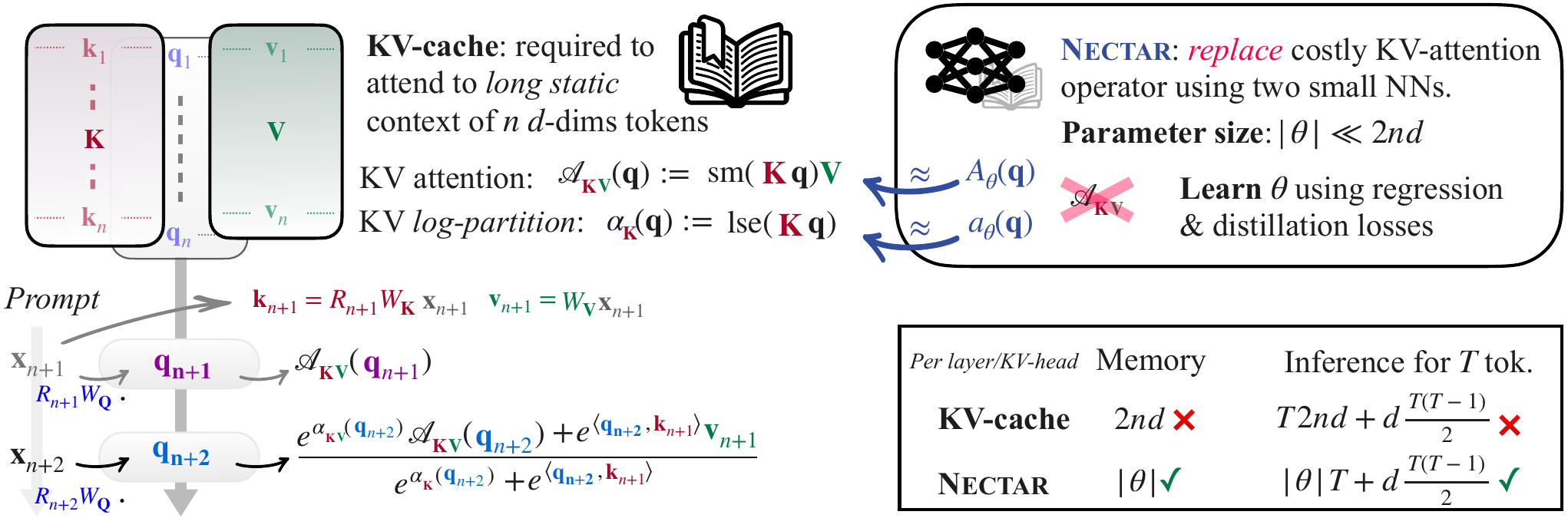}
    \caption{\RegAtt{} replaces the KV-cache attention operators $\mathcal{A}^{\ell,h}$ and $\alpha^{\ell,h}$ with compact networks $A_\theta$ and $a_\theta$, reducing both memory and inference cost from $O(nLHd)$ to $O(|\theta|LH)$. Notice that we have replaced what would have been the standard attention operator over extended keys/values, $\mathcal{A}_{[{\color{red!70!black}\mathbf{K}};\, {\color{red!70!black}\bk_{n+1}}],\, [{\color{green!50!black}\mathbf{V}};\, {\color{green!50!black}\bv_{n+1}}]}$, with a decomposition that emphasizes the log-normalizer $\alpha$ of the original {\color{red!70!black}keys} and the softmax-weighted output of the original {\color{red!70!black}keys} and {\color{green!50!black}values}; see \eqref{eq:inference}. Rewriting attention in that form, we can replace expensive KV-cache attention operators by NNs that can blend in with local causal attention over newly generated tokens.}
    \label{fig:overview}
\end{figure}

\begin{enumerate}[leftmargin=1.5em]
    \item We cast per-context attention approximation as function regression on the attention operator itself. At each layer~$\ell$ and KV-head~$h$, we decompose the attention computation into a normalized output $\mathcal{A}^{\ell,h}$ and a log-normalizer~$\alpha^{\ell,h}$, both determined by the KV-cache of the long context, and learn separate networks for each. Learning $\alpha^{\ell,h}$ separately is needed at inference time: the predicted log-normalizer enters the softmax denominator alongside local causal-attention logits, enabling principled blending of regressed long-range and exact short-range attention~(\S\ref{sec:method},~\S\ref{sec:method:inference}).
    \item We propose \RegAtt{}, a per-context neural function approximator that predicts softmax attention outputs for any query, replacing the $O(n)$ attention computation with a forward pass whose cost does not depend on context length~(\S\ref{sec:method:models}).
    \item We formulate a training objective that combines score and target regression with a next-token-distillation term and report the effect of each component~(\S\ref{sec:method:training}).
    \item We evaluate \RegAtt{} on models from $1.7$B to $8$B parameters across five long-context datasets~(\S\ref{sec:experiments}), and run an ablation on non-uniform capacity allocation across transformer layers~(\S\ref{app:layer-groups}). To evaluate \RegAtt{} we review first the interplay between regression and distillation losses and simple metrics such as next token accuracy~\S\ref{sec:exp:results}. These metrics are, however, not satisfactory per se, as they do not show whether \RegAtt{} modules can help a LLM generate meaningful text at all. We cover that important angle in~\S\ref{sec:exp:generation}, and demonstrate the ability of \RegAtt{} modules to retrieve critical information and package it seamlessly into LLM inference.
\end{enumerate}

\section{Background: Attention, KV-Caching, and Context Distillation}
\label{sec:background}

\subsection{Attention, RoPE, and KV-Caching}
\label{sec:background:attention}

\paragraph{Setup.}
Consider a transformer with $L$ layers and $H$ key-value heads per layer, processing a long context $\bY = (\by_1, \dots, \by_n)$. At each layer~$\ell$, hidden states undergo query, key, and value projections, RoPE-rotated \citep{su2024roformer} masked causal self-attention, and a feed-forward block. We write $\by_j^{(\ell)}$ for the hidden state of token $j$ at the input of layer~$\ell$, with $\by_j^{(1)} = \by_j$.

\paragraph{Keys, values, and attention.}
At layer~$\ell$ and KV-head~$h$, projection matrices $\mathbf{W}_K^{\ell,h}, \mathbf{W}_V^{\ell,h} \!\in\! \R^{d \times d_{\text{model}}}$ produce keys and values, with the key rotated by the position-dependent RoPE matrix:
\begin{equation}
    \bk_j^{\ell,h} = \Rot{j}\,\mathbf{W}_K^{\ell,h} \by_j^{(\ell)}\,, \qquad
    \bv_j^{\ell,h} = \mathbf{W}_V^{\ell,h} \by_j^{(\ell)}\,,
    \label{eq:kv}
\end{equation}
and the query defined symmetrically below. Stacked over the $n$ context tokens, these vectors form the matrices $\mathbf{K}^{\ell,h}, \mathbf{V}^{\ell,h} \!\in\! \R^{n \times d}$. A query token $\bx$ at position $i$ is projected as $\bq = \Rot{i}\,\mathbf{W}_Q^{\ell,h}\bx^{(\ell)}$, and the attention output at layer~$\ell$, head~$h$ is
\begin{equation}
    \text{Attention}(\bq,\, \mathbf{K}^{\ell,h},\, \mathbf{V}^{\ell,h})
    \;=\;
    \softmax\!\left(\tfrac{1}{\sqrt{d}}\,\mathbf{K}^{\ell,h}\,\bq\right)^T \mathbf{V}^{\ell,h}\,.
    \label{eq:attention}
\end{equation}

\paragraph{KV-caching.}
During autoregressive generation, keys and values are computed once over the context and stored. The \emph{KV-cache} is the collection $\{(\mathbf{K}^{\ell,h}, \mathbf{V}^{\ell,h})\}_{\ell,h}$; it requires $2nLHd$ float entries and must reside in memory throughout inference, which is the primary bottleneck for long contexts.

\subsection{Context Distillation: Cartridges and Related Approaches}
\label{sec:background:distillation}

\paragraph{Quadrature view of the KV-cache.}
Several recent methods replace the KV-cache with a smaller learned representation \citep{mu2023learning, chevalier2023adapting, caccia2025training, charakorn2025text, eyuboglu2026cartridges, zweiger2026fast}. Attention can be viewed as an expectation under the softmax distribution, and the sum over $n$ context tokens as its Monte Carlo estimate. Replacing the $n$ cached pairs with $p \ll n$ learnable pairs is analogous to replacing a Monte Carlo sum with a quadrature rule on fewer, optimized nodes \citep{stroud1971}. We refer to this family of approaches as \emph{quadrature attention}. Architecturally, this idea goes back to Perceiver IO \citep{jaegle2021perceiver}, where queries attend to a small set of learned latent vectors; Perceiver IO learns the query projections end-to-end, whereas in our setting all base-model weights are frozen.

\paragraph{Cartridges.}
\citet{eyuboglu2026cartridges} instantiate this idea: at each layer~$\ell$ and head~$h$, the KV-cache $(\mathbf{K}^{\ell,h}, \mathbf{V}^{\ell,h}) \!\in\! \R^{n \times d}$ is replaced with smaller learnable matrices $(\mathbf{W}^{\ell,h}, \mathbf{Z}^{\ell,h}) \!\in\! \R^{p \times d}$ with $p \ll n$, and attention is computed as $\text{Attention}(\bq,\, \mathbf{W}^{\ell,h},\, \mathbf{Z}^{\ell,h})$ using \eqref{eq:attention}. These matrices are the only trainable parameters; all other model weights are frozen. They are fit by distillation: given an instruction--response pair, the model with the virtual KV-cache decodes the response under teacher forcing, and the loss matches its next-token distribution to that of the full-cache model,
\begin{equation}
    \mathcal{L}_{\text{distill}} = \sum_{t=1}^{T} \text{KL}\!\bigl(p_{\text{full}}(\cdot \mid x_{<t}) \;\|\; p_{\text{cart}}(\cdot \mid x_{<t},\, \mathbf{W},\, \mathbf{Z})\bigr)\,.
\end{equation}
Training pairs are generated by self-study, i.e.\ the LLM produces question--answer pairs from document chunks; we adopt the same procedure to define the query distribution used to train \RegAtt{} (\Cref{sec:method:training}).

\paragraph{Related distillation approaches.}
\citet{zweiger2026fast} reduce the cost of training cartridges by matching attention weights rather than output logits. \citet{caccia2025training} train LoRA-style adapters that emulate the full-context model via next-token distillation, and \citet{charakorn2025text} extend this with a hypernetwork that generates adapter weights from the context, amortizing the training cost across contexts. Unlike these methods, \RegAtt{} regresses the attention function at each layer and head, with a next-token-distillation term that can be added on top.

\section{Method: Regressing Attention over Long Contexts}
\label{sec:method}

\paragraph{Log-normalizer and attention.}
Consider a long context $\bY = (\by_1, \dots, \by_n)$ whose KV-caches $(\mathbf{K}^{\ell,h}, \mathbf{V}^{\ell,h})$ have been precomputed at every layer~$\ell$ and KV-head~$h$ as in \eqref{eq:kv}. For a new query token $\bx$ at position $i$, with rotated query $\bq = \Rot{i}\,\mathbf{W}_Q^{\ell,h}\bx^{(\ell)}$, define
\begin{equation}
    \alpha^{\ell,h}(\bq) \;=\; \log \mathbf{1}_n^T \exp\!\left(\tfrac{1}{\sqrt{d}}\,\mathbf{K}^{\ell,h}\,\bq\right)\,,
    \qquad
    \mathcal{A}^{\ell,h}(\bq) \;=\; \softmax\!\left(\tfrac{1}{\sqrt{d}}\,\mathbf{K}^{\ell,h}\,\bq\right)^T \mathbf{V}^{\ell,h}\,,
    \label{eq:lognorm-att}
\end{equation}
where $\mathbf{1}_n$ is the all-ones vector and $\exp$ is elementwise. The \emph{score} $\alpha^{\ell,h}: \R^d \to \R$ and the \emph{target} $\mathcal{A}^{\ell,h}: \R^d \to \R^d$ are deterministic functions of $\bq$, determined by the cached $(\mathbf{K}^{\ell,h}, \mathbf{V}^{\ell,h})$.

\paragraph{Score and target networks.}
We learn, at each layer~$\ell$ and KV-head~$h$, two networks parameterized by $\theta$: a \emph{score} network $a^{\ell,h}_\theta(\bq) \approx \alpha^{\ell,h}(\bq)$ and a \emph{target} network $A^{\ell,h}_\theta(\bq) \approx \mathcal{A}^{\ell,h}(\bq)$. A pair $(a^{\ell,h}_\theta, A^{\ell,h}_\theta)$ forms a \RegAtt{} module. Under grouped-query attention (GQA), a single KV-head is shared by several query heads $h'$: the rotated query $\bq = \Rot{i}\,\mathbf{W}_Q^{\ell,h'}\bx^{(\ell)}$ may come from any of them, but $\mathcal{A}^{\ell,h}$ and $\alpha^{\ell,h}$ depend only on the shared $(\mathbf{K}^{\ell,h}, \mathbf{V}^{\ell,h})$, so a single module per KV-head suffices.

\subsection{Training: Combining Regression with Distillation}
\label{sec:method:training}

\paragraph{Query distribution.}
Training $a^{\ell,h}_\theta$ and $A^{\ell,h}_\theta$ requires ground-truth attention intermediates. We sample queries $(\bx,i)$ from a task-relevant distribution meant to reflect the kind of instruction--response pairs that will be appended after the long context at inference time. These pairs are produced by the self-study procedure~\citep{eyuboglu2026cartridges}, with full construction details in \Cref{app:data}. Although \RegAtt{} replaces only the attention over the long context, the query vectors $\bq^{\ell,h'}_i$ at positions $i>n$ are themselves shaped by causal attention to earlier post-context tokens (e.g.\ the instruction tokens), so the queries seen during training match those produced at inference.

\paragraph{Regression loss.}
Writing $\bq^{\ell,h'}_{i} = \Rot{i}\,\mathbf{W}_Q^{\ell,h'}\bx^{(\ell)}$ for the rotated query at layer~$\ell$, query head~$h'$ (associated with KV-head~$h$), position~$i$, the target regression loss is
\begin{equation}
    \mathcal{L}_{\mathcal{A}}(\theta) = \E_{(\bx,\, i)}\sum_{\ell,\, h,\, h'} \bigl\| A^{\ell,h}_\theta(\bq^{\ell,h'}_i) - \mathcal{A}^{\ell,h}(\bq^{\ell,h'}_i) \bigr\|^2\,.
    \label{eq:regression-loss}
\end{equation}
The score loss $\mathcal{L}_\alpha(\theta)$ is defined analogously, with $(\alpha^{\ell,h}, a^{\ell,h}_\theta)$ in place of $(\mathcal{A}^{\ell,h}, A^{\ell,h}_\theta)$ and a squared difference in place of the squared norm. We combine them into
$\mathcal{L}_{\text{reg}}(\theta) = \lambda_\alpha \mathcal{L}_\alpha + \lambda_{\mathcal{A}} \mathcal{L}_{\mathcal{A}}$.

\paragraph{Total loss.}
We additionally consider a next-token-distillation term $\mathcal{L}_{\text{KL}}(\theta)$ that matches, under teacher forcing, the next-token distribution of the \RegAtt{}-plugged model to that of the full-attention model. The total loss is
\begin{equation}
    \mathcal{L}(\theta)
    \;=\;
    \mathcal{L}_{\text{reg}}(\theta)
    \;+\;
    \lambda_{\text{KL}}\,\mathcal{L}_{\text{KL}}(\theta)\,,
    \label{eq:total-loss}
\end{equation}
and only \RegAtt{} parameters $\theta$ are updated. Setting $\lambda_{\text{KL}}{=}0$ recovers pure regression on the attention computation; setting $\lambda_\alpha{=}\lambda_{\mathcal{A}}{=}0$ recovers the next-token-distillation objective used by \citet{eyuboglu2026cartridges,caccia2025training,charakorn2025text}. The combined loss uses both signals.

\subsection{Inference: Blending \RegAtt{} with Local Causal Attention}
\label{sec:method:inference}

\paragraph{Rewriting standard attention.}
At inference, the long context occupies positions $1,\dots,n$ and newly generated tokens sit at positions $n{+}1,n{+}2,\dots$ At position $t{+}1$ with $t\geq n$, the standard attention output at layer~$\ell$, head~$h$, with $\bq_{t+1} = \Rot{t+1}\,\mathbf{W}_Q^{\ell,h}\bx_{t+1}$ and $\bk_i, \bv_i$ denoting the usual rotated keys and values, reads
\begin{equation}
    \frac{\sum_{i=1}^{t} \exp\bigl(\tfrac{1}{\sqrt{d}}\ip{\bq_{t+1}}{\bk_i}\bigr)\,\bv_i}{\sum_{i=1}^{t} \exp\bigl(\tfrac{1}{\sqrt{d}}\ip{\bq_{t+1}}{\bk_i}\bigr)}
    \;=\;
    \frac{
      \exp\bigl(\alpha^{\ell,h}(\bq_{t+1})\bigr)\,\mathcal{A}^{\ell,h}(\bq_{t+1})
      \;+\;
      \sum_{i=n+1}^{t} \exp\bigl(\tfrac{1}{\sqrt{d}}\ip{\bq_{t+1}}{\bk_i}\bigr)\,\bv_i
    }{
      \exp\bigl(\alpha^{\ell,h}(\bq_{t+1})\bigr)
      \;+\;
      \sum_{i=n+1}^{t} \exp\bigl(\tfrac{1}{\sqrt{d}}\ip{\bq_{t+1}}{\bk_i}\bigr)
    }\,,
    \label{eq:inference-exact}
\end{equation}
where the equality is the standard softmax-aggregation identity, obtained by splitting the sum at $i=n$ and using the definitions in~\eqref{eq:lognorm-att}: $\sum_{i=1}^{n} \exp(\tfrac{1}{\sqrt{d}}\ip{\bq_{t+1}}{\bk_i}) = \exp(\alpha^{\ell,h}(\bq_{t+1}))$ and $\sum_{i=1}^{n} \exp(\tfrac{1}{\sqrt{d}}\ip{\bq_{t+1}}{\bk_i})\,\bv_i = \exp(\alpha^{\ell,h}(\bq_{t+1}))\,\mathcal{A}^{\ell,h}(\bq_{t+1})$. The right-hand side is mathematically identical to the left, but groups the $n$ long-context terms into a single pair $(\alpha^{\ell,h}, \mathcal{A}^{\ell,h})$ that is then renormalized against the local-token logits.
Replacing the exact long-context pair with its regressed counterpart $(a^{\ell,h}_\theta, A^{\ell,h}_\theta)$ gives the \RegAtt{} blended attention output:
\begin{equation}
    \text{ATT}^{\ell,h}(\bx_{t+1};\, \bx_{i\leq t},\, \bY)
    \;=\;
    \frac{
      \exp\bigl(a^{\ell,h}_\theta(\bq_{t+1})\bigr)\,A^{\ell,h}_\theta(\bq_{t+1})
      \;+\;
      \sum_{i=n+1}^{t} \exp\bigl(\tfrac{1}{\sqrt{d}}\ip{\bq_{t+1}}{\bk_i}\bigr)\,\bv_i
    }{
      \exp\bigl(a^{\ell,h}_\theta(\bq_{t+1})\bigr)
      \;+\;
      \sum_{i=n+1}^{t} \exp\bigl(\tfrac{1}{\sqrt{d}}\ip{\bq_{t+1}}{\bk_i}\bigr)
    }\,.
    \label{eq:inference}
\end{equation}
Having access to a scalar log-normalizer for the long-context contribution is what enables this blending, since it sets the relative weight of the long-context summary against the local logits in a single softmax. We therefore learn $a^{\ell,h}_\theta$ jointly with $A^{\ell,h}_\theta$, unless the architecture yields it in closed form from the parameters of $A^{\ell,h}_\theta$ (as in the quadrature family of \S\ref{sec:method:models}). Equation \ref{eq:inference} is implemented by prepending $a^{\ell,h}_\theta$ as a virtual logit and $A^{\ell,h}_\theta$ as its corresponding value to the local attention arrays, then running the standard softmax. No change to the base-model attention kernel is required.

\subsection{Architectures for Score and Target Networks}
\label{sec:method:models}

\paragraph{Quadrature.}
Following the cartridge form of \S\ref{sec:background:distillation}, we store $p \ll n$ learnable pairs $(\mathbf{W}, \mathbf{Z}) \!\in\! \R^{p \times d}$ per head, initialized, as in \citet{eyuboglu2026cartridges}, from the first $p$ rows of the ground-truth KV-cache $(\mathbf{K}^{\ell,h}, \mathbf{V}^{\ell,h})$. Then $\theta = (\mathbf{W}, \mathbf{Z})$, and the score and target networks take the closed form
$a^{\ell,h}_\theta(\bq) = \log \mathbf{1}_p^T \exp(\tfrac{1}{\sqrt{d}}\mathbf{W}\bq)$ and $A^{\ell,h}_\theta(\bq) = \softmax(\tfrac{1}{\sqrt{d}}\mathbf{W}\bq)^T \mathbf{Z}$. Unlike cartridges, these parameters are fit by the regression loss on $(\alpha^{\ell,h}, \mathcal{A}^{\ell,h})$, with an optional next-token-distillation term. The parameter count is $2pd$ per head. Since the score follows from $\mathbf{W}$ already used for the target, we set $\lambda_\alpha{=}0$ and $\lambda_{\mathcal{A}}{=}1$ for quadrature modules.

\paragraph{MLP.}
The score function $\alpha^{\ell,h}(\bq) = \log \mathbf{1}_n^T \exp(\tfrac{1}{\sqrt{d}}\mathbf{K}^{\ell,h}\bq)$ is a log-sum-exp of linear forms in $\bq$, and therefore convex in $\bq$. In recent work, \citet{amips2025} exploit this convexity for top-1 retrieval to amortize maximum inner product search (MIPS), a close relative of the log-sum-exp function. They do so by regressing the MIPS objective with an input-convex network \citep{amos2017input}, in which each hidden layer re-injects the input via a skip term. Our MLP family borrows the same skip-to-input structure, applying it to both score and target heads, with SiLU activations. For such a network of depth $D$ and output dimension $q$, the input query $\bq$ is mapped to $\mathbf{h}\!\in\!\R^q$ via
\begin{equation}
    \mathbf{h}_0 = \sigma\bigl(\mathbf{U}_0 \bq + \mathbf{b}_0\bigr),\qquad
    \mathbf{h}_{k+1} = \sigma\bigl(\mathbf{V}_k \bq + \mathbf{U}_k \mathbf{h}_k + \mathbf{b}_k\bigr),\;\;
    \mathbf{h} = \mathbf{h}_{D-1}\,.
    \label{eq:mlp-backbone}
\end{equation}
The output dimension is equal to the input dimension $q=d$ for the target head, whereas $q=1$ for the score head. All intermediate representations have a fixed width $w$ that is computed automatically to satisfy, given the user-defined depth, a certain parameter count (see \citep[\S3]{amips2025}). Optionally, we have considered models in which target and score heads share a common backbone MLP of the same type, mapping query to an intermediate representation. In addition to the depth of that backbone $D_b$, we consider the $D_s$ hidden layers of the same score head and $D_t$ for the target head. We write configurations as $(D_b, D_s, D_t)$, e.g.\ $(0, 4, 4)$ for no backbone with depth-4 heads; the connection between this score head and the amortized MIPS limit is detailed in \Cref{app:support}. For the MLP family we use $\lambda_\alpha{=}0.1$, $\lambda_{\mathcal{A}}{=}1$.

\section{Experiments}
\label{sec:experiments}

We evaluate \RegAtt{} on its ability to approximate softmax attention over long contexts, training on models from $1.7$B to $8$B parameters across five long-context datasets.

\subsection{Setup: Base Models, Datasets and Hyperparameter Choices}
\label{sec:exp:setup}

We train \RegAtt{} modules on models spanning 1.7B to 8B parameters: \texttt{Qwen3-1.7B} (28 layers, $d{=}128$, 8~KV heads), \texttt{Qwen3-4B} (36 layers, $d{=}128$, 8~KV heads), \texttt{Qwen2.5-7B-1M} (28 layers, $d{=}128$, 4~KV heads), and \texttt{Qwen3-8B} (64 layers, $d{=}128$, 8~KV heads) \citep{yang2025qwen3}.

\textbf{Datasets.}
We evaluate on five long-context datasets: three public-domain novels---\emph{The Great Gatsby} \citep{fitzgerald1925gatsby}, \emph{The Time Machine} \citep{wells1895timemachine}, and \emph{Heart of Darkness} \citep{conrad1899darkness}---and two benchmark documents: \emph{Introduction to Intellectual Property} \citep{kline2024introduction} (hereafter IP~Intro) and a clinical case compilation from LongHealth \citep{adams2024longhealth}. For each, we construct synthetic instruction-following pairs that largely follow the self-study procedure of \citet{eyuboglu2026cartridges}, and cache ground-truth attention intermediates from the base model (details in Appendix~\ref{app:data}, \ref{app:caching}). \Cref{tab:datasets} summarizes the context lengths and training set sizes.

\begin{wraptable}{r}{0.43\textwidth}
\vspace{-12pt}
\centering
\caption{Dataset statistics.}
\label{tab:datasets}
\footnotesize
\begin{tabular}{@{}l r r@{}}
\toprule
Dataset & Tokens & Samples \\
\midrule
Time Machine        & 40\,022  & 117\,111 \\
Heart of Darkness   & 49\,313  & 117\,120 \\
LongHealth          & 59\,031  & 103\,592 \\
Great Gatsby        & 61\,651  & 114\,896 \\
IP Intro            & 122\,466 & 228\,193 \\
\bottomrule
\end{tabular}
\vspace{-7pt}
\end{wraptable}

\begin{figure}[t]
    \centering
    \raisebox{.7cm}{\rotatebox{90}{\fontsize{6}{7}\selectfont Token Acc.\ Gap (\%)}}\,%
    \includegraphics[width=0.23\linewidth]{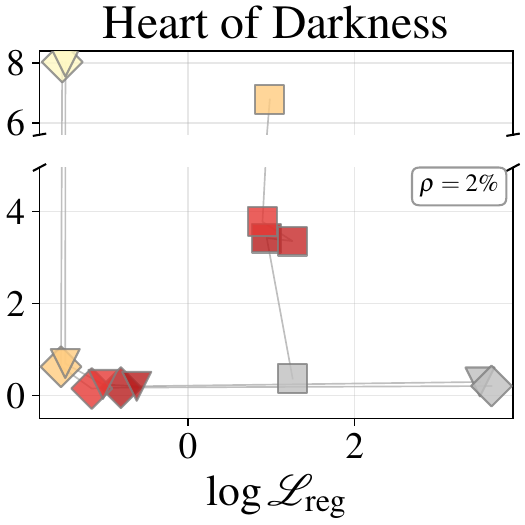}
    \includegraphics[width=0.23\linewidth]{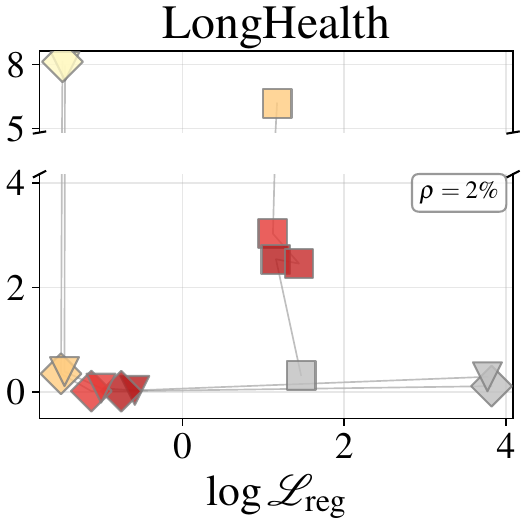}
    \includegraphics[width=0.23\linewidth]{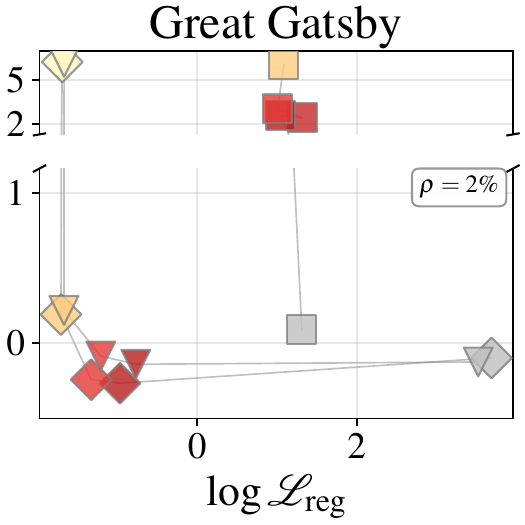}
    \includegraphics[width=0.23\linewidth]{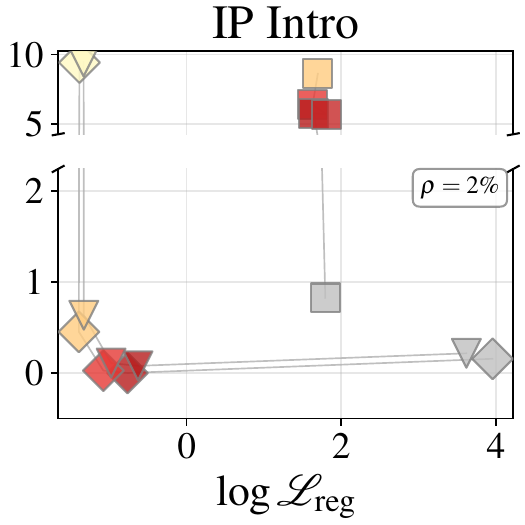}
    \\\vskip.1cm
    \includegraphics[width=\linewidth]{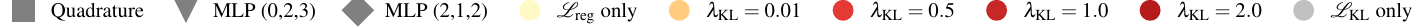}
    \caption{Token-accuracy gap (\%, lower is better) vs.\ $\log \mathcal{L}_{\text{reg}}$ for \texttt{Qwen3-1.7B} (with YaRN $4{\times}$) at $\rho{=}2\%$. Marker shape denotes architecture; color denotes $\lambda_{\text{KL}}$.}
    \label{fig:rl-1.7B}
\end{figure}

\begin{figure}[t]
    \centering
    \raisebox{.7cm}{\rotatebox{90}{\fontsize{6}{7}\selectfont Token Acc.\ Gap (\%)}}\,%
    \includegraphics[width=0.23\linewidth]{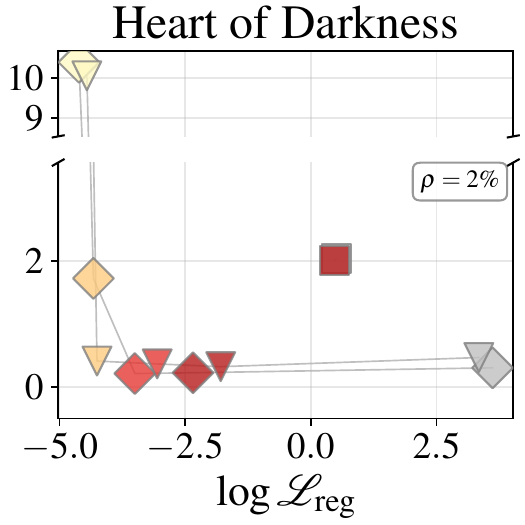}
    \includegraphics[width=0.23\linewidth]{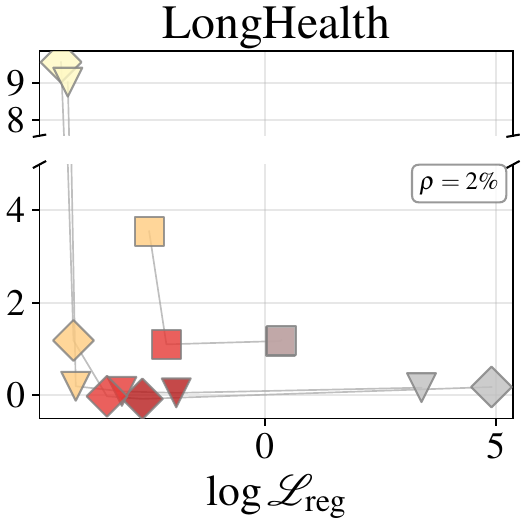}
    \includegraphics[width=0.23\linewidth]{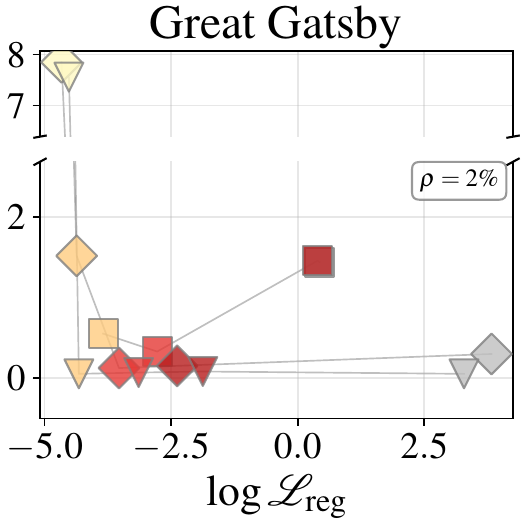}
    \includegraphics[width=0.23\linewidth]{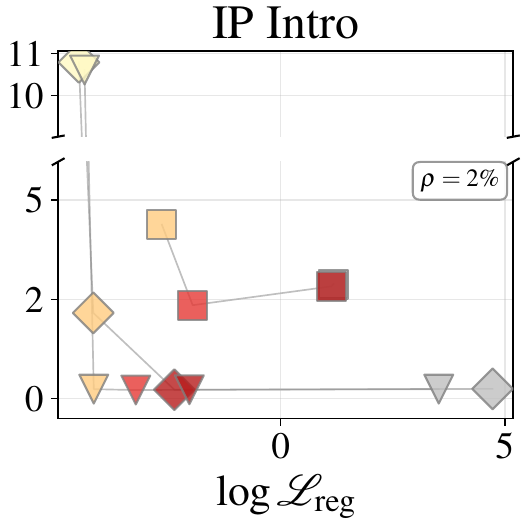}
    \\\vskip.1cm
    \includegraphics[width=\linewidth]{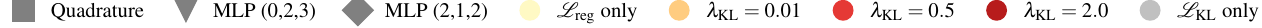}
    \caption{Token-accuracy gap (\%) vs.\ $\log \mathcal{L}_{\text{reg}}$ for \texttt{Qwen3-4B} (with YaRN $4{\times}$) at $\rho{=}2\%$.}
    \label{fig:rl-4B}
\end{figure}

\begin{figure}[t]
    \centering
    \raisebox{.7cm}{\rotatebox{90}{\fontsize{6}{7}\selectfont Token Acc.\ Gap (\%)}}\,%
    \includegraphics[width=0.23\linewidth]{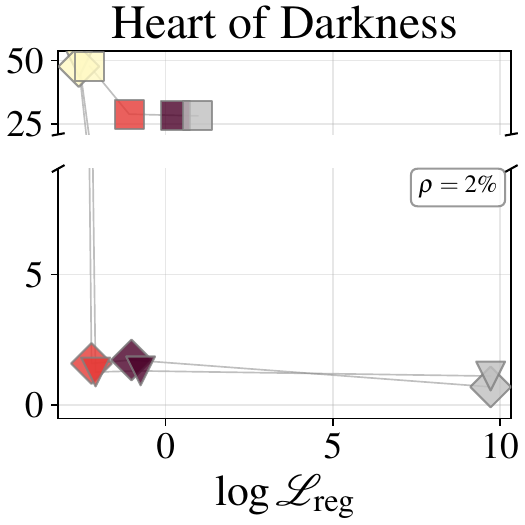}
    \includegraphics[width=0.23\linewidth]{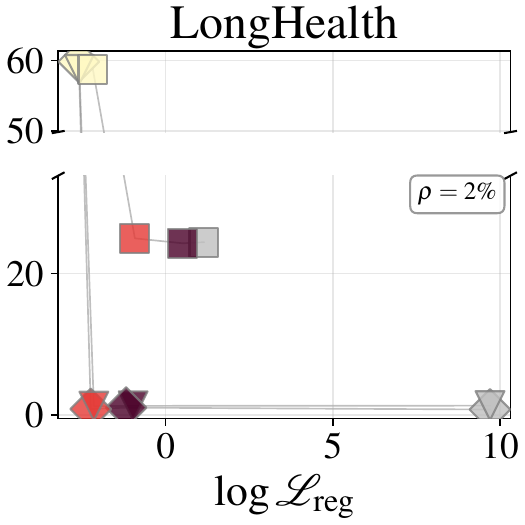}
    \includegraphics[width=0.23\linewidth]{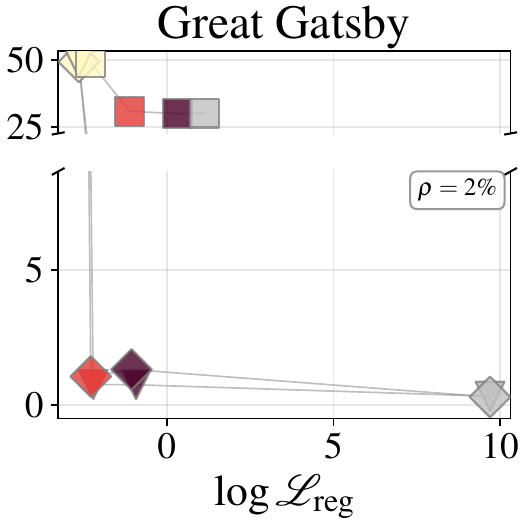}
    \includegraphics[width=0.23\linewidth]{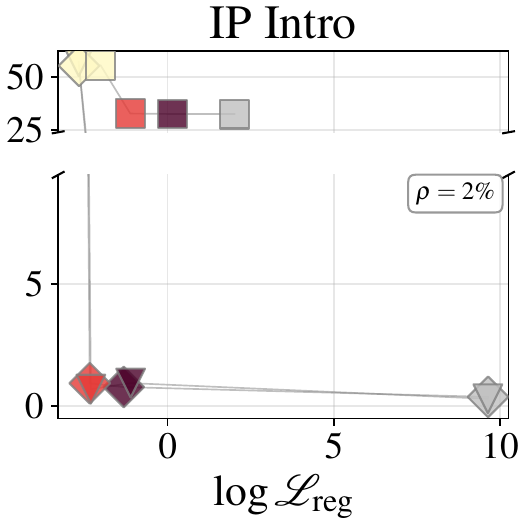}
    \\\vskip.1cm
    \includegraphics[width=\linewidth]{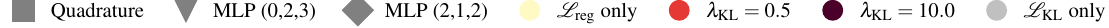}
    \caption{Token-accuracy gap (\%) vs.\ $\log \mathcal{L}_{\text{reg}}$ for \texttt{Qwen2.5-7B-1M} at $\rho{=}2\%$.}
    \label{fig:rl-7B-pf2}
\end{figure}

\textbf{Context-length extension with YaRN.}
The Qwen3 models are pretrained with an attention context length of 32\,768 tokens. All our documents exceed this limit. We apply YaRN \citep{peng2023yarn} with a $4{\times}$ scaling factor, which rescales the rotary-embedding frequency basis to support contexts of up to 125\,000 tokens without retraining. \texttt{Qwen2.5-7B-1M} natively supports up to 1M tokens and needs no extension, so its results are not confounded by YaRN; for the Qwen3 models, YaRN is shared between the full-attention reference and the \RegAtt{}-equipped model, so gap metrics remain directly comparable.

\textbf{Parameter fraction.}
We define the \emph{parameter fraction} $\rho$ as the \RegAtt{} module's parameter budget expressed as a percentage of the per-head KV-cache size ($2nd$ parameters per layer, where $n$ is context length and $d$ the head dimension). We evaluate at various $\rho\!\in\!\{0.5\%, 2\%, 5\%, 10\%\}$ using two architecture families: MLP and Quadrature (Appendix~\ref{app:arch}). When using non-uniform capacity allocation across layers, we normalize the per-group multipliers so that the average $\rho$ is preserved; see Appendix~\ref{app:layer-groups} for an ablation comparing weighted and uniform allocation.

\textbf{Training.}
All \RegAtt{} modules are trained with AdamW, cosine learning-rate schedule with peak rate at $10^{-4}$, and gradient clipping. Only \RegAtt{} parameters are updated. Full hyperparameters are in Appendix~\ref{app:hyperparameters}. The sweep spans three loss configurations: pure regression ($\lambda_{\text{KL}}{=}0$); regression plus distillation ($\lambda_{\text{KL}}{>}0$ with $\lambda_\alpha{=}0.1$, $\lambda_{\mathcal{A}}{=}1$); distillation only ($\lambda_\alpha{=}\lambda_{\mathcal{A}}{=}0, \lambda_{\text{KL}}=1$). The distillation-only setting with the Quadrature architecture is identical to the Cartridges training \citep{eyuboglu2026cartridges} and serves as our reference baseline (gray squares in all figures).

\textbf{Metrics.}
We focus on the \emph{token-accuracy gap}: the difference in next-token prediction accuracy between the model using \RegAtt{} and the model using the full KV-cache, evaluated on held-out question-answer pairs. A gap of zero means \RegAtt{} results in a similar behavior to full attention; positive values indicate degradation. As complementary diagnostics we also report, in the appendix, an LM cross-entropy gap (\Cref{app:scatter}) and a per-layer/head \emph{relative transport error} $\mathcal{E}_{\text{rel}}$ used as an alternative evaluation metric for regression (\Cref{app:rte-def}).

\begin{figure}[t]
    \centering
    \raisebox{.7cm}{\rotatebox{90}{\fontsize{6}{7}\selectfont Token Acc.\ Gap (\%)}}\,%
    \includegraphics[width=0.23\linewidth]{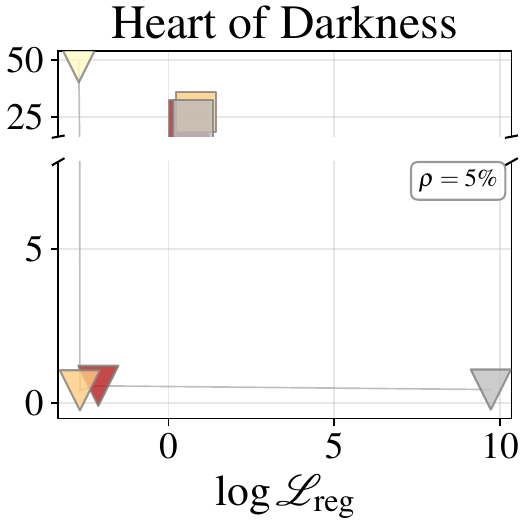}
    \includegraphics[width=0.23\linewidth]{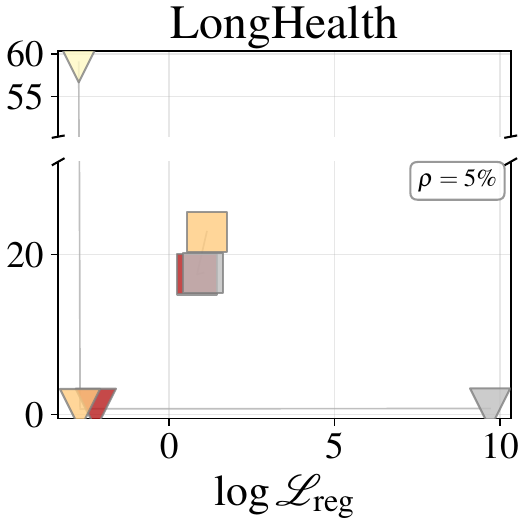}
    \includegraphics[width=0.23\linewidth]{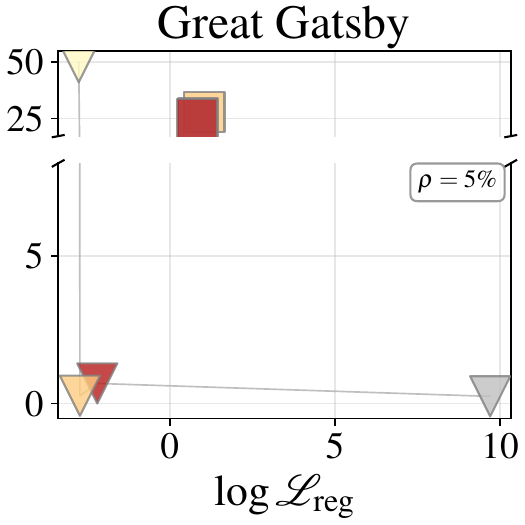}
    \includegraphics[width=0.23\linewidth]{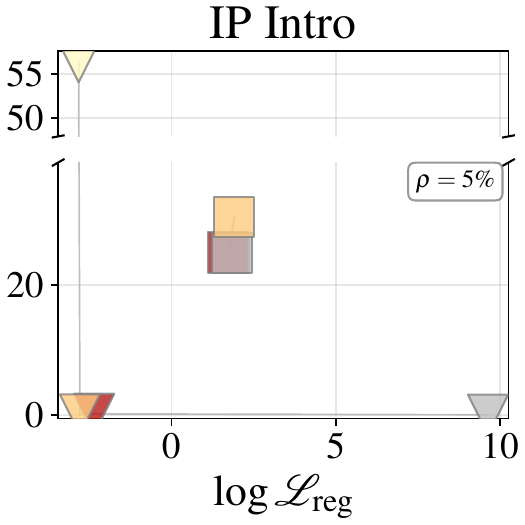}
    \\\vskip.1cm
    \includegraphics[width=\linewidth]{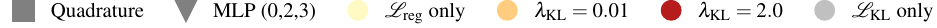}
    \caption{Token-accuracy gap (\%) vs.\ $\log \mathcal{L}_{\text{reg}}$ for \texttt{Qwen2.5-7B-1M} at $\rho{=}5\%$.}
    \label{fig:rl-7B-pf5}
\end{figure}

\subsection{Approximation Fidelity vs.\ Next-Token Agreement on Evaluation Q/As}
\label{sec:exp:results}

\paragraph{Setup of the plots.}
\Cref{fig:rl-1.7B,fig:rl-4B,fig:rl-7B-pf2,fig:rl-7B-pf5} plot the token-accuracy gap against the total regression loss $\mathcal{L}_{\text{reg}} = 0.1\,\mathcal{L}_\alpha + \mathcal{L}_{\mathcal{A}}$ for \texttt{Qwen3-1.7B}, \texttt{Qwen3-4B}, and \texttt{Qwen2.5-7B-1M} at $\rho \in \{2\%, 5\%\}$ (score loss ignored for quadrature). Datasets are ordered by context length.
The corresponding plots for \texttt{Qwen2.5-7B-1M} at $\rho{=}10\%$ and for \texttt{Qwen3-8B} at $\rho{=}2\%$ are in \Cref{fig:rl-7B-pf10,fig:rl-8B} (Appendix~\ref{app:scatter}).

\textbf{Results.} Lower $\log\mathcal{L}_{\text{reg}}$ is associated with a smaller token-accuracy gap across the four models and datasets we evaluate. MLP architectures reach the lowest regression loss in most cases; Quadrature models track the trend from higher. Corresponding plots against target $\mathcal{E}_{\text{rel}}$ are in Appendix~\ref{app:ta-vs-rte}. MLP models often attain their best token-accuracy gap (specifically if it cannot be saturated to 0) when including regression in their objectives. This is not necessarily the case for quadrature modules, whose performance tends to improve with a smaller regression loss term, except for \texttt{Qwen3-4B}. While the best quadrature modules perform similarly to the best \RegAtt{} MLP modules with the smallest model \texttt{Qwen3-1.7B}, the gap widens significantly as the model size is increased, supporting our hypothesis that the attention operator for larger depth and heads is harder to model with quadrature points and likely requires a general-purpose MLP with more capacity for the same parameter count.

\subsection{Assessing Generation Quality using LLM as a Judge}
\label{sec:exp:generation}

Token-accuracy gap, next-token cross-entropy, and per-head MSE/RTE on cached attention targets are convenient surrogates: they are cheap to compute, align with the training loss, and reward any approximator that reproduces the base model's internal states. What a user of a KV-cache replacement actually cares about, however, is far more fine-grained: given the long context or its approximation, can the module support \emph{free-form} generation that remains faithful to the source document? This requires coherent multi-token rollouts over a distribution that drifts away from the teacher-forced one used at training time, under arbitrary instructions, and cannot be read off from a single-step regression or next-token metric. We therefore sound a note of caution: a module can attain a near-zero token-accuracy gap and low attention MSE while producing generations that are, in substance, nonsense.

\textbf{Low metrics can still result in nonsensical generation.} Every Qwen3 model we trained (\texttt{Qwen3-1.7B}, \texttt{Qwen3-4B}, \texttt{Qwen3-8B}) failed this generation test regardless of approximator: both Cartridges (Quadrature with distillation-only loss) and \RegAtt{}~(MLP) with low regression loss and low token-accuracy gap produced outputs unrelated to the long context when asked to answer, summarize, quote, or paraphrase. We hypothesize that the culprit is not the approximator but the context-length extension: the Qwen3 models must be YaRN-extended \citep{peng2023yarn} to ingest our documents, and we observe that the base \emph{full-attention} Qwen3 model, given the same YaRN'd long contexts, \textit{fails} to produce coherent answers, while it remains fluent on short prompts. In other words, the ceiling set by the YaRN'd base model is already below the threshold of usable generation, so any approximator trained against it inherits the failure. \texttt{Qwen2.5-7B-1M} \citep{qwen2.5-1m}, by contrast, natively supports contexts up to 1M tokens and requires no YaRN rescaling; We observe that its full-attention outputs on our long contexts are coherent and on-topic. We therefore restrict the generation-quality evaluation in this section to \texttt{Qwen2.5-7B-1M}. We view this as an important remark for any line of work that replaces the KV-cache: matching attention outputs or next-token distributions is necessary but not sufficient, and an end-to-end generation-quality protocol against the full-attention baseline is, in our view, required before claiming that a KV-cache replacement works.

\textbf{Evaluation pipeline.} We evaluate \RegAtt{} modules on \texttt{Qwen2.5-7B-1M} via an LLM-as-judge protocol. To limit the cost of LLM judge queries, we select a representative subset of three datasets: Heart of Darkness, LongHealth, and Great Gatsby. For a set of held-out instructions, both the \RegAtt{}-augmented model and the base model (with access to the full KV-cache) generate responses autoregressively. \Cref{tab:gen-sample} shows a representative QA example from Great Gatsby; both models retrieve the correct answer, with \RegAtt{} producing a semantically equivalent response (additional examples in Appendix~\ref{app:gen-side-by-side}). A commercial-grade LLM judge then rates each response against the ground-truth answer on a $1$ to $5$ semantic-similarity scale (Appendix~\ref{app:eval-pipeline}). We define the \emph{similarity score gap} $\Delta_s = \bar{s}_{\text{base}} - \bar{s}_{\RegAtt{}}$ \textit{(lower is better)}, where $\bar{s}$ denotes the mean judge score; a gap of zero indicates that \RegAtt{} matches the generation quality of full attention, and a gap of $k$ means that \RegAtt{} responses are, on average, $k$ points less similar to the ground-truth than the base model's. Evaluations span four task types: \textit{question answering, summarization, quoting, and paraphrasing}; covering various generation demands from short factual recall to long-form synthesis.

\begin{table}[]
\centering
\caption{QA generation example from Great Gatsby. In this example, the production of a \texttt{Qwen2.5-7B-1M} model equipped with a pre-fitted \RegAtt{} module is semantically equivalent to the same model attending to the entire KV-cache of Great Gatsby. On the other hand, the quadrature/cartridges approximation is not able to generate meaningful content.}
\label{tab:gen-sample}
\footnotesize
\begin{tabular}{@{}p{0.14\linewidth} p{0.82\linewidth}@{}}
\toprule
\textbf{Instruction} & What physical accomplishment is Tom Buchanan known for? \\
\midrule
\textbf{Ground truth} & Tom Buchanan was one of the most powerful ends that ever played football at New Haven. \\
\midrule
\textbf{\RegAtt{} (MLP)} & Based on the passage, Tom Buchanan is known for being one of the most powerful ends that ever played football at New Haven. Specifically, he was a powerful football player who had played in the finals at New Haven. \\
\midrule
\textbf{Quadrature} & Based of being driving side-room trains and everythingage Platt and lemon polographer over the St. Specifically sand breath, and driveser. Specifically ignore tradingcyour coach. The photograph.17 vitality in Gatsby's Series. Wolfshiemts Wilson Buchanan Buchanan, Doc Carson, and sitting with Myrtle with Meyer \\
\bottomrule
\end{tabular}

\end{table}

\textbf{Aggregation.} Each reported $\Delta_s$ selects the best configuration (lowest gap) \emph{within each aggregation unit} and averages over those units: \Cref{tab:gen-pf} takes the best run per dataset at a fixed capacity and averages over the three datasets, whereas \Cref{tab:gen-distill-w} fixes $\lambda_{\text{KL}}$, takes the best run in each (dataset, capacity) cell, and averages over all such cells---comparing distillation weights across capacities rather than reporting best-case performance. \Cref{tab:gen-quality} instead fixes a \emph{single} best-overall run per dataset (chosen by the four-task mean) and reports that run's per-task gaps averaged over the three datasets, so a model's task cells all come from one configuration. Reported $\pm$ values are $95\%$ confidence intervals (Student's $t$) over those same units, and are correspondingly wide for the small-sample aggregations.

\textbf{Results.} \Cref{tab:gen-pf} shows that MLP modules improve monotonically with capacity, going from $\Delta_s{=}0.28$ at $\rho{=}2\%$ to $\Delta_s{=}{-}0.14$ at $\rho{=}10\%$---crossing into negative territory, indicating that at sufficient capacity MLP \RegAtt{} surpasses the base model on average. Quadrature modules remain near $\Delta_s{\approx}2.2$ regardless of capacity, consistent with the qualitative outputs shown in Appendix~\ref{app:gen-side-by-side}.

\Cref{tab:gen-distill-w} examines the interaction between the regression and distillation objectives. The ``Pure $\mathcal{L}_{\text{KL}}$'' column trains with the KL loss alone ($\lambda_\alpha {=} \lambda_{\mathcal{A}} {=} 0$, $\lambda_{\text{KL}} {=} 1$); the remaining columns combine the full regression loss with distillation at varying $\lambda_{\text{KL}}$. Averaged over datasets and capacities, adding the regression term improves over pure distillation: the mixed objective reaches $\Delta_s{=}0.25$ at $\lambda_{\text{KL}}{=}0.01$ and $0.28$ at $\lambda_{\text{KL}}{=}2$, both below pure distillation's $0.36$, indicating that the regression component boosts generation quality. The advantage shrinks only when distillation is heavily over-weighted ($\lambda_{\text{KL}}{=}10$, $\Delta_s{=}0.53$). The best $\lambda_{\text{KL}}$ nonetheless varies by dataset and capacity (see Appendix~\ref{app:gen-breakdown} for the full breakdown). For Quadrature, no objective drives $\Delta_s$ below $2.2$.

\begin{table}[t]
\centering
\footnotesize
\begin{minipage}[t]{0.38\linewidth}
\centering
\caption{$\Delta_s$ ($\downarrow$) vs.\ $\rho$.}
\label{tab:gen-pf}
\resizebox{\linewidth}{!}{%
\begin{tabular}{@{}l ccc@{}}
\toprule
& \multicolumn{3}{c}{$\rho$} \\
\cmidrule(lr){2-4}
& $2\%$ & $5\%$ & $10\%$ \\
\midrule
MLP    & $\mathbf{0.28}${\tiny$\pm0.26$} & $\mathbf{-0.02}${\tiny$\pm0.39$} & $\mathbf{-0.14}${\tiny$\pm0.34$} \\
Quad.\ & $2.25${\tiny$\pm0.22$} & $2.23${\tiny$\pm0.21$} & $2.22${\tiny$\pm0.26$} \\
\bottomrule
\end{tabular}}
\end{minipage}\hfill
\begin{minipage}[t]{0.60\linewidth}
\centering
\caption{$\Delta_s$ ($\downarrow$) vs.\ distill.\ weight. ``Pure $\mathcal{L}_{\text{KL}}$'' uses only the KL loss; others combine with regression.}
\label{tab:gen-distill-w}
\resizebox{\linewidth}{!}{%
\begin{tabular}{@{}l c ccc@{}}
\toprule
& Pure $\mathcal{L}_{\text{KL}}$ & \multicolumn{3}{c}{$\mathcal{L}_{\text{reg}} + \lambda_{\text{KL}}\,\mathcal{L}_{\text{KL}}$} \\
\cmidrule(lr){2-2} \cmidrule(lr){3-5}
&  & $0.01$ & $2$ & $10$ \\
\midrule
MLP  & $\mathbf{0.36}${\tiny$\pm0.33$} & $\mathbf{0.25}${\tiny$\pm0.13$} & $\mathbf{0.28}${\tiny$\pm0.37$} & $\mathbf{0.53}${\tiny$\pm0.24$} \\
Quad.\ & $2.22${\tiny$\pm0.10$} & $2.26${\tiny$\pm0.14$} & $2.22${\tiny$\pm0.13$} & $2.26${\tiny$\pm0.21$} \\
\bottomrule
\end{tabular}}
\end{minipage}
\end{table}

\begin{table}[h]
\centering
\caption{Similarity score gap $\Delta_s$ ($\downarrow$) by task type, averaged across datasets. Negative values indicate \RegAtt{} surpasses the base model.}
\label{tab:gen-quality}
\footnotesize
\begin{tabular}{@{}l ccccc@{}}
\toprule
& Paraphr.\ & QA & Quoting & Summ.\ & Overall \\
\midrule
\RegAtt{} (MLP) & $\mathbf{0.19}${\tiny$\pm0.59$} & $\mathbf{-0.46}${\tiny$\pm0.66$} & $\mathbf{-0.10}${\tiny$\pm1.01$} & $\mathbf{-0.33}${\tiny$\pm1.06$} & $\mathbf{-0.18}${\tiny$\pm0.46$} \\
Quadrature      & $2.60${\tiny$\pm0.53$} & $2.01${\tiny$\pm0.82$} & $2.09${\tiny$\pm0.41$} & $2.16${\tiny$\pm0.27$} & $2.22${\tiny$\pm0.42$} \\
\bottomrule
\end{tabular}
\end{table}

\begin{table}[h]
\centering
\caption{Similarity score gap $\Delta_s$ ($\downarrow$) per dataset, best run per architecture. Negative values indicate \RegAtt{} surpasses the base model. $\pm$: 95\% CI. Best per column in bold.}
\label{tab:gen-quality-per-dataset}
\footnotesize
\begin{tabular}{@{}l cccc@{}}
\toprule
& Heart of Dark.\ & LongHealth & Great Gatsby & Mean \\
\midrule
\RegAtt{} (MLP) & $\mathbf{-0.15}${\tiny$\pm0.87$} & $\mathbf{-0.30}${\tiny$\pm0.38$} & $\mathbf{-0.08}${\tiny$\pm0.67$} & $\mathbf{-0.18}${\tiny$\pm0.28$} \\
Quadrature      & $2.17${\tiny$\pm0.68$} & $2.14${\tiny$\pm0.35$} & $2.33${\tiny$\pm0.43$} & $2.22${\tiny$\pm0.26$} \\
\bottomrule
\end{tabular}
\end{table}

\Cref{tab:gen-quality} reports $\Delta_s$ for the best MLP and best Quadrature configurations, broken down by task type and averaged across datasets. The overall MLP gap is $-0.18$, meaning \RegAtt{} (MLP) surpasses the base model on average; the effect is strongest on QA ($\Delta_s{=}{-}0.46$) and summarization ($\Delta_s{=}{-}0.33$), while paraphrasing shows a modest degradation ($\Delta_s{=}0.19$). Quadrature remains at $\Delta_s{\approx}2.2$ on every task, consistent with its incoherent outputs (Appendix~\ref{app:gen-side-by-side}). \Cref{tab:gen-quality-per-dataset} unpacks the cross-task average by dataset: MLP surpasses the base model on all three datasets, most strongly on LongHealth ($-0.30$). Per-task and per-dataset breakdowns are in Appendix~\ref{app:gen-breakdown}. These results use a uniform random subsample of held-out instructions; on a deliberately out-of-distribution \emph{challenge} split given by the held-out instructions least similar to any training pair, although small, a gap can be observed (Appendix~\ref{app:challenge}).

\section{Conclusion}
\label{sec:conclusion}
We presented \RegAtt{}, which fits, per context, two networks per layer and KV-head: a score network that predicts the log-normalizer of attention and a target network that predicts the attention output. This formulation includes quadrature attention approximations for KV-caches as a special case~\citep{eyuboglu2026cartridges} and proposes to learn these modules using direct input-output regression, in addition to end-to-end losses. At inference the pair replaces the $O(n)$ attention over the KV-cache with a forward pass whose cost does not depend on $n$, and blends with local causal attention through the standard softmax. When looking at regression/accuracy metrics, across \texttt{Qwen3-1.7B}, \texttt{Qwen3-4B}, \texttt{Qwen2.5-7B-1M}, and \texttt{Qwen3-8B} on five long-context datasets, lower regression loss is associated with a smaller next-token accuracy gap relative to full attention, notably as models get larger and the task is harder. Training with the combined regression-plus-next-token-distillation loss reaches lower accuracy gap than either term alone in our runs, and allocating per-layer capacity non-uniformly reduces the gap compared to a uniform allocation in the ablation of Appendix \ref{app:layer-groups}. While worth having as hill-climbing metrics, these accuracies do not reflect practical use cases, which is why we turn instead to the only realistic setting that truly matters: whether a base LLM equipped with a \RegAtt{} module is able to generate accurate text that would be comparable to that generated with the full KV cache. Going beyond these token-level metrics, we show that MLP \RegAtt{} modules surpass the base model's open-ended generation quality: on \texttt{Qwen2.5-7B-1M} the LLM-as-judge similarity-score gap reaches $\Delta_s{=}{-}0.14$ at $\rho{=}10\%$ on a $1$--$5$ scale (negative means \RegAtt{} scores higher than the base model), while the Quadrature baseline (equivalent to the Cartridges recipe under our pair generation with a pure distillation loss) remains above $\Delta_s{\approx}2.2$ regardless of capacity. These gains come with substantial inference speedups (Appendix~\ref{app:generation-bmk}).

\textbf{Limitations and Future work.} Each \RegAtt{} module is trained for a single fixed context and training requires both selecting relevant question/answer pairs through self-study and evaluating ground-truth attention targets on such pairs, which has $O(n)$ cost per sequence. Generation quality is uneven across task types and degrades more on the longest context (IP~Intro, ${\sim}122$k tokens), a counter-performance that will require further study. Potential directions include cross-context generalization via meta-learning, amortized conditioning on the context, fusion of multiple \RegAtt{} modules by direct summation or MoE type mechanisms gated on \RegAtt{} log-normalizer or attention values, as well as exploiting document structure to build compositional approximators.

\bibliographystyle{plainnat}
\bibliography{biblio}

\appendix
\section{Additional Experimental Details}
\label{app:experiments}

\subsection{Data Preparation}
\label{app:data}

Training and evaluating \RegAtt{} requires long-context sequences paired with ground-truth attention targets. We construct synthetic instruction-following datasets from public-domain novels---\emph{The Great Gatsby} \citep{fitzgerald1925gatsby}, \emph{The Time Machine} \citep{wells1895timemachine}, and \emph{Heart of Darkness} \citep{conrad1899darkness}---as well as benchmark documents from \emph{Introduction to Intellectual Property} \citep{kline2024introduction} (IP~Intro) and LongHealth \citep{adams2024longhealth}.

\paragraph{Chunk sampling.}
Each document is split into overlapping word-level chunks of approximately 2\,000 words. Multiple chunks are drawn per document with configurable overlap, ensuring broad coverage of the source text.

\paragraph{Pair generation.}
An instruction-tuned LLM (\texttt{Qwen3-8B}; \citealp{yang2025qwen3}) generates instruction--response pairs from each chunk across four task types: question answering~(QA), summarization, quoting, and paraphrasing. Each chunk yields approximately ten pairs (five QA, one summarization, two quoting, two paraphrasing). To improve coverage and diversity, the model first enumerates key facts in a \emph{scratchpad} before producing the final pairs. All instructions are written to be self-contained: they reference specific entities or events rather than ``the passage,'' so that a reader with access to the full document can locate the relevant information. Full prompt templates are provided in Appendix~\ref{app:prompts}.

\paragraph{Verification and filtering.}
A separate long-context LLM (\texttt{Qwen2.5-7B-Instruct-1M}; \citealp{qwen2.5-1m}) re-answers each instruction using the \emph{entire} document rather than the originating chunk. A consistency check compares the chunk-based and full-context responses with task-specific criteria (e.g.\ factual match for QA, semantic equivalence for paraphrasing). Pairs whose consistency confidence falls below $0.7$ are discarded.

\paragraph{Dataset statistics.}
After filtering, each dataset contains approximately 100\,000 instruction--response pairs. We use a random 80/20 train/test split. During caching, each pair is concatenated with its source document to form sequences of up to 122\,500 context tokens (our largest context) and up to 512 query tokens.

\subsection{Caching Attention Intermediates}
\label{app:caching}

To train \RegAtt{} we cache ground-truth attention targets from the base language model. For every (context, instruction, response) sequence we run a full forward pass and, at each layer $\ell\in\{1,\dots,L\}$, extract two quantities for every query position~$i$ (non-context tokens): (1) the log-normalizer $\alpha^{\ell,h}(\bq)$ and (2) the attention output $\mathcal{A}^{\ell,h}(\bq)$, both computed via \texttt{dot\_product\_attention} with residuals. Targets are stored as \texttt{float32} arrays alongside the tokenized dataset.

\subsection{\RegAtt{} Architecture Details}
\label{app:arch}

\paragraph{Shape of the MLP family.}
For the MLP variant of \S\ref{sec:method:models}, each layer~$\ell$ and KV-head~$h$ carries a module with three parts:
\begin{itemize}
    \item \textbf{Backbone} (optional): an MLP of depth $D_b$ and width $h$ mapping $\bq\!\in\!\R^{d}$ to $\mathbf{h}\!\in\!\R^{h}$. When $D_b{=}0$, $\mathbf{h}=\bq$.
    \item \textbf{Score head}: an MLP of depth $D_s$ mapping $\mathbf{h}$ to $a^{\ell,h}_\theta\!\in\!\R$.
    \item \textbf{Target head}: an MLP of depth $D_t$ mapping $\mathbf{h}$ to $A^{\ell,h}_\theta\!\in\!\R^{d}$.
\end{itemize}
All MLPs use SiLU activations. Skip connections re-inject $\bq$ at intermediate depths, and residual blocks plus layer normalization can be enabled per configuration.

\paragraph{Integration with standard attention.}
The \RegAtt{} output is inserted into the standard attention arrays by prepending $a^{\ell,h}_\theta$ to the $n$-dimensional logit vector and $A^{\ell,h}_\theta$ to the value matrix. A single softmax over the extended $(n{+}1)$-dimensional logit vector produces the blended output, so no change to the base-model attention kernel is required.

\paragraph{Grouped-query \RegAtt{}.}
Under grouped-query attention \citep[GQA;][]{ainslie2023gqa}, all query heads in the same KV group share a single \RegAtt{} pair, matching the structure of the attention operation (see \S\ref{sec:method}). This reduces the per-layer module count from $H$ query heads to the number of KV heads.

\subsection{Hyperparameters}
\label{app:hyperparameters}

\paragraph{Optimizer and schedule.}
All \RegAtt{} modules are trained with Adam, NaN-gradient masking, and global-norm gradient clipping at $1.0$. The learning rate follows a warmup--cosine--decay schedule with initial and end values set to $0$ and a peak value of $1\!\times\!10^{-4}$. The peak is rescaled per run by $\sqrt{b/b_{\text{ref}}}$, where $b$ is the effective per-device batch size and $b_{\text{ref}}$ the reference batch size (Table~\ref{tab:hparam-bands}), to compensate for batch-size-dependent gradient noise. Warmup is set to $\min(N/40,\;10\,N_{\text{samples}}/b)$ iterations, i.e.\ the smaller of $2.5\%$ of the training budget or ten epochs of the dataset. Only \RegAtt{} parameters are updated; the base model is frozen. The total training budget is fixed at $10^{6}$ datapoints across all runs.

\paragraph{Loss weights.}
Runs sweep three combinations of $(\lambda_\alpha,\lambda_{\mathcal{A}},\lambda_{\text{KL}})$: pure regression $(0.1,1.0,0)$, regression plus distillation $(0.1,1.0,2.0)$, and distillation only $(0,0,1.0)$. For quadrature architectures $\lambda_\alpha$ is set to $0$ at runtime, as the score is computed analytically from $\mathbf{W}$.

\paragraph{Batch size and sequence length.}
The effective per-device batch size is scaled inversely with context length and $\sqrt{\rho}$, and clipped to $[1, 32]$. When either distillation or cross-entropy is active, batch size is further halved to accommodate the additional logits dataloader. Query sequence length (instruction $+$ response tokens) is capped at $196$ for training; longer samples are filtered and shorter ones padded.

\paragraph{Capacity allocation.}
Per-layer $\rho$ is allocated with group multipliers $(1, 2, 5, 2)$ on four contiguous layer groups, normalized to preserve the average $\rho$ (see \Cref{app:layer-groups}).

\paragraph{Representative configurations.}
Table~\ref{tab:hparam-bands} summarizes the band of effective batch sizes and iteration counts induced by the scaling rules above, across the $\rho\!\in\!\{0.5\%, 2\%\}$ and dataset (context length $\in [20\text{k}, 122\text{k}]$) sweep axes. Lower values in each band correspond to longer-context datasets (\texttt{ip\_intro}) or larger $\rho$; upper values correspond to shorter-context datasets (\texttt{great\_gatsby}) or smaller $\rho$.

\begin{table}[t]
\centering
\caption{Per-model reference batch size and effective-hyperparameter bands across the sweep. Reference batch size is calibrated at ctx\_len${\approx}62$k, $\rho{=}5\%$, and query length $128$; the effective batch size in each run follows the inverse-context, inverse-$\sqrt{\rho}$, and distillation-halving rules. Iteration counts assume a $10^{6}$ datapoint budget on 8 devices.}
\label{tab:hparam-bands}
\small
\begin{tabular}{@{}l c c c c@{}}
\toprule
Model & Ref.\ $b_{\text{ref}}$ & Eff.\ per-device bs & Iterations (K) & Hardware \\
\midrule
\texttt{Qwen3-1.7B} & $4$  & $2$--$16$ & $8$--$63$ & H100 ($8\times$) \\
\texttt{Qwen3-4B}   & $2$  & $1$--$8$  & $16$--$125$ & B200 ($8\times$) \\
\texttt{Qwen3-8B}   & $1$  & $1$--$4$  & $31$--$125$ & B200 ($8\times$) \\
\bottomrule
\end{tabular}
\end{table}

\subsection{Computational Resources}
\label{app:compute}

\paragraph{Hardware.}
Training is performed on single 8-GPU nodes with data parallelism. \texttt{Qwen3-1.7B} runs on NVIDIA H100 nodes; \texttt{Qwen3-4B} and \texttt{Qwen3-8B} run on NVIDIA B200 nodes, whose larger HBM fits the base-model forward pass together with the cached attention targets and logits at context lengths up to ${\sim}122$k tokens.

\paragraph{Precision and kernels.}
Runs use bfloat16 compute with float32 parameters and cached targets. The base-model forward pass uses Flash Attention 3; the smallest quadrature variant (at $\rho{=}0.5\%$) uses a standard softmax implementation. XLA runs with Triton GEMMs disabled.

\subsection{Wall-clock Training Time}
\label{app:wallclock}

We report the average per-run wall-clock time observed in our sweeps for \texttt{Qwen3-1.7B}, \texttt{Qwen3-4B}, and \texttt{Qwen2.5-7B-1M}, grouped by loss configuration. Values are averaged across datasets and \RegAtt{} architectures (MLP and Quadrature), restricted to runs sharing a common total-sample budget (reported in the Samples column) so that timings are comparable. Each reported time subtracts one hour to remove the dataloading and compilation overhead that is common to all runs regardless of configuration; the reported value therefore approximates optimizer time on that training budget. Hardware is fixed per model, as listed in \S\ref{app:compute}.

\begin{table}[h]
\centering
\caption{Average wall-clock training time (hours, $-1$h dataloading) on a single 8-GPU node, at a fixed total-sample budget per row. Pure regression: $\lambda_\alpha{=}0.1,\lambda_\mathcal{A}{=}1,\lambda_{\text{KL}}{=}0$. Mixed: $\lambda_\alpha{=}0.1,\lambda_\mathcal{A}{=}1,\lambda_{\text{KL}}{>}0$. Pure distillation: $\lambda_\alpha{=}\lambda_\mathcal{A}{=}0,\lambda_{\text{KL}}{>}0$.}
\label{tab:wallclock}
\small
\begin{tabular}{@{}l c c c c c@{}}
\toprule
Model & $\rho$ & Samples & Pure reg. (h) & Mixed (h) & Pure dist. (h) \\
\midrule
\texttt{Qwen3-1.7B} & $2\%$  & $1\text{M}$   & $13.2$ & $12.9$ & $11.8$ \\
\texttt{Qwen3-4B}   & $2\%$  & $1\text{M}$   & $26.2$ & $25.0$ & $24.8$ \\
\texttt{Qwen2.5-7B-1M}   & $2\%$  & $1\text{M}$   & $16.4$ & $16.3$ & $16.5$ \\
\texttt{Qwen2.5-7B-1M}   & $5\%$  & $1.5\text{M}$ & $23.7$ & $23.8$ & $24.0$ \\
\texttt{Qwen2.5-7B-1M}   & $10\%$ & $1.5\text{M}$ & $23.7$ & $24.0$ & $23.8$ \\
\bottomrule
\end{tabular}
\end{table}

The three loss configurations are within ${\sim}5\%$ of each other across all models and parameter fractions: the extra cost of loading attention targets (for regression) roughly matches the extra cost of loading teacher logits (for distillation), and the batch-size halving triggered when either is active (\S\ref{app:hyperparameters}) further equalizes per-step cost. Wall-clock scales with base-model size and, for \texttt{Qwen2.5-7B-1M}, with the larger $1.5$M-sample budget used at $\rho\!\in\!\{5\%,10\%\}$.

\subsection{Memory Footprint}
\label{app:memory}

We report the in-memory size of the base model, its full KV-cache at a representative context length ($n{=}62\text{k}$ tokens), and the size of a \RegAtt{} module at each parameter fraction. All sizes assume bfloat16 storage (2~bytes per parameter). The \RegAtt{} module size is computed as $\rho \cdot 2nd \cdot L \cdot H_{\text{kv}}$ parameters, matching the definition of $\rho$ in \S\ref{sec:exp:setup}.

\begin{table}[h]
\centering
\caption{Base-model size, full KV-cache footprint at $n{=}62$k, and \RegAtt{} module size at $\rho\!\in\!\{2,5,10\}\%$, in MB (bfloat16).}
\label{tab:memory}
\small
\begin{tabular}{@{}l r r r r r@{}}
\toprule
Model & Base & KV-cache & \multicolumn{3}{c}{\RegAtt{} (MB)} \\
\cmidrule(lr){4-6}
      & (GB) & $n{=}62$k (GB) & $\rho{=}2\%$ & $\rho{=}5\%$ & $\rho{=}10\%$ \\
\midrule
\texttt{Qwen3-1.7B} & $3.44$  & $7.11$  & $142$ & $356$  & $711$  \\
\texttt{Qwen3-4B}   & $8.04$  & $9.14$  & $183$ & $457$  & $914$  \\
\texttt{Qwen2.5-7B-1M}   & $14.14$ & $3.56$  & $71$  & $178$  & $355$  \\
\texttt{Qwen3-8B}   & $16.38$ & $16.25$ & $325$ & $813$  & $1625$ \\
\bottomrule
\end{tabular}
\end{table}

The KV-cache-to-base-model ratio varies with architecture: \texttt{Qwen3-1.7B} and \texttt{Qwen3-8B} have KV-caches of comparable size to the model weights themselves at this context length, while \texttt{Qwen2.5-7B-1M} (which uses a higher GQA factor) has a much smaller cache. In all cases, \RegAtt{} at $\rho{=}2\%$ reduces the long-context footprint by one to two orders of magnitude relative to the full cache; at $\rho{=}10\%$ the footprint remains well below the KV-cache and the base model.

\section{Proofs and Derivations}
\label{app:proofs}

\subsection{Properties of Log-Sum-Exp}
\label{app:lse}

The log-sum-exp function has several important properties that inform our approach. $\text{LSE}(\bq) = \log \sum_{j=1}^n \exp(\ip{\bq}{\bk_j})$ is convex in $\bq$ as the composition of a linear map with the standard log-sum-exp, and its gradient is the softmax-weighted average of keys,
\begin{equation}
    \nabla_{\bq} \mathrm{LSE}(\bq) = \sum_{j=1}^n \frac{\exp(\ip{\bq}{\bk_j})}{\sum_{j'} \exp(\ip{\bq}{\bk_{j'}})} \bk_j = \mathbf{K}^T\,\mathrm{softmax}(\mathbf{K}\bq)\,.
\end{equation}

\subsection{Connection to Support Functions and Amortized MIPS}
\label{app:support}

The score network $a^{\ell,h}_\theta(\bq)$ regresses the log-sum-exp $\alpha^{\ell,h}(\bq) = \log\sum_j \exp(\ip{\bq}{\bk_j})$. In the low-temperature limit,
\begin{equation}
    \lim_{\beta \to \infty} \frac{1}{\beta} \log \sum_{j=1}^n \exp(\beta \ip{\bq}{\bk_j}) = \max_j \ip{\bq}{\bk_j}\,,
\end{equation}
so LSE is a smoothed version of the maximum inner product $\max_j \ip{\bq}{\bk_j}$, i.e.\ the support function of the key set. Computing this maximum for a query $\bq$ against a fixed key set is precisely the problem solved by \emph{maximum inner product search} (MIPS), which is the dominant cost of greedy decoding over large fixed key/value stores. \citet{amips2025} amortize MIPS by training a neural network that predicts the support function $\bq \mapsto \max_j \ip{\bq}{\bk_j}$ from queries sampled from a task distribution. Our score head targets the finite-temperature analogue of the same quantity, and we reuse the MLP design proposed there.

\section{Additional Results}
\label{app:results}

\subsection{Non-Uniform Capacity Allocation Across Layers}
\label{app:layer-groups}

\paragraph{Setting.}
We partition the 28 layers of \texttt{Qwen3-1.7B} into four contiguous groups, $[0,9)$, $[9,15)$, $[15,27)$, and $[27,28)$, and assign per-group multipliers, normalized so that the average $\rho$ matches across settings. The \emph{weighted} scheme uses multipliers $(1, 2, 5, 2)$ on these four groups (larger capacity in middle-to-late layers); the \emph{uniform} scheme uses $(1, 1, 1, 1)$. We compare MLP-based \RegAtt{} modules with $\rho{\approx}2\%$ and $\lambda_{\text{KL}}{=}0.01$, evaluated at convergence. \Cref{tab:layer-groups} reports raw score and target MSE and the token-accuracy gap.

\begin{table}[t]
\centering
\caption{Weighted vs.\ uniform layer-group allocation at convergence (MLP, $\rho{\approx}2\%$, $\lambda_{\text{KL}}{=}0.01$). Positive $\Delta$ means uniform is worse (higher error). Weighted allocation reduces target MSE and token-accuracy gap at the expense of slightly higher score MSE.}
\label{tab:layer-groups}
\small
\begin{tabular}{@{}l cc c cc c cc c@{}}
\toprule
& \multicolumn{3}{c}{Score MSE} & \multicolumn{3}{c}{Target MSE} & \multicolumn{3}{c}{Token Acc.\ Gap (\%)} \\
\cmidrule(lr){2-4} \cmidrule(lr){5-7} \cmidrule(lr){8-10}
Dataset & W & U & $\Delta$ & W & U & $\Delta$ & W & U & $\Delta$ \\
\midrule
Heart of Dark.   & $0.063$ & $0.049$ & $-0.015$ & $0.224$ & $0.251$ & $+0.027$ & $0.95$ & $1.03$ & $+0.09$ \\
LongHealth       & $0.070$ & $0.048$ & $-0.022$ & $0.229$ & $0.270$ & $+0.041$ & $0.45$ & $0.57$ & $+0.12$ \\
Time Machine     & $0.089$ & $0.072$ & $-0.017$ & $0.218$ & $0.246$ & $+0.028$ & $0.41$ & $0.67$ & $+0.26$ \\
\midrule
\textbf{Average} & $0.074$ & $0.056$ & $\mathbf{-0.018}$ & $0.224$ & $0.256$ & $\mathbf{+0.032}$ & $0.60$ & $0.76$ & $\mathbf{+0.16}$ \\
\bottomrule
\end{tabular}
\end{table}

\paragraph{Observations.}
Weighted allocation has lower target MSE (by ${\sim}3$--$4\!\times\!10^{-2}$) and a smaller token-accuracy gap than uniform allocation on all three datasets in the table. Uniform allocation has lower score MSE, consistent with its larger share of capacity on early layers where scores are harder to fit.

\subsection{Score vs.\ Target Parameter Allocation}
\label{app:score-target-alloc}

\paragraph{Setting.}
The MLP architecture allocates separate parameter budgets to the score and target heads. At a total $\rho{=}10\%$, we compare two splits on \texttt{Qwen2-7B}: (i) score $1\%$ / target $9\%$, and (ii) score $2.5\%$ / target $7.5\%$. Both use the same layer-group multipliers $(1,2,8,12)$. We run each split under two training regimes (see \S\ref{app:hyperparameters}): \emph{pure distillation} ($\lambda_\alpha{=}\lambda_{\mathcal{A}}{=}0,\lambda_{\text{KL}}{=}1$), where neither head is directly supervised, and \emph{mixed training} ($\lambda_\alpha{=}0.1,\lambda_{\mathcal{A}}{=}1,\lambda_{\text{KL}}{=}2$), where regression supervises each head in addition to the KL term. \Cref{tab:score-target-alloc} reports the token-accuracy gap at convergence.

\begin{table}[h]
\centering
\caption{Score/target parameter split ablation at $\rho{=}10\%$ (\texttt{Qwen2-7B}). Reported values are token-accuracy gaps (lower is better). Mixed training ($\lambda_{\text{KL}}{=}2$) is the regime in which the split directly affects how regression capacity is allocated across heads; pure distillation ($\lambda_{\text{KL}}{=}1$) leaves the heads unsupervised except through the KL term.}
\label{tab:score-target-alloc}
\small
\begin{tabular}{@{}l cc cc@{}}
\toprule
& \multicolumn{2}{c}{Pure distillation ($\lambda_{\text{KL}}{=}1$)} & \multicolumn{2}{c}{Mixed training ($\lambda_{\text{KL}}{=}2$)} \\
\cmidrule(lr){2-3} \cmidrule(lr){4-5}
Dataset & $1/9$ & $2.5/7.5$ & $1/9$ & $2.5/7.5$ \\
\midrule
Heart of Dark.   & $0.42$ & $0.22$ & $0.32$ & $0.47$ \\
LongHealth       & $0.55$ & $0.45$ & $0.51$ & $0.50$ \\
Great Gatsby     & $0.03$ & $0.16$ & $0.41$ & $0.63$ \\
IP Intro         & $-0.04$ & $-0.05$ & $0.10$ & $0.26$ \\
\midrule
\textbf{Average} & $0.24$ & $0.20$ & $0.34$ & $0.47$ \\
\bottomrule
\end{tabular}
\end{table}

\paragraph{Observations.}
The mixed-training regime is the more informative of the two, since it is the setting in which the split directly controls how much regression supervision each head receives. There, the $1/9$ split attains an average token-accuracy gap of $0.34\%$ against $0.47\%$ for the $2.5/7.5$ split, indicating that allocating most of the MLP budget to the target head is preferable when the score head is already well-constrained by the score regression term. Under pure distillation the two splits are within noise of each other ($0.24\%$ vs.\ $0.20\%$), as expected: with $\lambda_\alpha{=}\lambda_{\mathcal{A}}{=}0$ the split only redistributes unsupervised capacity.

\subsection{Training Dynamics}
\label{app:training-dynamics}

\begin{figure}[h]
    \centering
    \includegraphics[width=\linewidth]{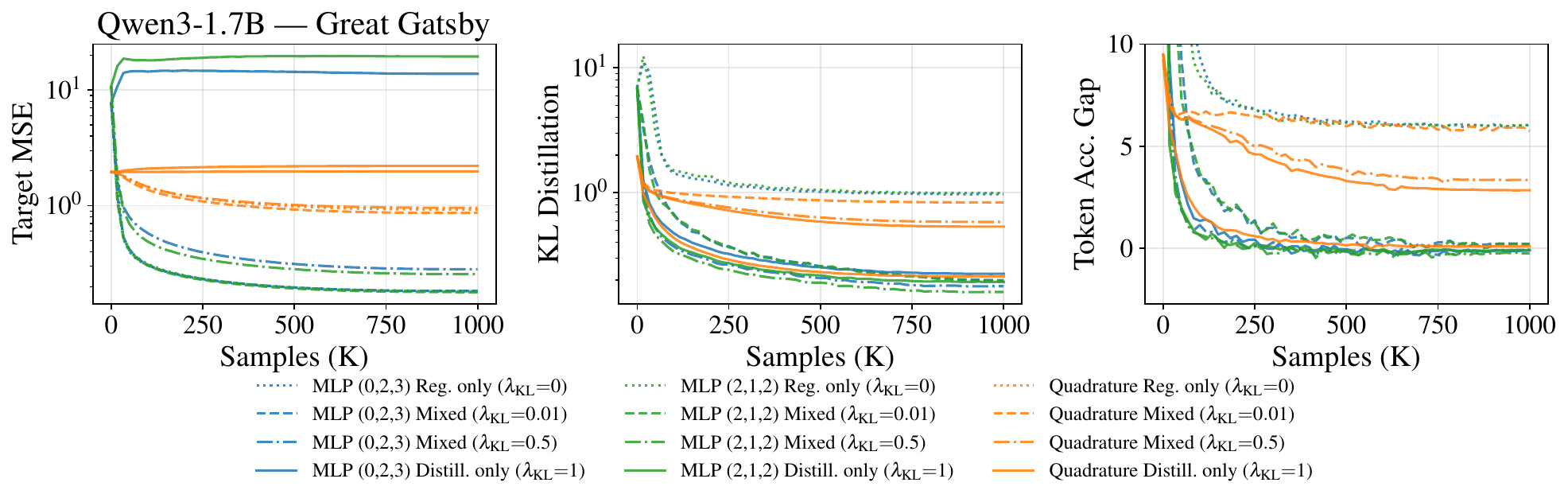}\\[4pt]
    \includegraphics[width=\linewidth]{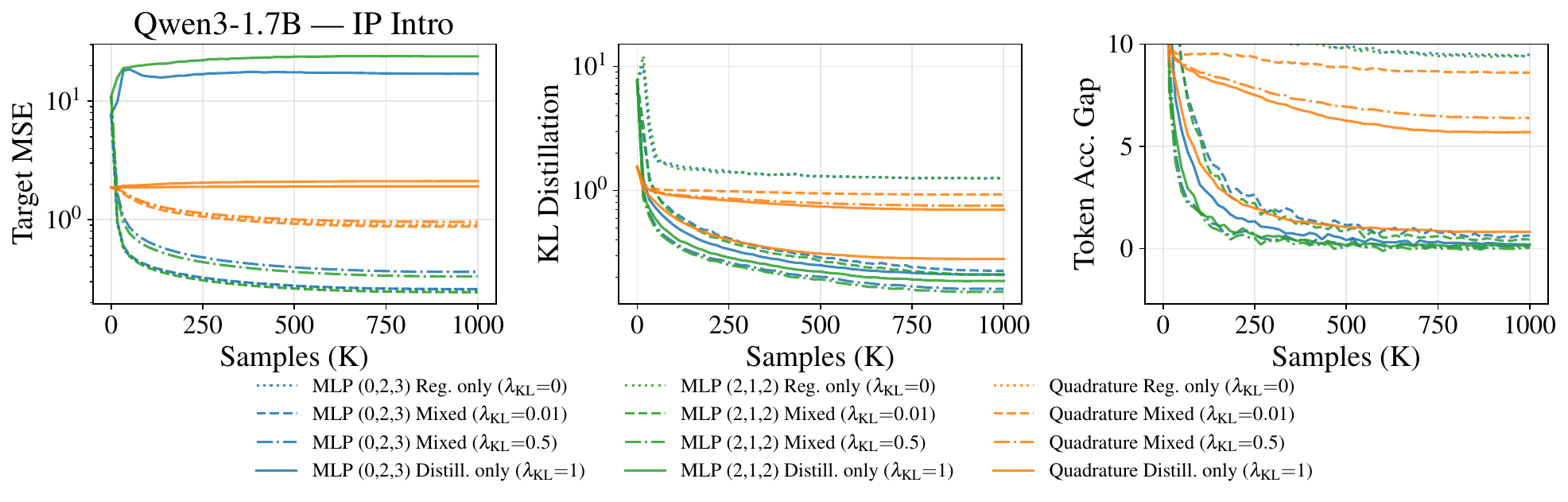}
    \caption{Training dynamics for \texttt{Qwen3-1.7B} at $\rho{=}2\%$ on Great Gatsby (top) and IP Intro (bottom).}
    \label{fig:dyn-1.7B}
\end{figure}

\begin{figure}[h]
    \centering
    \includegraphics[width=\linewidth]{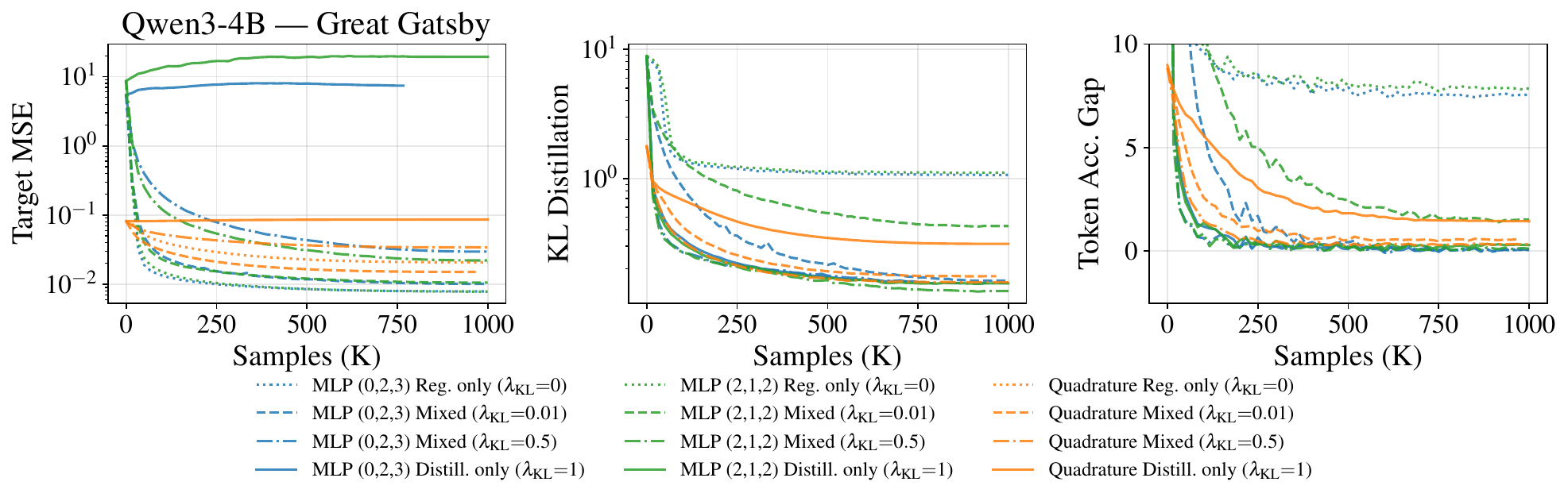}\\[4pt]
    \includegraphics[width=\linewidth]{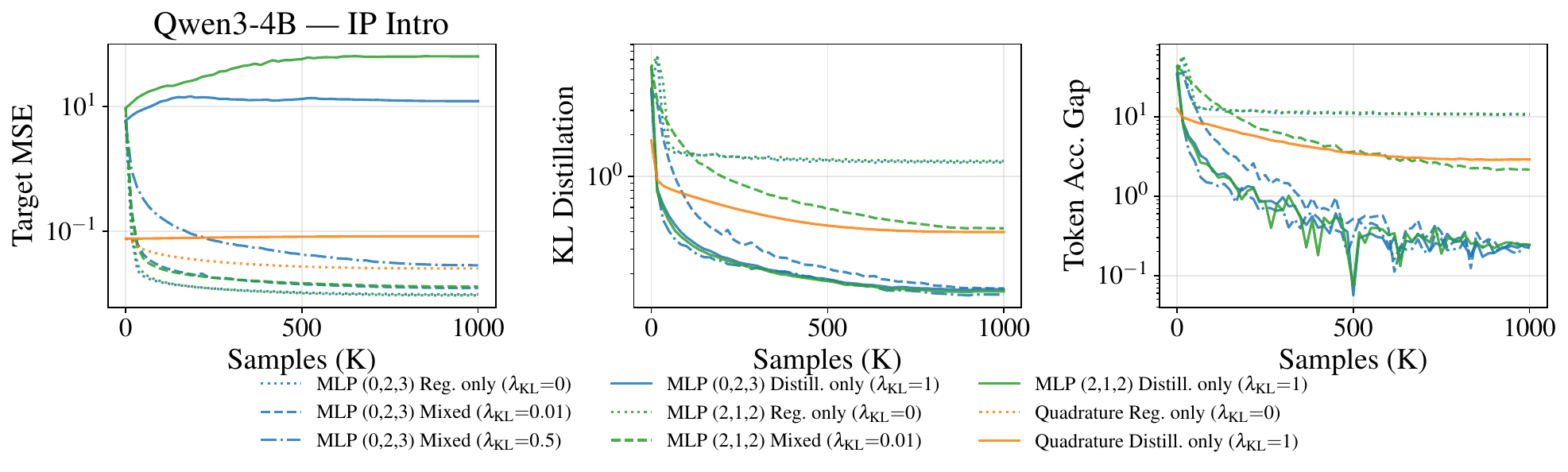}
    \caption{Training dynamics for \texttt{Qwen3-4B} at $\rho{=}2\%$ on Great Gatsby (top) and IP Intro (bottom).}
    \label{fig:dyn-4B}
\end{figure}

\begin{figure}[h]
    \centering
    \includegraphics[width=\linewidth]{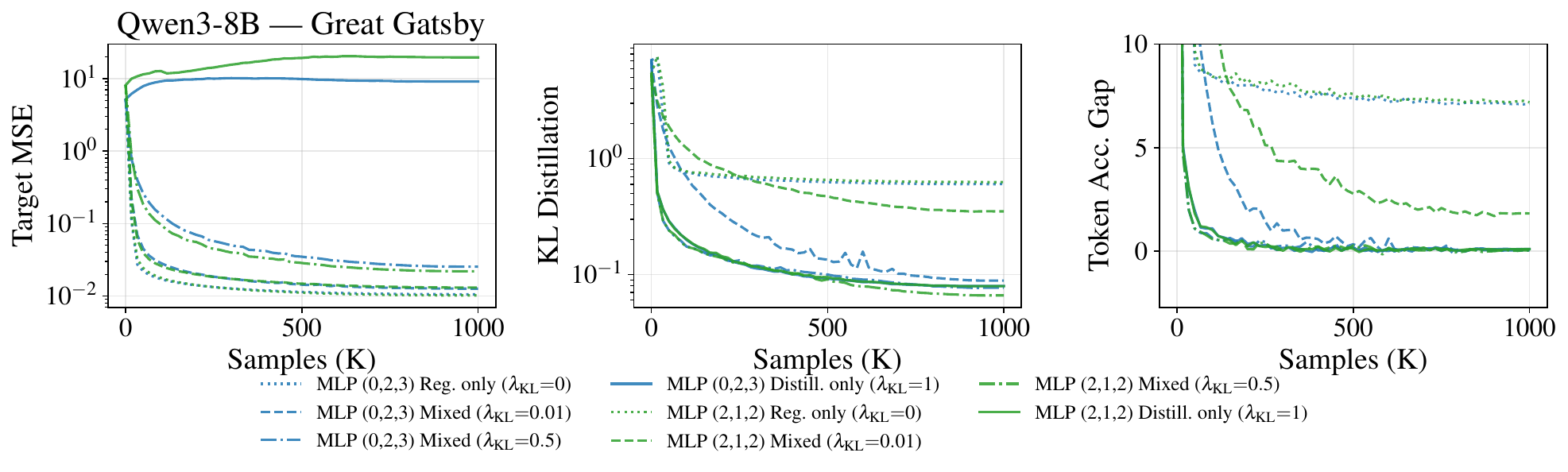}\\[4pt]
    \includegraphics[width=\linewidth]{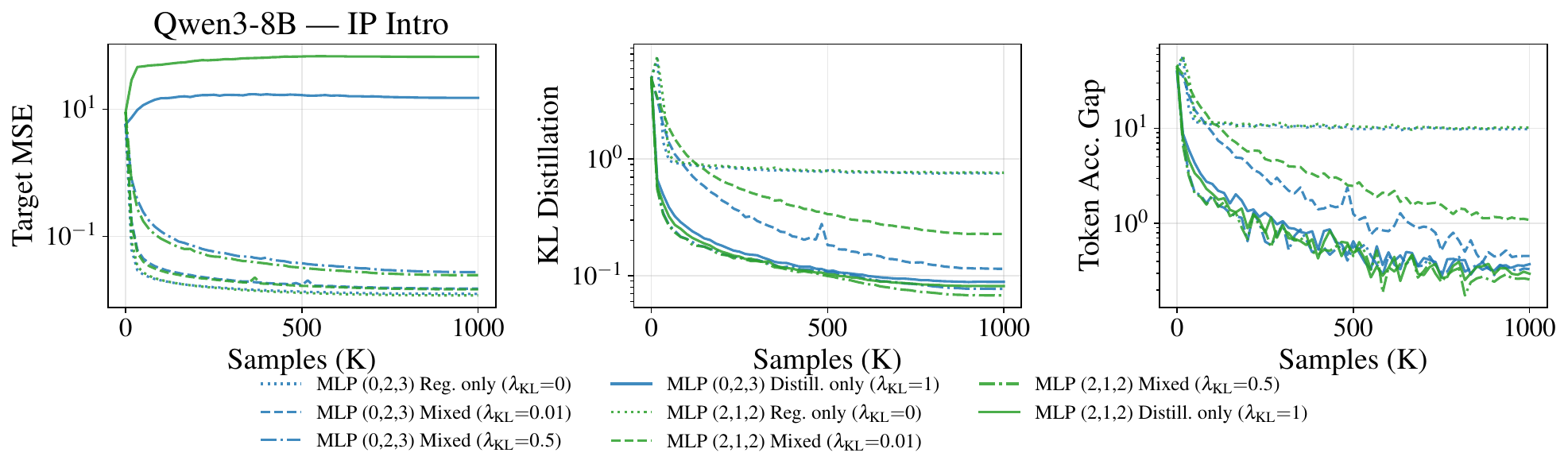}
    \caption{Training dynamics for \texttt{Qwen3-8B} at $\rho{=}2\%$ on Great Gatsby (top) and IP Intro (bottom).}
    \label{fig:dyn-8B}
\end{figure}

\begin{figure}[h]
    \centering
    \includegraphics[width=\linewidth]{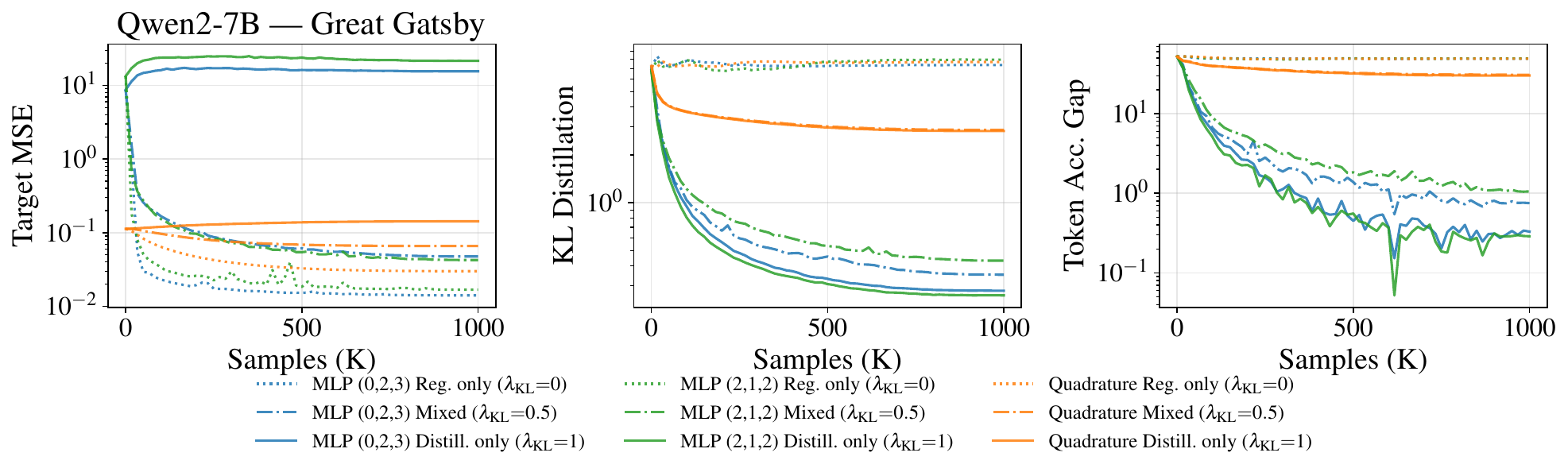}\\[4pt]
    \includegraphics[width=\linewidth]{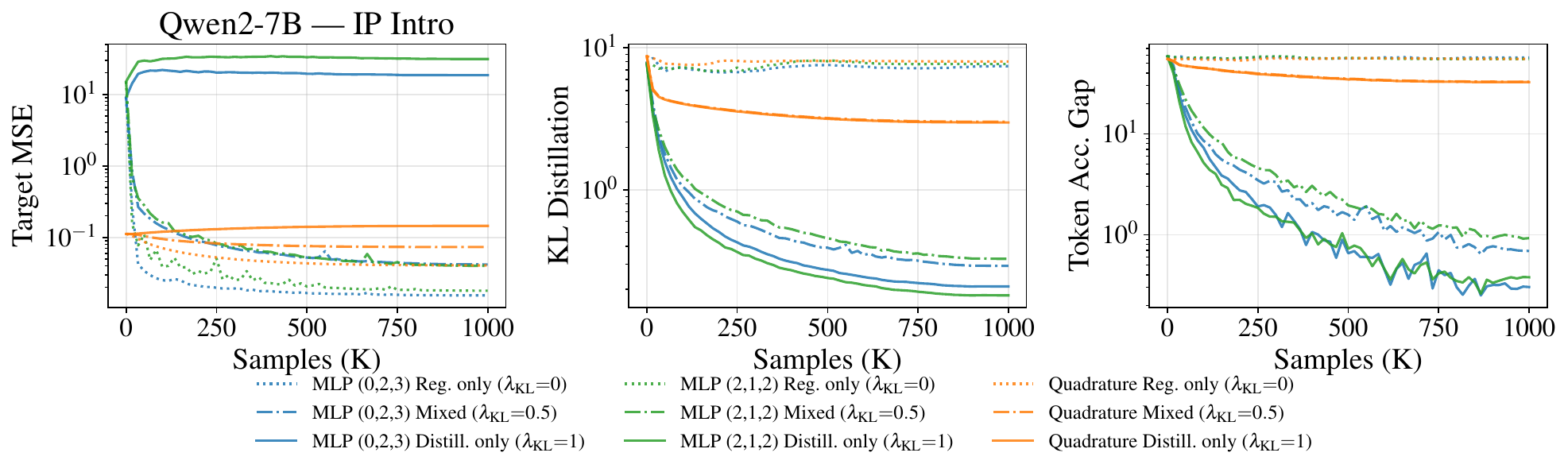}
    \caption{Training dynamics for \texttt{Qwen2-7B} at $\rho{=}2\%$ on Great Gatsby (top) and IP Intro (bottom).}
    \label{fig:dyn-7B}
\end{figure}

\Cref{fig:dyn-1.7B,fig:dyn-4B,fig:dyn-8B,fig:dyn-7B} show representative training curves for \RegAtt{} modules at $\rho{=}2\%$. Each panel plots an eval metric (target MSE, KL distillation loss, token-accuracy gap) against the number of training samples, where one sample is a single question--answer instruction pair (typically fewer than $196$ tokens). Curves are shown for MLP and Quadrature architectures at three distillation weights: $\lambda_{\text{KL}}\!\in\!\{0, 0.01, 1\}$.

Across all models and datasets, target MSE decreases monotonically and converges within ${\sim}500$K samples for MLP and slightly faster for Quadrature. The KL distillation loss follows a similar trajectory when distillation is active ($\lambda_{\text{KL}}{>}0$), but remains flat at high values for pure-regression runs ($\lambda_{\text{KL}}{=}0$). The token-accuracy gap shows the sharpest early improvement, with most gains realized in the first $200$K samples; further training yields diminishing returns. Runs with $\lambda_{\text{KL}}{=}1$ (pure distillation) converge fastest in token-accuracy gap despite having higher target MSE, consistent with the observation that distillation optimizes directly for generation quality.

\subsection{Additional Scatter Plots}
\label{app:scatter}

\begin{figure}[h]
    \centering
    \raisebox{.7cm}{\rotatebox{90}{\fontsize{6}{7}\selectfont Token Acc.\ Gap (\%)}}\,%
    \includegraphics[width=0.23\linewidth]{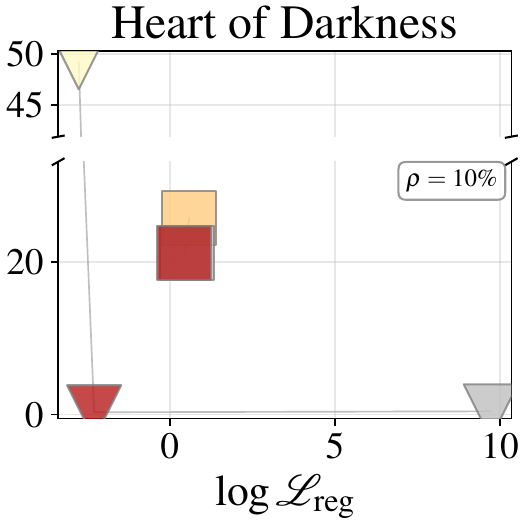}
    \includegraphics[width=0.23\linewidth]{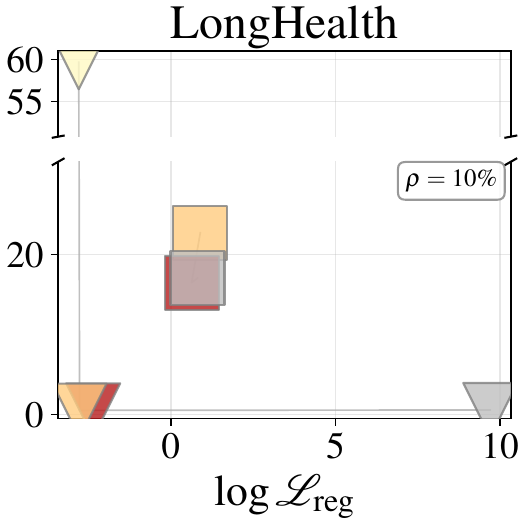}
    \includegraphics[width=0.23\linewidth]{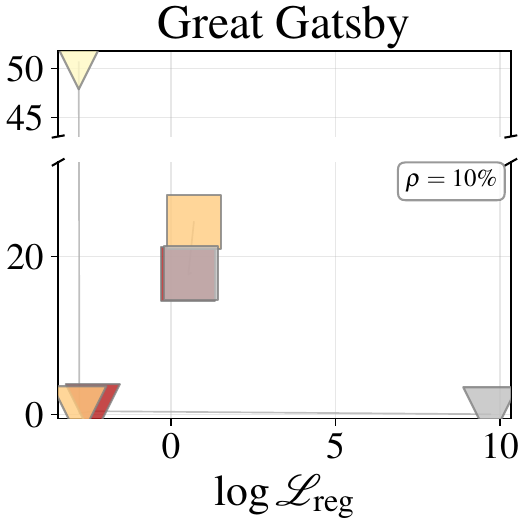}
    \includegraphics[width=0.23\linewidth]{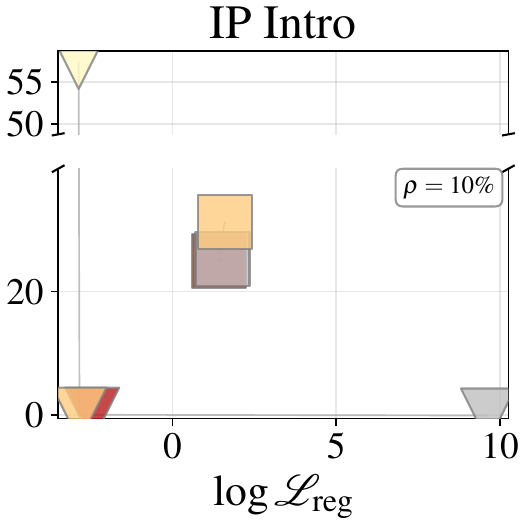}
    \\\vskip.1cm
    \includegraphics[width=\linewidth]{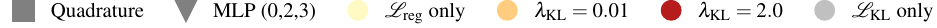}
    \caption{Token-accuracy gap (\%) vs.\ $\log \mathcal{L}_{\text{reg}}$ for \texttt{Qwen2.5-7B-1M} at $\rho{=}10\%$.}
    \label{fig:rl-7B-pf10}
\end{figure}

\begin{figure}[h]
    \centering
    \raisebox{.7cm}{\rotatebox{90}{\fontsize{6}{7}\selectfont Token Acc.\ Gap (\%)}}\,%
    \includegraphics[width=0.23\linewidth]{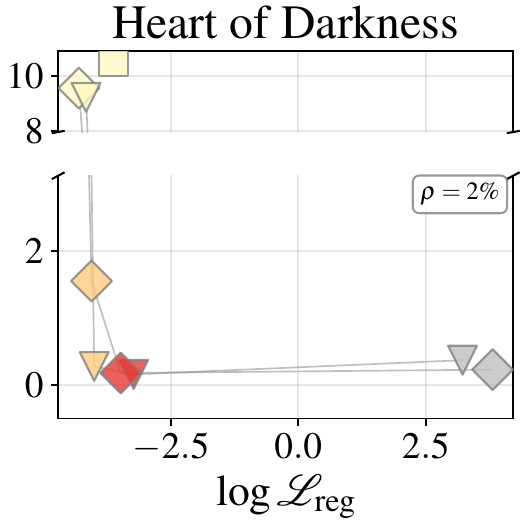}
    \includegraphics[width=0.23\linewidth]{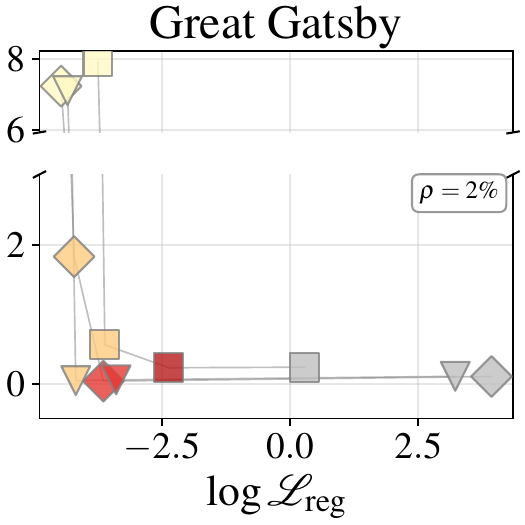}
    \includegraphics[width=0.23\linewidth]{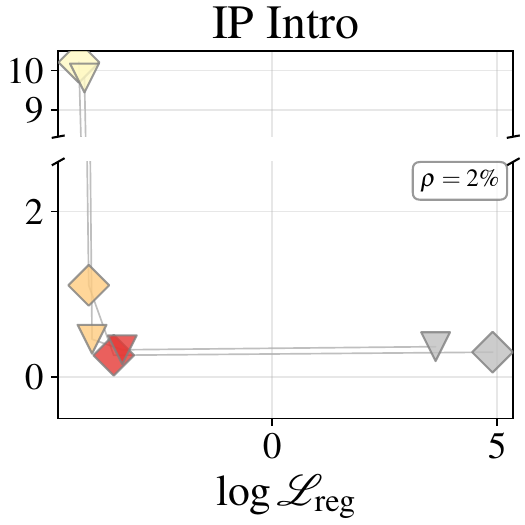}
    \includegraphics[width=0.23\linewidth]{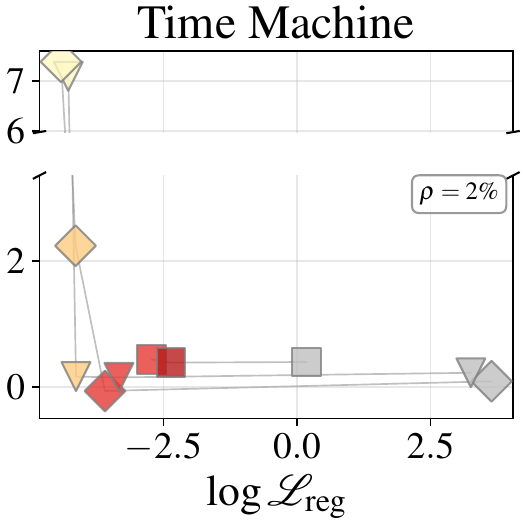}
    \\\vskip.1cm
    \includegraphics[width=\linewidth]{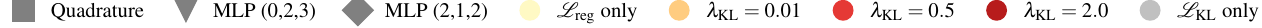}
    \caption{Token-accuracy gap (\%) vs.\ $\log \mathcal{L}_{\text{reg}}$ for \texttt{Qwen3-8B} (with YaRN $4{\times}$) at $\rho{=}2\%$.}
    \label{fig:rl-8B}
\end{figure}

\paragraph{\texttt{Qwen2.5-7B-1M} at $\rho{=}10\%$ and \texttt{Qwen3-8B}.}
\Cref{fig:rl-7B-pf10} extends the main-text scatter plots to $\rho{=}10\%$ on \texttt{Qwen2.5-7B-1M}. \Cref{fig:rl-8B} shows the same token-accuracy gap vs.\ $\log\mathcal{L}_{\text{reg}}$ relationship on \texttt{Qwen3-8B} at $\rho{=}2\%$.

\paragraph{Relative transport error.}
\label{app:rte-def}
Several of the plots below use the \emph{relative transport error} (RTE) as a training-time diagnostic for the regression heads, summed over layers and heads:
\begin{equation}
    \mathcal{E}_{\text{rel}}
    \;=\;
    \sum_{\ell, h}
    \mathbb{E}_{\bq}\!\left[\log \frac{\bigl\lVert A^{\ell,h}_\theta(\bq) - \mathcal{A}^{\ell,h}(\bq) \bigr\rVert_2^2}{\bigl\lVert \bq - \mathcal{A}^{\ell,h}(\bq) \bigr\rVert_2^2}\right],
    \label{eq:rte}
\end{equation}
and analogously for the score head, replacing the squared norm with a squared difference: $\bigl(a^{\ell,h}_\theta(\bq) - \alpha^{\ell,h}(\bq)\bigr)^2 / \bigl(\alpha^{\ell,h}(\bq)\bigr)^2$. The quantity $\exp(\mathcal{E}_{\text{rel}})$ is the geometric mean of the per-query squared-error ratio; for instance, $\mathcal{E}_{\text{rel}} = -4$ corresponds to a geometric-mean ratio of $e^{-4} \approx 0.018$, meaning the prediction error is roughly fifty times smaller than the query-to-target baseline distance.

\begin{figure}[h]
    \centering
    \raisebox{.7cm}{\rotatebox{90}{\fontsize{6}{7}\selectfont Token Acc.\ Gap (\%)}}\,%
    \includegraphics[width=0.23\linewidth]{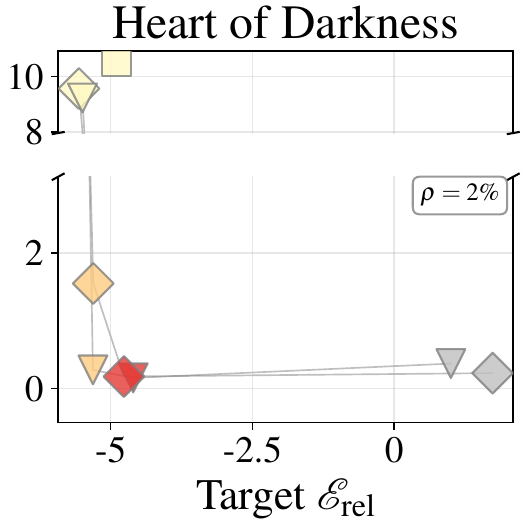}
    \includegraphics[width=0.23\linewidth]{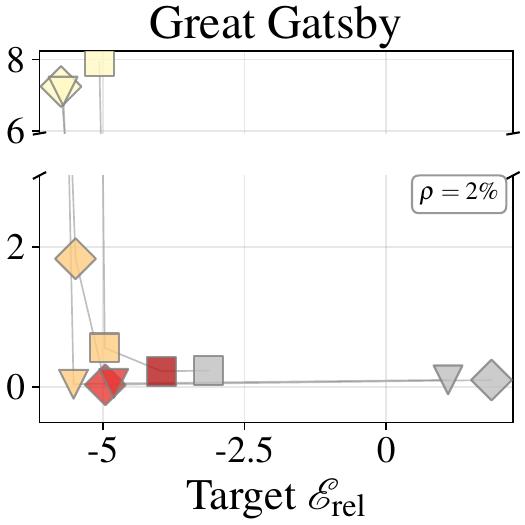}
    \includegraphics[width=0.23\linewidth]{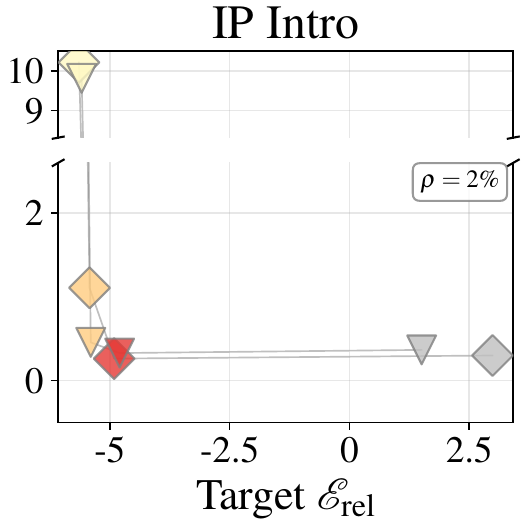}
    \includegraphics[width=0.23\linewidth]{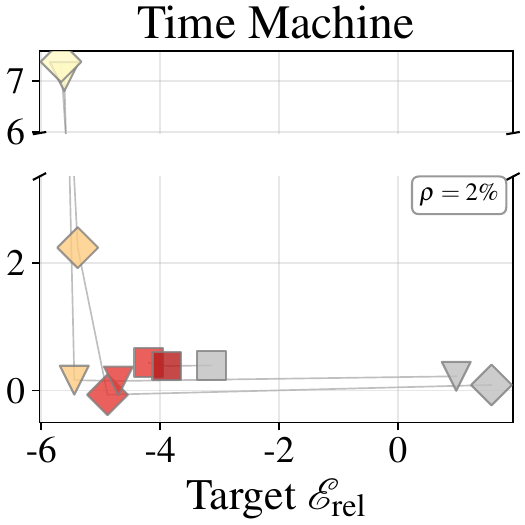}
    \\\vskip.1cm
    \includegraphics[width=\linewidth]{figures/scatter/komainu_pf2_legend.pdf}
    \caption{Token-accuracy gap (\%) vs.\ target $\mathcal{E}_{\text{rel}}$ for \texttt{Qwen3-8B} at $\rho{=}2\%$.}
    \label{fig:ta-8B}
\end{figure}

\paragraph{Token-accuracy gap vs.\ target $\mathcal{E}_{\text{rel}}$.}
\label{app:ta-vs-rte}
\Cref{fig:ta-8B} plots the token-accuracy gap against target $\mathcal{E}_{\text{rel}}$ for \texttt{Qwen3-8B}.

\begin{figure}[h]
    \centering
    \raisebox{.7cm}{\rotatebox{90}{\fontsize{6}{7}\selectfont Token Acc.\ Gap (\%)}}\,%
    \includegraphics[width=0.32\linewidth]{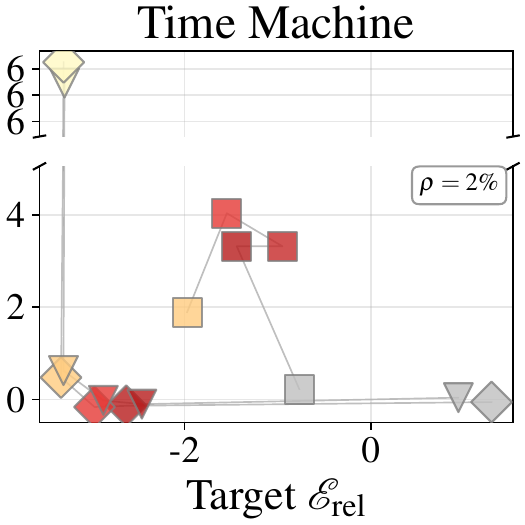}
    \includegraphics[width=0.32\linewidth]{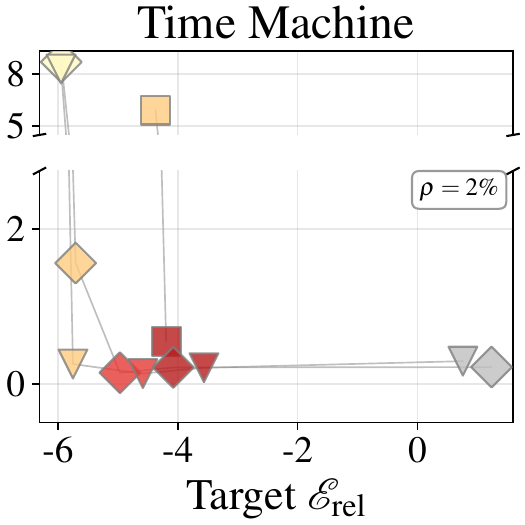}
    \includegraphics[width=0.32\linewidth]{figures/scatter/komainu_pf2_ta_qwen3_8B_time_machine.pdf}
    \\\vskip.1cm
    \includegraphics[width=\linewidth]{figures/scatter/kappa_pf2_legend.pdf}
    \caption{Token-accuracy gap (\%) vs.\ target $\mathcal{E}_{\text{rel}}$ on \texttt{Time Machine} for \texttt{Qwen3-1.7B} (left), \texttt{Qwen3-4B} (center), and \texttt{Qwen3-8B} (right), all at $\rho{=}2\%$.}
    \label{fig:ta-time-machine}
\end{figure}

\paragraph{Time Machine.}
\label{app:time-machine}
\Cref{fig:ta-time-machine} shows the token-accuracy gap vs.\ target $\mathcal{E}_{\text{rel}}$ on Time Machine for all three models.

\begin{figure}[h]
    \centering
    \raisebox{.7cm}{\rotatebox{90}{\fontsize{6}{7}\selectfont LM CE Gap}}\,%
    \includegraphics[width=0.19\linewidth]{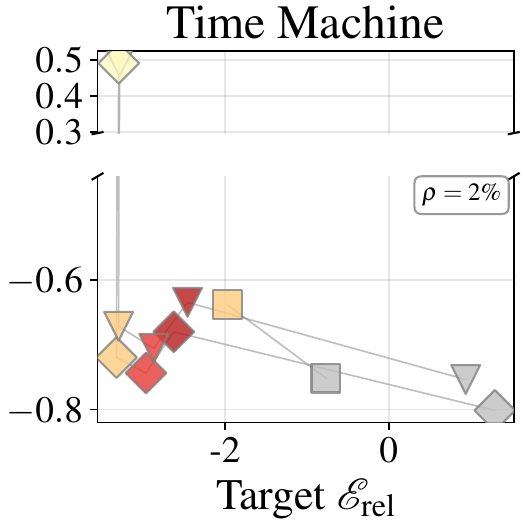}
    \includegraphics[width=0.19\linewidth]{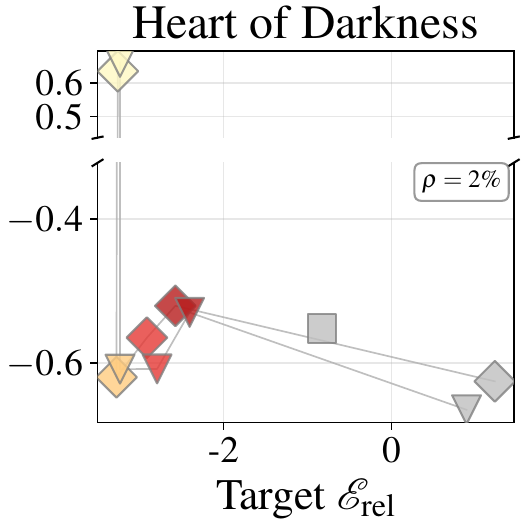}
    \includegraphics[width=0.19\linewidth]{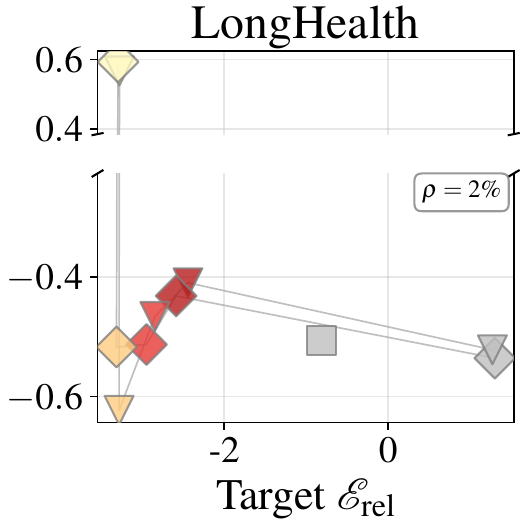}
    \includegraphics[width=0.19\linewidth]{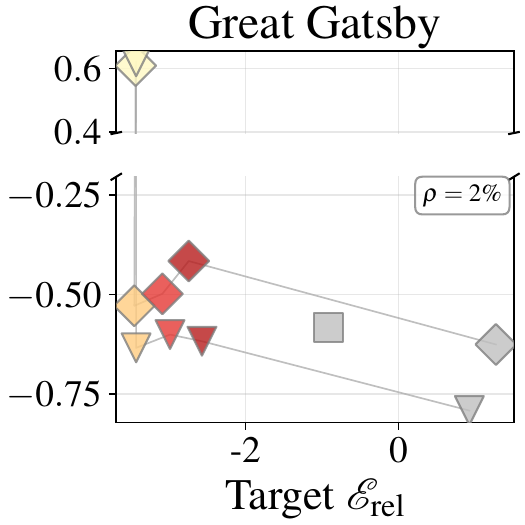}
    \includegraphics[width=0.19\linewidth]{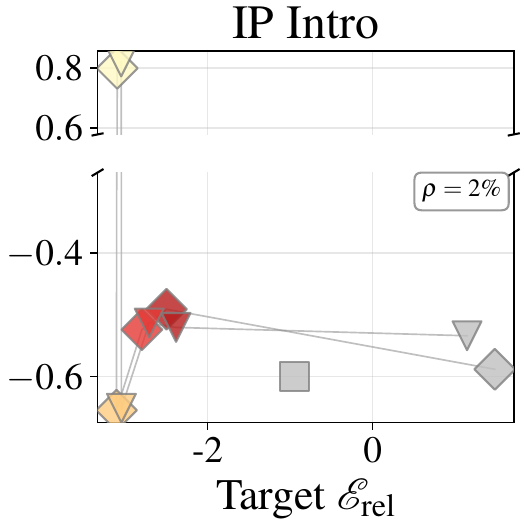}
    \\\vskip.1cm
    \includegraphics[width=\linewidth]{figures/scatter/kappa_pf2_legend.pdf}
    \caption{LM cross-entropy gap vs.\ target $\mathcal{E}_{\text{rel}}$ for \texttt{Qwen3-1.7B}.}
    \label{fig:ce-1.7B}
\end{figure}

\begin{figure}[h]
    \centering
    \raisebox{.7cm}{\rotatebox{90}{\fontsize{6}{7}\selectfont LM CE Gap}}\,%
    \includegraphics[width=0.24\linewidth]{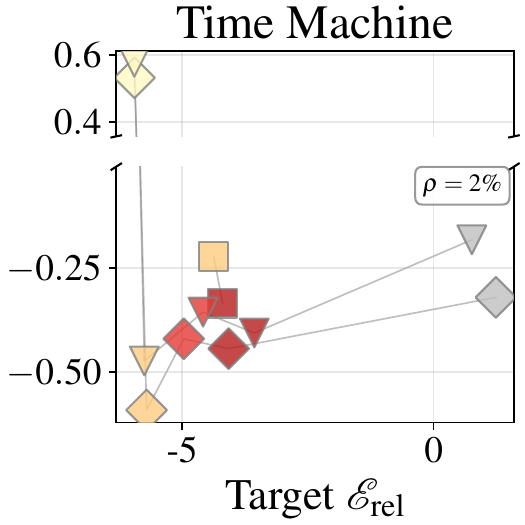}
    \includegraphics[width=0.24\linewidth]{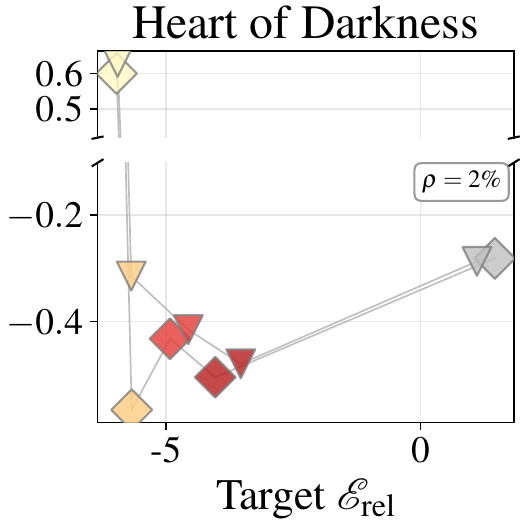}
    \includegraphics[width=0.24\linewidth]{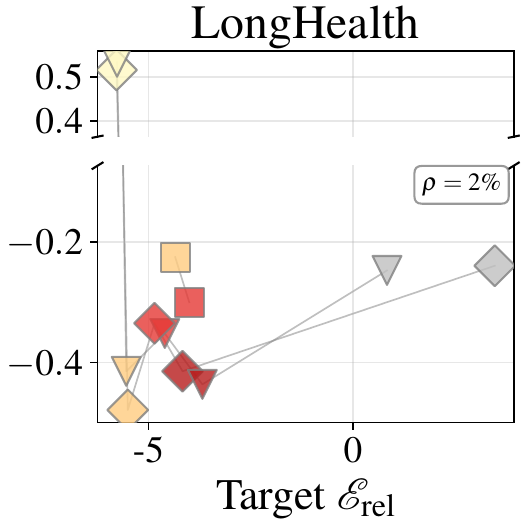}
    \includegraphics[width=0.24\linewidth]{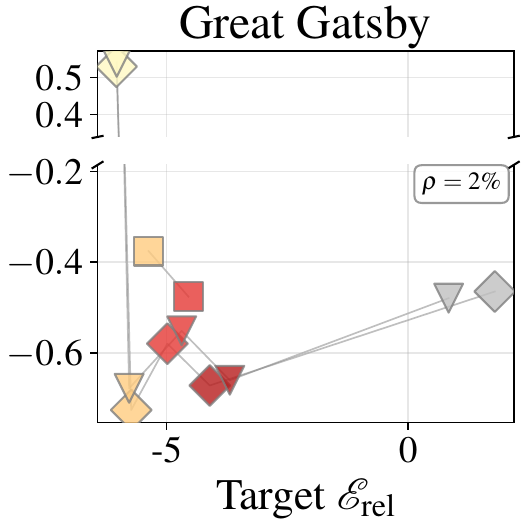}
    \\\vskip.1cm
    \includegraphics[width=\linewidth]{figures/scatter/raijin_pf2_legend.pdf}
    \caption{LM cross-entropy gap vs.\ target $\mathcal{E}_{\text{rel}}$ for \texttt{Qwen3-4B}.}
    \label{fig:ce-4B}
\end{figure}

\paragraph{LM cross-entropy gap.}
\Cref{fig:ce-1.7B,fig:ce-4B} show the LM cross-entropy gap vs.\ target $\mathcal{E}_{\text{rel}}$.

\begin{figure}[h]
    \centering
    \raisebox{.7cm}{\rotatebox{90}{\fontsize{6}{7}\selectfont KL Distill Loss}}\,%
    \includegraphics[width=0.23\linewidth]{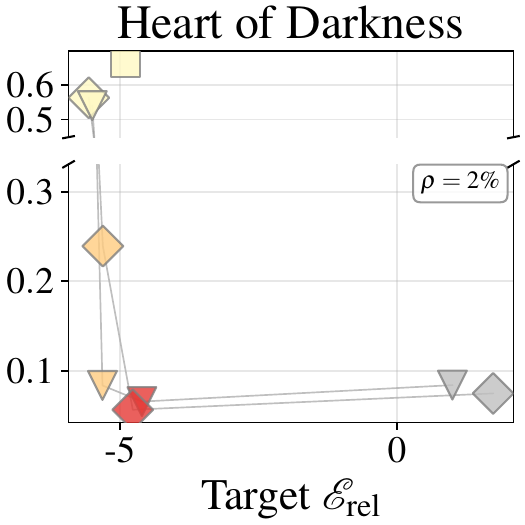}
    \includegraphics[width=0.23\linewidth]{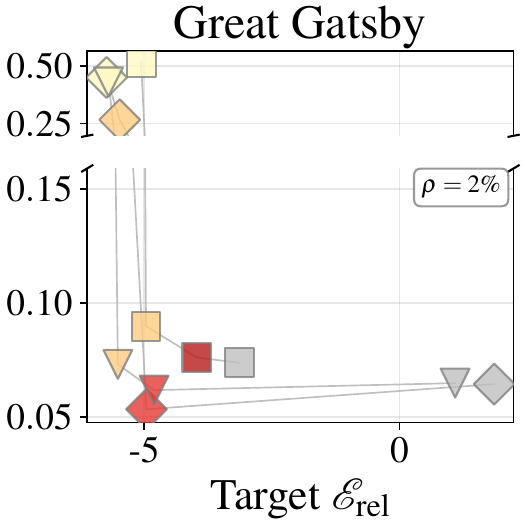}
    \includegraphics[width=0.23\linewidth]{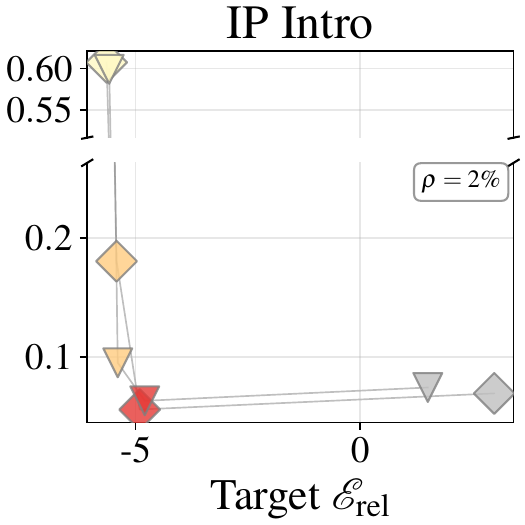}
    \includegraphics[width=0.23\linewidth]{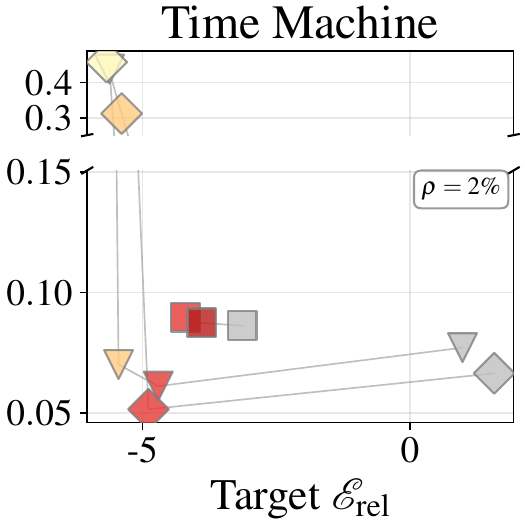}
    \\\vskip.1cm
    \includegraphics[width=\linewidth]{figures/scatter/komainu_pf2_legend.pdf}
    \caption{Eval KL distillation loss vs.\ target $\mathcal{E}_{\text{rel}}$ for \texttt{Qwen3-8B} at $\rho{=}2\%$.}
    \label{fig:kl-8B}
\end{figure}

\paragraph{KL distillation loss.}
\Cref{fig:kl-8B} shows the eval KL divergence between the teacher (full-attention) and student (\RegAtt{}) top-$K$ logit distributions, plotted against target $\mathcal{E}_{\text{rel}}$ for \texttt{Qwen3-8B}. This metric is computed on the held-out test set regardless of which losses were used during training. Lower KL indicates that \RegAtt{} produces next-token distributions closer to full attention.

\section{Generation Benchmarks}
\label{app:generation-bmk}

We benchmark end-to-end autoregressive generation comparing standard full-cache inference (Base) with \RegAtt{}-augmented inference.
\Cref{tab:generation-bmk} reports time to first token (TTFT), decode throughput, and peak GPU memory for \texttt{Qwen3-1.7B} and \texttt{Qwen3-4B} on three evaluation documents spanning 40k to 122k tokens.
In the base configuration the model performs a full prefill over all cached tokens before generating; with \RegAtt{}, the stored KV cache is replaced by the trained approximation modules, eliminating the prefill bottleneck.

\begin{table}[h]
\centering
\caption{End-to-end generation benchmarks comparing standard full-cache inference (Base) with \RegAtt{}-augmented inference on \texttt{Qwen3-1.7B} and \texttt{Qwen3-4B}. TTFT denotes time to first token, Tok/s the decode throughput, and Mem the peak GPU memory. Evaluation documents and context lengths correspond to \Cref{tab:datasets}.}
\label{tab:generation-bmk}
\footnotesize
\begin{tabular}{@{}l r rrr rrr@{}}
\toprule
& & \multicolumn{3}{c}{Base} & \multicolumn{3}{c}{\RegAtt{}} \\
\cmidrule(lr){3-5} \cmidrule(lr){6-8}
Model & Ctx. Len. & TTFT (ms) & Tok/s & Mem (GB) & TTFT (ms) & Tok/s & Mem (GB) \\
\midrule
\multirow{3}{*}{\texttt{Qwen3-1.7B}}
 & 40\,022  & 183\,000 & 9.5  & 24.9 & 26.4 & 44.2 & 7.7 \\
 & 61\,651  & 214\,100 & 7.3  & 34.2 & 26.4 & 42.9 & 7.9 \\
 & 122\,466 & 299\,900 & 4.4  & 60.4 & 29.3 & 37.7 & 8.2 \\
\midrule
\multirow{3}{*}{\texttt{Qwen3-4B}}
 & 40\,022  & 347\,900 & 6.3  & 38.6 & 44.6 & 18.4 & 16.7 \\
 & 61\,651  & 388\,700 & 4.0  & 50.6 & 47.0 & 17.5 & 16.8 \\
 & 122\,466 & 557\,600 & 1.5  & 84.2 & 50.1 & 15.7 & 17.1 \\
\bottomrule
\end{tabular}
\end{table}

\RegAtt{} reduces TTFT by three to four orders of magnitude---from hundreds of seconds to tens of milliseconds---by replacing the sequential prefill with a single forward pass through the compressed representation. Decode throughput improves by $3$--$10{\times}$, with larger gains at longer contexts where the full KV cache becomes increasingly expensive to attend over: at 122k tokens, \texttt{Qwen3-1.7B} improves from $4.4$ to $37.7$\,tok/s ($8.6{\times}$) and \texttt{Qwen3-4B} from $1.5$ to $15.7$\,tok/s ($10.5{\times}$). Peak memory follows the same trend, dropping by $4.9$--$7.4{\times}$ at the longest context. Crucially, both the TTFT and memory footprint of \RegAtt{} remain nearly constant across context lengths, confirming the fixed-cost inference predicted by the method.

\section{Sample Quality Breakdown}
\label{app:gen-breakdown}

\begin{figure}[h]
    \centering
    \includegraphics[width=\linewidth]{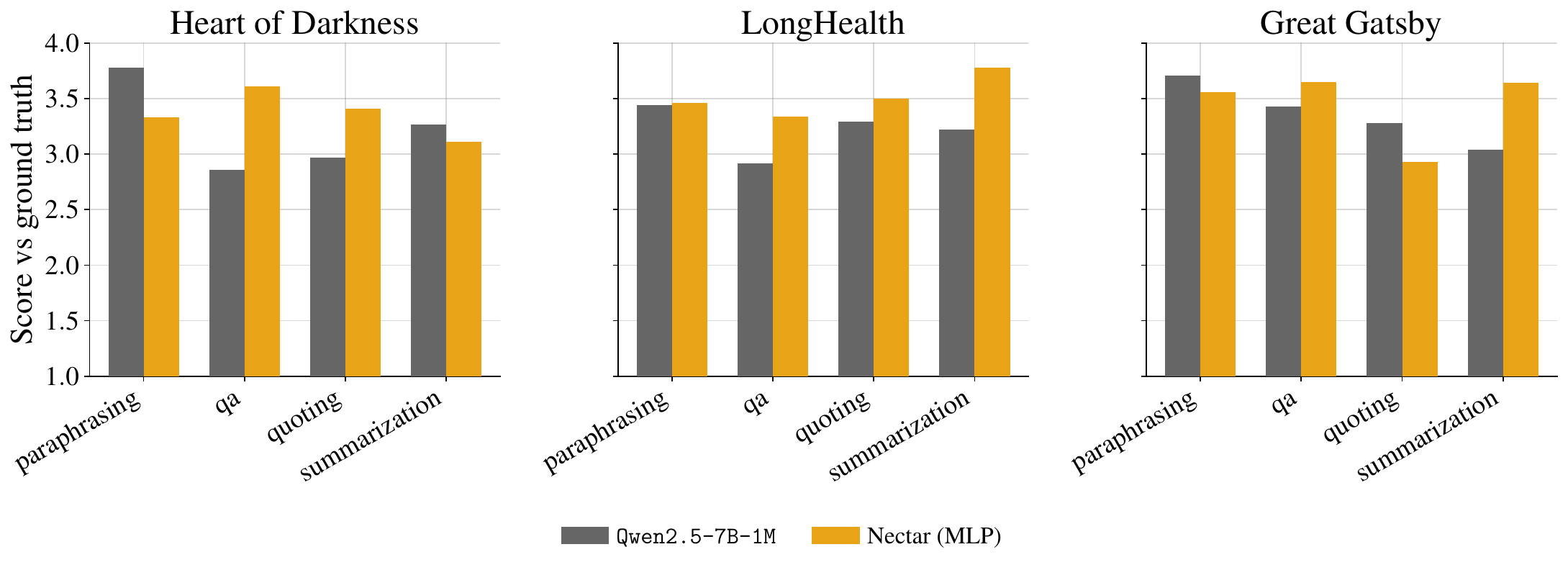}
    \caption{LLM-judge score (1--5, higher is better) per task type for the original model equipped with the full KV-cache and best MLP \RegAtt{} across three datasets.}
    \label{fig:gen-breakdown-mlp}
\end{figure}

Figure~\ref{fig:gen-breakdown-mlp} breaks down the LLM-judge scores (\S\ref{sec:exp:generation}) by task type for the best MLP \RegAtt{} configuration across all datasets. The MLP approximation preserves generation quality unevenly across task types: on QA and summarization it can match or occasionally surpass the base model. For instance on Great Gatsby, MLP scores are higher than full attention on both tasks. Quoting consistently shows the largest gap, as reproducing specific passages requires high approximation fidelity. LongHealth is the most challenging dataset, with larger gaps across all task types.

\subsection{Per-dataset breakdowns}

Table~\ref{tab:gen-pf-per-dataset} unpacks the cross-dataset average of Table~\ref{tab:gen-pf} into per-dataset values. MLP improves steadily with capacity on all three datasets, reaching negative $\Delta_s$ by $\rho{=}10\%$ on every dataset (and already at $\rho{=}5\%$ on Heart of Darkness and Great Gatsby); LongHealth is slowest, only crossing below zero at $\rho{=}10\%$. Quadrature remains above $2.1$ across all datasets and capacities.

\begin{table}[h]
\centering
\caption{Per-dataset $\Delta_s$ ($\downarrow$) vs.\ parameter fraction $\rho$. Best per (dataset, $\rho$) pair in bold.}
\label{tab:gen-pf-per-dataset}
\small
\begin{tabular}{@{}l l ccc@{}}
\toprule
Dataset & Model & $\rho{=}2\%$ & $\rho{=}5\%$ & $\rho{=}10\%$ \\
\midrule
\multirow{2}{*}{Heart of Darkness} & MLP    & $\mathbf{0.18}$         & $\mathbf{-0.14}$ & $\mathbf{-0.05}$ \\
                                   & Quad.\ & $2.21$ & $2.18$         & $2.18$          \\
\midrule
\multirow{2}{*}{LongHealth}        & MLP    & $\mathbf{0.39}$ & $\mathbf{0.16}$  & $\mathbf{-0.30}$ \\
                                   & Quad.\ & $2.19$         & $2.19$          & $2.14$          \\
\midrule
\multirow{2}{*}{Great Gatsby}      & MLP    & $\mathbf{0.26}$ & $\mathbf{-0.08}$ & $\mathbf{-0.08}$ \\
                                   & Quad.\ & $2.35$         & $2.33$          & $2.34$          \\
\bottomrule
\end{tabular}
\end{table}

Table~\ref{tab:gen-distill-w-per-dataset} breaks down the distillation-weight sweep by dataset and $\rho$. The best $\lambda_{\text{KL}}$ per (model, $\rho$) row is bolded. The optimal $\lambda_{\text{KL}}$ for MLP varies across dataset--capacity combinations: at $\rho{=}2\%$ the near-zero weight ($\lambda_{\text{KL}}{=}0.01$) is best on all three datasets, whereas at higher capacity pure distillation wins on Heart of Darkness ($\rho{=}5\%$--$10\%$) and LongHealth ($\rho{=}10\%$), and $\lambda_{\text{KL}}{=}2$ wins on Great Gatsby ($\rho{=}5\%$--$10\%$) and LongHealth ($\rho{=}5\%$). Quadrature is largely insensitive to $\lambda_{\text{KL}}$, remaining near $2.2$ throughout. The leftmost ``Pure $\mathcal{L}_{\text{reg}}$'' column ($\lambda_{\text{KL}}{=}0$, MLP only) trains with no distillation signal at all: the modules collapse to degenerate text and $\Delta_s$ stays at $2.2$--$2.4$ regardless of capacity (the values are identical across $\rho$ within each dataset), underscoring that the distillation term is essential for usable generations.

\begin{table}[h]
\centering
\caption{Per-dataset $\Delta_s$ ($\downarrow$) by $\rho$ and $\lambda_{\text{KL}}$. Empty cells indicate configurations not run. The ``Pure $\mathcal{L}_{\text{reg}}$'' column ($\lambda_{\text{KL}}{=}0$, regression only) is MLP-only and capacity-independent. Best $\lambda_{\text{KL}}$ per (model, $\rho$) row in bold.}
\label{tab:gen-distill-w-per-dataset}
\small
\begin{tabular}{@{}l c cc ccc@{}}
\toprule
& & \multicolumn{2}{c}{Pure} & \multicolumn{3}{c}{$\mathcal{L}_{\text{reg}} + \lambda_{\text{KL}}\,\mathcal{L}_{\text{KL}}$} \\
\cmidrule(lr){3-4} \cmidrule(lr){5-7}
Model & $\rho$ & $\mathcal{L}_{\text{reg}}$ & $\mathcal{L}_{\text{KL}}$ & $0.01$ & $2$ & $10$ \\
\midrule
\multicolumn{7}{@{}l}{\textit{Heart of Darkness}} \\
\midrule
\multirow{3}{*}{MLP}     & $2\%$  & $2.22$ & $0.98$ & $\mathbf{0.18}$ & $1.15$ & $1.05$ \\
                         & $5\%$  & $2.22$ & $\mathbf{-0.14}$ & $0.07$ & $-0.02$ & $0.26$ \\
                         & $10\%$ & $2.22$ & $\mathbf{-0.05}$ & $0.11$ & $0.04$ & $0.04$ \\
\addlinespace
\multirow{3}{*}{Quad.\ } & $2\%$  & --     & $\mathbf{2.21}$ & --     & --     & $\mathbf{2.21}$ \\
                         & $5\%$  & --     & --             & $2.18$ & --     & -- \\
                         & $10\%$ & --     & --             & --     & $2.18$ & -- \\
\midrule
\multicolumn{7}{@{}l}{\textit{LongHealth}} \\
\midrule
\multirow{3}{*}{MLP}     & $2\%$  & $2.22$ & $1.04$ & $\mathbf{0.39}$ & --     & $1.01$ \\
                         & $5\%$  & $2.22$ & $0.73$ & $0.45$ & $\mathbf{0.16}$ & $0.52$ \\
                         & $10\%$ & $2.22$ & $\mathbf{-0.30}$ & $0.61$ & $-0.03$ & $0.08$ \\
\addlinespace
\multirow{3}{*}{Quad.\ } & $2\%$  & --     & $\mathbf{2.19}$ & --     & --     & $2.22$ \\
                         & $5\%$  & --     & $2.22$ & $\mathbf{2.19}$ & $2.20$ & -- \\
                         & $10\%$ & --     & $\mathbf{2.14}$ & --     & $2.15$ & -- \\
\midrule
\multicolumn{7}{@{}l}{\textit{Great Gatsby}} \\
\midrule
\multirow{3}{*}{MLP}     & $2\%$  & $2.36$ & $0.62$ & $\mathbf{0.26}$ & $1.11$ & $0.80$ \\
                         & $5\%$  & $2.36$ & $0.24$ & $0.02$ & $\mathbf{-0.08}$ & $0.50$ \\
                         & $10\%$ & $2.36$ & $0.08$ & $0.14$ & $\mathbf{-0.08}$ & $0.49$ \\
\addlinespace
\multirow{3}{*}{Quad.\ } & $2\%$  & --     & $\mathbf{2.35}$ & --     & --     & $2.36$ \\
                         & $5\%$  & --     & --             & $2.34$ & $\mathbf{2.33}$ & -- \\
                         & $10\%$ & --     & --             & $2.34$ & --     & -- \\
\bottomrule
\end{tabular}
\end{table}

\subsection{Breakdowns restricted to QA}

The training data has a 5:1 ratio of QA pairs to the other task types (\S\ref{app:data}), and one might ask how the metric changes when restricted to QA. Tables~\ref{tab:gen-pf-qa} and~\ref{tab:gen-distill-w-qa} reproduce the main-text $\rho$- and $\lambda_{\text{KL}}$-breakdowns with $\Delta_s$ computed on QA samples only. MLP achieves negative QA $\Delta_s$ at every $\rho$ ($-0.14$, $-0.45$, $-0.44$), and on the distillation-weight axis the $\lambda_{\text{KL}}{=}2$ mixed regime gives the most negative QA gap ($-0.06$), with pure distillation and $\lambda_{\text{KL}}{=}0.01$ also marginally negative (both $-0.03$). Table~\ref{tab:gen-quality-qa} shows the per-task breakdown when the best configuration is selected by QA performance; the overall QA-selected gap is $-0.18$, matching the cross-task selection of Table~\ref{tab:gen-quality}. Table~\ref{tab:gen-quality-per-dataset-qa} unpacks the QA $\Delta_s$ per dataset: MLP surpasses the base model on all three, most strongly on Heart of Darkness ($-0.75$).

\begin{table}[h]
\centering
\caption{$\Delta_s$ ($\downarrow$) on QA samples only, vs.\ parameter fraction.}
\label{tab:gen-pf-qa}
\begin{tabular}{@{}l ccc@{}}
\toprule
& \multicolumn{3}{c}{$\rho$} \\
\cmidrule(lr){2-4}
& $2\%$ & $5\%$ & $10\%$ \\
\midrule
MLP    & $\mathbf{-0.14}${\tiny$\pm0.41$} & $\mathbf{-0.45}${\tiny$\pm0.66$} & $\mathbf{-0.44}${\tiny$\pm0.36$} \\
Quad.\ & $2.04${\tiny$\pm0.76$} & $2.02${\tiny$\pm0.80$} & $2.01${\tiny$\pm0.82$} \\
\bottomrule
\end{tabular}
\end{table}

\begin{table}[h]
\centering
\caption{$\Delta_s$ ($\downarrow$) on QA samples only, vs.\ distillation weight.}
\label{tab:gen-distill-w-qa}
\begin{tabular}{@{}l c ccc@{}}
\toprule
& Pure $\mathcal{L}_{\text{KL}}$ & \multicolumn{3}{c}{$\mathcal{L}_{\text{reg}} + \lambda_{\text{KL}}\,\mathcal{L}_{\text{KL}}$} \\
\cmidrule(lr){2-2} \cmidrule(lr){3-5}
&  & $0.01$ & $2$ & $10$ \\
\midrule
MLP    & $\mathbf{-0.03}${\tiny$\pm0.32$} & $\mathbf{-0.03}${\tiny$\pm0.19$} & $\mathbf{-0.06}${\tiny$\pm0.39$} & $\mathbf{0.17}${\tiny$\pm0.28$} \\
Quad.\ & $1.98${\tiny$\pm0.29$} & $2.12${\tiny$\pm0.51$} & $1.98${\tiny$\pm0.43$} & $2.05${\tiny$\pm0.80$} \\
\bottomrule
\end{tabular}
\end{table}

\begin{table}[h]
\centering
\caption{$\Delta_s$ ($\downarrow$) by task type, best configuration selected by QA performance. Counterpart to Table~\ref{tab:gen-quality}.}
\label{tab:gen-quality-qa}
\begin{tabular}{@{}l ccccc@{}}
\toprule
& Paraphr.\ & QA & Quoting & Summ.\ & Overall \\
\midrule
\RegAtt{} (MLP) & $\mathbf{0.16}${\tiny$\pm0.63$} & $\mathbf{-0.49}${\tiny$\pm0.58$} & $\mathbf{-0.06}${\tiny$\pm1.18$} & $\mathbf{-0.31}${\tiny$\pm1.02$} & $\mathbf{-0.18}${\tiny$\pm0.45$} \\
Quadrature      & $2.61${\tiny$\pm0.55$} & $2.01${\tiny$\pm0.83$} & $2.10${\tiny$\pm0.48$} & $2.18${\tiny$\pm0.30$} & $2.22${\tiny$\pm0.42$} \\
\bottomrule
\end{tabular}
\end{table}

\begin{table}[h]
\centering
\caption{QA $\Delta_s$ ($\downarrow$) per dataset, best run per architecture. Counterpart to Table~\ref{tab:gen-quality-per-dataset}.}
\label{tab:gen-quality-per-dataset-qa}
\begin{tabular}{@{}l cccc@{}}
\toprule
& Heart of Dark.\ & LongHealth & Great Gatsby & Mean \\
\midrule
\RegAtt{} (MLP) & $\mathbf{-0.75}$ & $\mathbf{-0.42}$ & $\mathbf{-0.22}$ & $\mathbf{-0.46}$ \\
Quadrature      & $1.80$ & $1.84$ & $2.39$ & $2.01$ \\
\bottomrule
\end{tabular}
\end{table}

\section{Out-of-Distribution Generalization: The Challenge Split}
\label{app:challenge}

The generation-quality results in the main text (\S\ref{sec:exp:generation}) are computed on a uniform random subsample of $500$ held-out instructions per dataset. To probe how \RegAtt{} behaves on prompts deliberately far from the training distribution, we additionally construct a \emph{Challenge} split for each dataset. Every held-out test instruction is embedded with a sentence encoder and scored by its maximum cosine similarity to any training instruction; the $500$ instructions with the \emph{lowest} such maximum similarity, those least similar to anything seen during self-study, are selected per dataset. Namely, semantic similarities between train and test instructions are measured with cosine similarities due to embedding using the \texttt{all-MiniLM-L6-v2}~\citep{reimers-2019-sentence-bert} encoder.

Because each module is fit to a single context and its self-study pairs are drawn from the same document, this split stresses precisely the regime in which the learned approximation is least supported by training pairs. We re-evaluate all models using this split. Accordingly, the base-model reference scores $\bar{s}_{\text{base}}$ are recomputed on the challenge instructions and are higher than on the random split ($3.82$/$3.87$/$3.68$ for Heart of Darkness/LongHealth/Great Gatsby, vs.\ $3.22$/$3.22$/$3.36$), reflecting the harder, more dissimilar question mix.

\Cref{tab:gen-pf-challenge,tab:gen-distill-w-challenge,tab:gen-quality-challenge,tab:gen-quality-per-dataset-challenge} reproduce the four main-text generation tables on the challenge split, and \Cref{tab:gen-distill-w-per-dataset-challenge} gives the full per-dataset breakdown (\RegAtt{} modules trained with pure regression were not evaluated on this split, so the corresponding column is omitted).

The findings of \S\ref{sec:exp:generation} carry over. For instance, \RegAtt{} outperforms the Quadrature baseline by a wide margin and the MLP gap still shrinks monotonically with capacity ($1.40 {\to} 1.28 {\to} 1.18$ as $\rho$ grows from $2\%$ to $10\%$). What changes are the gaps with respect to baseline performance which increase on these deliberately out-of-distribution prompts. Consistent with the random split, \emph{quoting} is the hardest task, as verbatim reproduction of unfamiliar passages demands the highest approximation fidelity. The advantage of adding regression over pure distillation is also observed in several cases here. Overall, the challenge split indicates that the per-context approximation generalizes well within a document's distribution but, as one would expect, the gap increases as one pushes models out of the training distribution.

\begin{table}[h]
\centering
\caption{\textbf{Challenge split.} $\Delta_s$ ($\downarrow$) vs.\ parameter fraction $\rho$. $\pm$: 95\% CI. Best per column in bold.}
\label{tab:gen-pf-challenge}
\footnotesize
\begin{tabular}{@{}l ccc@{}}
\toprule
& \multicolumn{3}{c}{$\rho$} \\
\cmidrule(lr){2-4}
& $2\%$ & $5\%$ & $10\%$ \\
\midrule
MLP    & $\mathbf{1.40}${\tiny$\pm0.54$} & $\mathbf{1.28}${\tiny$\pm0.49$} & $\mathbf{1.18}${\tiny$\pm0.43$} \\
Quad.\ & $2.79${\tiny$\pm0.23$} & $2.78${\tiny$\pm0.23$} & $2.78${\tiny$\pm0.23$} \\
\bottomrule
\end{tabular}
\end{table}

\begin{table}[h]
\centering
\caption{\textbf{Challenge split.} $\Delta_s$ ($\downarrow$) vs.\ distillation weight. ``Pure $\mathcal{L}_{\text{KL}}$'' uses only the KL loss; the others combine the regression loss with distillation. $\pm$: 95\% CI. Best per column in bold.}
\label{tab:gen-distill-w-challenge}
\footnotesize
\begin{tabular}{@{}l c ccc@{}}
\toprule
& Pure $\mathcal{L}_{\text{KL}}$ & \multicolumn{3}{c}{$\mathcal{L}_{\text{reg}} + \lambda_{\text{KL}}\,\mathcal{L}_{\text{KL}}$} \\
\cmidrule(lr){2-2} \cmidrule(lr){3-5}
&  & $0.01$ & $2$ & $10$ \\
\midrule
MLP    & $\mathbf{1.49}${\tiny$\pm0.25$} & $\mathbf{1.46}${\tiny$\pm0.16$} & $\mathbf{1.52}${\tiny$\pm0.31$} & $\mathbf{1.77}${\tiny$\pm0.22$} \\
Quad.\ & $2.82${\tiny$\pm0.10$} & $2.76${\tiny$\pm0.15$} & $2.81${\tiny$\pm0.14$} & $2.79${\tiny$\pm0.24$} \\
\bottomrule
\end{tabular}
\end{table}

\begin{table}[h]
\centering
\caption{\textbf{Challenge split.} $\Delta_s$ ($\downarrow$) by task type, averaged across datasets. $\pm$: 95\% CI. Best per column in bold.}
\label{tab:gen-quality-challenge}
\footnotesize
\begin{tabular}{@{}l ccccc@{}}
\toprule
& Paraphr.\ & QA & Quoting & Summ.\ & Overall \\
\midrule
\RegAtt{} (MLP) & $\mathbf{0.79}${\tiny$\pm0.42$} & $\mathbf{0.95}${\tiny$\pm0.53$} & $\mathbf{1.98}${\tiny$\pm0.49$} & $\mathbf{0.78}${\tiny$\pm0.54$} & $\mathbf{1.12}${\tiny$\pm0.91$} \\
Quadrature      & $2.91${\tiny$\pm0.23$} & $2.57${\tiny$\pm0.07$} & $3.15${\tiny$\pm0.40$} & $2.50${\tiny$\pm1.38$} & $2.78${\tiny$\pm0.49$} \\
\bottomrule
\end{tabular}
\end{table}

\begin{table}[h]
\centering
\caption{\textbf{Challenge split.} $\Delta_s$ ($\downarrow$) per dataset, best run per architecture. $\pm$: 95\% CI. Best per column in bold.}
\label{tab:gen-quality-per-dataset-challenge}
\footnotesize
\begin{tabular}{@{}l cccc@{}}
\toprule
& Heart of Dark.\ & LongHealth & Great Gatsby & Mean \\
\midrule
\RegAtt{} (MLP) & $\mathbf{0.97}${\tiny$\pm1.17$} & $\mathbf{1.15}${\tiny$\pm0.66$} & $\mathbf{1.25}${\tiny$\pm0.92$} & $\mathbf{1.12}${\tiny$\pm0.35$} \\
Quadrature      & $2.81${\tiny$\pm0.64$} & $2.86${\tiny$\pm0.38$} & $2.68${\tiny$\pm0.83$} & $2.78${\tiny$\pm0.23$} \\
\bottomrule
\end{tabular}
\end{table}

\begin{table}[h]
\centering
\caption{\textbf{Challenge split.} Per-dataset $\Delta_s$ ($\downarrow$) by $\rho$ and $\lambda_{\text{KL}}$. Empty cells indicate configurations not run. Best $\lambda_{\text{KL}}$ per (model, $\rho$) row in bold.}
\label{tab:gen-distill-w-per-dataset-challenge}
\small
\begin{tabular}{@{}l c c ccc@{}}
\toprule
& & Pure $\mathcal{L}_{\text{KL}}$ & \multicolumn{3}{c}{$\mathcal{L}_{\text{reg}} + \lambda_{\text{KL}}\,\mathcal{L}_{\text{KL}}$} \\
\cmidrule(lr){3-3} \cmidrule(lr){4-6}
Model & $\rho$ & & $0.01$ & $2$ & $10$ \\
\midrule
\multicolumn{6}{@{}l}{\textit{Heart of Darkness}} \\
\midrule
\multirow{3}{*}{MLP}     & $2\%$  & $1.99$ & $\mathbf{1.51}$ & $2.24$ & $2.07$ \\
                         & $5\%$  & $\mathbf{1.10}$ & $1.21$ & $1.12$ & $1.47$ \\
                         & $10\%$ & $\mathbf{0.98}$ & $1.38$ & $1.23$ & $1.30$ \\
\addlinespace
\multirow{3}{*}{Quad.\ } & $2\%$  & $\mathbf{2.82}$ & --     & --     & $\mathbf{2.82}$ \\
                         & $5\%$  & --     & $2.81$ & --     & -- \\
                         & $10\%$ & --     & --     & $2.81$ & -- \\
\midrule
\multicolumn{6}{@{}l}{\textit{LongHealth}} \\
\midrule
\multirow{3}{*}{MLP}     & $2\%$  & $2.09$ & $\mathbf{1.15}$ & --     & $2.35$ \\
                         & $5\%$  & $1.65$ & $1.88$ & $\mathbf{1.49}$ & $1.96$ \\
                         & $10\%$ & $\mathbf{1.28}$ & $1.80$ & $1.30$ & $1.48$ \\
\addlinespace
\multirow{3}{*}{Quad.\ } & $2\%$  & $\mathbf{2.86}$ & --     & --     & $2.87$ \\
                         & $5\%$  & $\mathbf{2.86}$ & $2.87$ & $2.87$ & -- \\
                         & $10\%$ & $\mathbf{2.86}$ & --     & $\mathbf{2.86}$ & -- \\
\midrule
\multicolumn{6}{@{}l}{\textit{Great Gatsby}} \\
\midrule
\multirow{3}{*}{MLP}     & $2\%$  & $1.72$ & $\mathbf{1.54}$ & $2.21$ & $1.95$ \\
                         & $5\%$  & $1.32$ & $1.31$ & $\mathbf{1.25}$ & $1.67$ \\
                         & $10\%$ & $\mathbf{1.28}$ & $1.39$ & $1.36$ & $1.65$ \\
\addlinespace
\multirow{3}{*}{Quad.\ } & $2\%$  & $\mathbf{2.68}$ & --     & --     & $\mathbf{2.68}$ \\
                         & $5\%$  & --     & $\mathbf{2.68}$ & $\mathbf{2.68}$ & -- \\
                         & $10\%$ & --     & $2.68$ & --     & -- \\
\bottomrule
\end{tabular}
\end{table}

\section{Side-by-Side \RegAtt{} (MLP) vs.\ Quadrature Generation Examples at $\rho{=}10\%$}
\label{app:gen-side-by-side}

\newcounter{compsample}
\renewcommand{\thecompsample}{\arabic{compsample}}

Tables~\ref{tab:comp-gatsby}--\ref{tab:comp-longhealth} present matched generation examples at the largest tested capacity ($\rho{=}10\%$), using \texttt{Qwen2.5-7B-1M} and the best configuration per architecture and dataset. Each sample shows the instruction, ground truth, Full KV-cache response, \RegAtt{} (MLP) response, and Quadrature response. Non-Latin glyphs in Quadrature outputs are rendered as bracketed placeholders (\texttt{[CJK]}, \texttt{[Arabic]}, \texttt{[symbol]}).

\textbf{Great Gatsby.} (Table~\ref{tab:comp-gatsby}) The MLP model retrieves correct factual content on all five samples and produces semantically equivalent responses to the full-KV-cache baseline; notably, sample~\ref{sample:comp-gatsby-vitality} returns the exact ground-truth passage verbatim. The Quadrature model produces incoherent text on all five with no discernible relation to the instruction.

\begin{table}[h]
\centering
\caption{Side-by-side generation examples on Great Gatsby.\label{tab:comp-gatsby}}
\setcounter{compsample}{0}
\scriptsize
\refstepcounter{compsample}\label{sample:comp-gatsby-myrtle}%
\begin{tabular}{@{}p{0.14\linewidth} p{0.82\linewidth}@{}}
\toprule
\textbf{Instruction} & Who was the principal witness at the inquest for Myrtle Wilson's death? \\
\midrule
\textbf{Ground truth} & The young Greek, Michaelis, was the principal witness at the inquest. \\
\midrule
\textbf{Full KV-cache} & the young Greek, Michaelis, who ran the coffee joint beside the ash-heaps was the principal witness at the inquest. \\
\midrule
\textbf{\RegAtt{} (MLP)} & Based on the information provided in the passage, the principal witness at the inquest for Myrtle Wilson's death was Michaelis, the young Greek who ran the coffee joint beside the ash-heaps. \\
\midrule
\textbf{Quadrature} & Based [Arabic] sits Jay Gatsby's death. The narrator attached the war, a green incident in a score identity. Specificallyatedable music. Specifically that her hope légèreires shoes. Specifically Myrtle. Eckleainlerers Wilson Buchanan recognizes a fountain of officer. Hereizes the justification and leadership. \\
\bottomrule
\end{tabular}

\vspace{6pt}
\refstepcounter{compsample}\label{sample:comp-gatsby-rain}%
\begin{tabular}{@{}p{0.14\linewidth} p{0.82\linewidth}@{}}
\toprule
\textbf{Instruction} & How does Gatsby react when the rain stops? \\
\midrule
\textbf{Ground truth} & Gatsby becomes overly excited about the stopped rain and the resulting sunshine, expressing pride in his house's appearance. \\
\midrule
\textbf{Full KV-cache} & According to the text, when the rain stops, Gatsby becomes excited and expresses joy at the change in weather. Specifically: Gatsby says, ``Has it?'' when Nick mentions the rain has stopped, showing his surprise. Gatsby then smiles like a weatherman, exclaiming, ``What do you think of that?'' \\
\midrule
\textbf{\RegAtt{} (MLP)} & Based on the passage, when Gatsby hears the rain stop, he initially reacts with excitement and hope. Specifically: He says ``Oh, it's stopped raining!'' and then looks around, as if to check on the weather. This indicates his eagerness to see Daisy. \\
\midrule
\textbf{Quadrature} & Based about the Colrtle Egg, drugside with Meyer history that widow overseas room. Specifically there and that if they should have the Arges guide. Some couples running. Specifically there West Egg Wilson's house. Specificallys beauty. He was brought out. His fingers dog. His car talks. Eckleburg Buchanan Cook \\
\bottomrule
\end{tabular}

\vspace{6pt}
\refstepcounter{compsample}\label{sample:comp-gatsby-tom}%
\begin{tabular}{@{}p{0.14\linewidth} p{0.82\linewidth}@{}}
\toprule
\textbf{Instruction} & What physical accomplishment is Tom Buchanan known for? \\
\midrule
\textbf{Ground truth} & Tom Buchanan was one of the most powerful ends that ever played football at New Haven. \\
\midrule
\textbf{Full KV-cache} & Tom Buchanan is known for being one of the most powerful football players, specifically an ``end,'' that ever played at New Haven (Yale). He was described as ``one of the most powerful ends that ever played football at New Haven---a national figure in a way.'' \\
\midrule
\textbf{\RegAtt{} (MLP)} & Based on the passage, Tom Buchanan is known for being one of the most powerful ends that ever played football at New Haven. Specifically, he was a powerful football player who had played in the finals at New Haven. \\
\midrule
\textbf{Quadrature} & Based of being driving side-room trains and everythingage Platt and lemon polographer over the St. Specifically sand breath, and driveser. Specifically ignore tradingcyour coach. The photograph.17 vitality in Gatsby's Series. Wolfshiemts Wilson Buchanan Buchanan, Doc Carson, and sitting with Myrtle with Meyer \\
\bottomrule
\end{tabular}

\vspace{6pt}
\refstepcounter{compsample}\label{sample:comp-gatsby-eyes}%
\begin{tabular}{@{}p{0.14\linewidth} p{0.82\linewidth}@{}}
\toprule
\textbf{Instruction} & What physical features are attributed to Doctor T.\,J.\ Eckleburg's eyes? \\
\midrule
\textbf{Ground truth} & Doctor T.\,J.\ Eckleburg's eyes are described as blue and gigantic, with retinas one yard high. \\
\midrule
\textbf{Full KV-cache} & Based on the description in the text, Doctor T.\,J.\ Eckleburg's eyes have the following physical features: 1.\ They are blue in color. 2.\ They are described as ``gigantic'' with retinas that are one yard high. 3.\ They ``look out of no face, but, instead, from a pair of enormous yellow spectacles.'' \\
\midrule
\textbf{\RegAtt{} (MLP)} & Based on the passage, Doctor T.\,J.\ Eckleburg's eyes are described as blue and gigantic, with retinas one yard high. They look out from a pair of enormous yellow spectacles that pass over a nonexistent nose. \\
\midrule
\textbf{Quadrature} & Based to go to the eyes of Ashes on the oven where they weren't Poleceral bridge out the narrator's yacht with a picture of Doctor T. J. Specificallyitableian, asking it signifies how this. Eckleburg Wilson get evidence that are taken back.300000000 \\
\bottomrule
\end{tabular}

\vspace{6pt}
\refstepcounter{compsample}\label{sample:comp-gatsby-vitality}%
\begin{tabular}{@{}p{0.14\linewidth} p{0.82\linewidth}@{}}
\toprule
\textbf{Instruction} & Quote the description of Myrtle Wilson's appearance and vitality. \\
\midrule
\textbf{Ground truth} & She was in the middle thirties, and faintly stout, but she carried her flesh sensuously as some women can. Her face, above a spotted dress of dark blue cr\^{e}pe-de-chine, contained no facet or gleam of beauty, but there was an immediately perceptible vitality about her as if the nerves of her body were continually smouldering. \\
\midrule
\textbf{Full KV-cache} & Mrs.\ Wilson had changed her costume some time before, and was now attired in an elaborate afternoon dress of cream-coloured chiffon, which gave out a continual rustle as she swept about the room. With the influence of the dress her personality had also undergone a change. The intense vitality that had been so remarkable in the garage was converted into impressive hauteur. \\
\midrule
\textbf{\RegAtt{} (MLP)} & She was in the middle thirties, and faintly stout, but she carried her flesh sensuously as some women can. Her face, above a spotted dress of dark blue cr\^{e}pe-de-chine, contained no facet or gleam of beauty, but there was an immediately perceptible vitality about her as if the nerves of her body were continually smouldering. \\
\midrule
\textbf{Quadrature} & Heres my servants Daisy's body, do you hear, with her armisting Wolfshiem's body. The photograph of windshieldsering towels thatches towels thatched her nose. The faces towelsing her strength and archives. She looks back as if she became her demeanor. She was linkedively, butleruchenneering, suggesting he was a scheduleied coat shoes, and her husbandless and gigantic sunlight, however that she said she was forever understood Gatsby in the road. \\
\bottomrule
\end{tabular}
\end{table}

\textbf{Heart of Darkness.} (Table~\ref{tab:comp-heart}) The MLP model retrieves the Kurtz phrase (sample~\ref{sample:comp-hod-horror}), the correct book title (sample~\ref{sample:comp-hod-book}), the exact painting description (sample~\ref{sample:comp-hod-painting}), and a fluent symbolic paraphrase (sample~\ref{sample:comp-hod-heads}); for the beetle metaphor (sample~\ref{sample:comp-hod-beetle}) it adds explanatory context beyond the exact quote. The Quadrature model is largely incoherent; sample~\ref{sample:comp-hod-horror} produces only a truncated fragment.

\begin{table}[h]
\centering
\caption{Side-by-side generation examples on Heart of Darkness.\label{tab:comp-heart}}
\setcounter{compsample}{0}
\scriptsize
\refstepcounter{compsample}\label{sample:comp-hod-beetle}%
\begin{tabular}{@{}p{0.14\linewidth} p{0.82\linewidth}@{}}
\toprule
\textbf{Instruction} & Quote the metaphor describing the steamboat as a sluggish beetle crawling on the floor of a lofty portico. \\
\midrule
\textbf{Ground truth} & like a sluggish beetle crawling on the floor of a lofty portico \\
\midrule
\textbf{Full KV-cache} & ``\dots\ crept the little begrimed steamboat, like a sluggish beetle crawling on the floor of a lofty portico.'' \\
\midrule
\textbf{\RegAtt{} (MLP)} & The metaphor used to describe the steamboat is ``like a sluggish beetle crawling on the floor of a lofty portico,'' which is used to emphasize the steamboat's slow and labored progress through the jungle. \\
\midrule
\textbf{Quadrature} & Here describe a file was seen just vegetation area ragazzoary of SeamanshipaddItem and windows. The motiven'tstop cellspacingled trees was brown bushes froked, red, red, and the back, patches on elbows, patches on elbows, patches on elbows, patches on elbowsages, patches on elbows, patches on elbows, patchesing at the bottom of light plankslean, and the sunshine made him look extremely patching behavior. \\
\bottomrule
\end{tabular}

\vspace{6pt}
\refstepcounter{compsample}\label{sample:comp-hod-horror}%
\begin{tabular}{@{}p{0.14\linewidth} p{0.82\linewidth}@{}}
\toprule
\textbf{Instruction} & What phrase did Kurtz cry out during his final moments? \\
\midrule
\textbf{Ground truth} & Kurtz cried out `The horror! The horror!' during his final moments. \\
\midrule
\textbf{Full KV-cache} & Kurtz cried out ``The horror! The horror!'' during his final moments, according to the passage. \\
\midrule
\textbf{\RegAtt{} (MLP)} & According to the text, Kurtz's last words were ``The horror! The horror!'' These words were whispered to Marlow just before Kurtz's death and represent his final recognition of the darkness within himself and the world. \\
\midrule
\textbf{Quadrature} & Baseded Kurtz's death \\
\bottomrule
\end{tabular}

\vspace{6pt}
\refstepcounter{compsample}\label{sample:comp-hod-painting}%
\begin{tabular}{@{}p{0.14\linewidth} p{0.82\linewidth}@{}}
\toprule
\textbf{Instruction} & Quote the description of the painting in the brickmaker's room. \\
\midrule
\textbf{Ground truth} & a small sketch in oils, on a panel, representing a woman, draped and blindfolded, carrying a lighted torch \\
\midrule
\textbf{Full KV-cache} & The movement of the woman was stately, and the effect of the torchlight on the face was sinister. \\
\midrule
\textbf{\RegAtt{} (MLP)} & Here was a small sketch in oils, on a panel, representing a woman, draped and blindfolded, carrying a lighted torch. The background was somber---almost black. The movement of the woman was stately, and the effect of the torchlight on the face was sinister. \\
\midrule
\textbf{Quadrature} & Here's wives, bricous clothes was as a complexed torch wire gaoked, moving at the seaico hut. It was surrounded by a club is depicted as if I justages notes.' \\
\bottomrule
\end{tabular}

\vspace{6pt}
\refstepcounter{compsample}\label{sample:comp-hod-book}%
\begin{tabular}{@{}p{0.14\linewidth} p{0.82\linewidth}@{}}
\toprule
\textbf{Instruction} & What is the title of the book found in the hut? \\
\midrule
\textbf{Ground truth} & The book is titled \emph{An Inquiry into some Points of Seamanship}. \\
\midrule
\textbf{Full KV-cache} & Towson's book found in the hut was titled ``An Inquiry into Some Points of Seamanship'' by a man named Towson or Towser. \\
\midrule
\textbf{\RegAtt{} (MLP)} & The title of the book found in the hut is ``An Inquiry into some Points of Seamanship,'' by a man named Tower or Towson. \\
\midrule
\textbf{Quadrature} & The title is titled ``Egypt silence on the stakes of Seamanship,' leaves tobaccoist beef, and science, and science named cipher. The horror! The boiler had been made him alone words is done. Specifically leave, was in Europe, appears as a book of the narrator suggests. This have been urgent from disease, \\
\bottomrule
\end{tabular}

\vspace{6pt}
\refstepcounter{compsample}\label{sample:comp-hod-heads}%
\begin{tabular}{@{}p{0.14\linewidth} p{0.82\linewidth}@{}}
\toprule
\textbf{Instruction} & Rephrase the explanation of the symbolic meaning of the heads on the stakes. \\
\midrule
\textbf{Ground truth} & The heads on the stakes symbolize Kurtz's loss of restraint and his descent into moral corruption, reflecting the destructive impact of his unchecked desires and the wilderness's retribution against his invasion. \\
\midrule
\textbf{Full KV-cache} & The heads on those stakes were not merely a cruel display of power, but a haunting reminder of Mr.\ Kurtz's lack of restraint. They symbolized a fundamental emptiness within him, a deficiency that, when the moment of need arrived, could not be found amid his grandiloquent words. Whether Kurtz was aware of this flaw himself is uncertain, but the wilderness had long since uncovered it. \\
\midrule
\textbf{\RegAtt{} (MLP)} & The heads on the stakes served as a powerful symbol of Kurtz's moral deterioration---a physical manifestation of the darkness that had consumed him. They reflected the unrestrained nature of his actions, highlighting the conflict between his initial ideals and the brutal reality of his conduct in the wilderness. \\
\midrule
\textbf{Quadrature} & The accountant's office paints a group of Seamanship and starvation' crawling gunmen with a blindfolded collar, disgust, Freslevenals booterative movement, a melanchous presence. They represent a spearing the earth. This interview that the expedition's trade. The Russian \texttt{fkk} Fresleven's personality, Freslequinles with the moral ambiguity of the Company's dreadful environment. \\
\bottomrule
\end{tabular}
\end{table}

\textbf{LongHealth.} (Table~\ref{tab:comp-longhealth}) The MLP model produces accurate responses for all five. The Quadrature model confuses patient identities, introduces unrelated conditions, and generates garbled medical terminology throughout.

\begin{table}[h]
\centering
\caption{Side-by-side generation examples on LongHealth.\label{tab:comp-longhealth}}
\setcounter{compsample}{0}
\scriptsize
\refstepcounter{compsample}\label{sample:comp-lh-braf}%
\begin{tabular}{@{}p{0.14\linewidth} p{0.82\linewidth}@{}}
\toprule
\textbf{Instruction} & Quote the statement about the BRAF V600E mutation and its therapeutic implications. \\
\midrule
\textbf{Ground truth} & BRAF V600E mutation. Our recommended therapeutic combination includes Vemurafenib and Cobimetinib, both of which are aimed at disrupting the aberrant BRAF-MEK signaling cascade. \\
\midrule
\textbf{Full KV-cache} & Our molecular profile has revealed the presence of the BRAF V600E mutation. Implications: Based on this molecular profile, our recommended therapeutic combination includes Vemurafenib and Cobimetinib, both of which are aimed at disrupting the aberrant BRAF-MEK signaling cascade. Complementing this, we suggest the administration of Pembrolizumab. \\
\midrule
\textbf{\RegAtt{} (MLP)} & The statement about the molecular profile of the BRAF V600E mutation and the recommended therapeutic combination includes Vemurafenib and Cobimetinib, both of which are aimed at disrupting the aberrant BRAF-MEK signaling cascade. Complementing this, we suggest the administration of Pembrolizumab. \\
\midrule
\textbf{Quadrature} & \#\#\#[symbol][symbol] \texttt{fkk} (IDHoblastoma, WHO CNS grade 40 GyLoginib, with a BRAgenmidoccusF, Margethasone, Margethidermidisalinalved 45.200 ALmax, and an DurRANO and SUVmax, and LVEmonitor of 18) translocation. \\
\bottomrule
\end{tabular}

\vspace{6pt}
\refstepcounter{compsample}\label{sample:comp-lh-immuno}%
\begin{tabular}{@{}p{0.14\linewidth} p{0.82\linewidth}@{}}
\toprule
\textbf{Instruction} & What immunotherapy drugs were prescribed to Mrs.\ Done starting in May 2020? \\
\midrule
\textbf{Ground truth} & Nivolumab and Ipilimumab were prescribed as a combination immunotherapy regimen. \\
\midrule
\textbf{Full KV-cache} & Based on the information provided in the documents, Mrs.\ Done started the following immunotherapy drugs in May 2020: 1.\ Nivolumab --- 60 mg, started on May 2, 2020. 2.\ Ipilimumab --- 20\,[\dots] \\
\midrule
\textbf{\RegAtt{} (MLP)} & Based on the patient record, Mrs.\ Done was prescribed Nivolumab (60 mg) and Ipilimumab as a combination immunotherapy regimen starting in May 2020 for treatment of her metastatic malignant melanoma. \\
\midrule
\textbf{Quadrature} & Basedoratory and Dexamethasone plates was recommended for the morning fall over a cumulative fatigue from September 20222218. Specifically, with temozolomide chemotherapy with the ribs. Specifically with Velidermidal drain. Specifically with a Demers catheter EC. \\
\bottomrule
\end{tabular}

\vspace{6pt}
\refstepcounter{compsample}\label{sample:comp-lh-fdg}%
\begin{tabular}{@{}p{0.14\linewidth} p{0.82\linewidth}@{}}
\toprule
\textbf{Instruction} & Quote the statement about the liver's FDG uptake and its potential causes. \\
\midrule
\textbf{Ground truth} & The liver, however, presents with increased FDG avidity, especially in the anterolateral segment. This uptake might represent active lymphomatous involvement or could be due to an inflammatory process. \\
\midrule
\textbf{Full KV-cache} & The liver, however, presents with increased FDG avidity, especially in the anterolateral segment. This uptake might represent active lymphomatous involvement or could be due to an inflammatory process. Given the differential, and to ascertain the etiology, further diagnostic evaluation, such as a liver MRI or biopsy, is recommended. \\
\midrule
\textbf{\RegAtt{} (MLP)} & The liver presents with increased FDG avidity, particularly in the anterolateral segment, which may indicate active lymphomatous involvement or an inflammatory process. \\
\midrule
\textbf{Quadrature} & The statement's melanoma lesion in the sense of the sense of melanoma. The mass shows a melanoma is about evaluating a tumor is in the right and the levels of T1. This was decreased mass effect. There is decreased impively in the right aftermarket prompted by a positive response. \\
\bottomrule
\end{tabular}

\vspace{6pt}
\refstepcounter{compsample}\label{sample:comp-lh-biopsy}%
\begin{tabular}{@{}p{0.14\linewidth} p{0.82\linewidth}@{}}
\toprule
\textbf{Instruction} & Paraphrase the histological diagnosis of Mrs.\ Done's skin biopsy and the recommendation for further excision. \\
\midrule
\textbf{Ground truth} & The biopsy analysis identified a dysplastic nevus with severe cellular atypia, which requires additional surgical removal to ensure complete eradication and confirm the absence of invasive melanoma. \\
\midrule
\textbf{Full KV-cache} & Mrs.\ Done's skin biopsy revealed multiple atypical melanocytes arranged in nests and single units at the dermoepidermal junction, with some cells infiltrating the papillary dermis. The histology was positive for HMB-45 and S-100, and focally positive for Melan A. The final diagnosis was dysplastic nevus with severe atypia, and the margins appeared clear. However, due to the suspicious nature of the lesion, the report recommended further excision to ensure complete removal and to rule out invasive melanoma. \\
\midrule
\textbf{\RegAtt{} (MLP)} & Diagnosis: Dysplastic nevus with severe atypia. The margins appear clear, but further excision is recommended to ensure complete removal and to rule out invasive melanoma. \\
\midrule
\textbf{Quadrature} & The histopathological results for Mrs.\ Done's melanoma (MCL) was recommended to have a Durieirates and Durieper-milimumabineinib, which was open PRRTXmlmonitor, which suggests a Durints score to monitor with Fentanylision.5015 Gyants. The melanoma on Days and Gemology. [\dots] \\
\bottomrule
\end{tabular}

\vspace{6pt}
\refstepcounter{compsample}\label{sample:comp-lh-laparotomy}%
\begin{tabular}{@{}p{0.14\linewidth} p{0.82\linewidth}@{}}
\toprule
\textbf{Instruction} & What complication led to the termination of the exploratory laparotomy in January 2022? \\
\midrule
\textbf{Ground truth} & Positive frozen section results from a biopsy near the distal mesenteric root led to the surgery being halted. \\
\midrule
\textbf{Full KV-cache} & Based on the surgery report provided, the exploratory laparotomy in January 2022 was terminated due to positive frozen section results. Specifically, the report states: ``Surgery was stopped due to positive frozen section results; gallbladder removal.''\,[\dots] \\
\midrule
\textbf{\RegAtt{} (MLP)} & The exploratory laparotomy performed in January 2022 was terminated due to positive frozen section results obtained from a biopsy near the distal mesenteric root, indicating the presence of adenocarcinoma infiltrates. \\
\midrule
\textbf{Quadrature} & Based for Mrs.\ Anderson's glioblastoma was halted bile duct narrowing in the right hemisphere in sizezzleopneumothorax. The keyoperative neuromonitoring dome marrow. Specifically, the BKEP4, CK7.4 Gy) was administered using the bile duct blockage of fluid or the \\
\bottomrule
\end{tabular}
\end{table}

\section{Details of Generation Quality Evaluation}
\label{app:eval-pipeline}

Beyond token-level metrics, we evaluate \RegAtt{} via open-ended generation: the model autoregressively generates responses to held-out instructions, and a commercial-grade LLM judge scores the outputs. This appendix describes the pipeline and reproduces the exact prompts used.

\paragraph{Generation.}
For samples in the test split, we generate responses from both the \RegAtt{}-augmented model and the base model attending to the full KV cache. We use nucleus sampling with $p{=}0.8$ and temperature $T{=}0.7$ (unless otherwise noted). Per-task maximum generation lengths are: 64 tokens for QA and 196 tokens for summarization, quoting, and paraphrasing.

\paragraph{Evaluation protocol.}
We compute three similarity scores per sample using GPT-5.2 \citep{singh2025openai} as a judge:
\begin{enumerate}[nosep]
    \item \textbf{\RegAtt{} vs.\ ground-truth}: how well the \RegAtt{} response matches the reference answer.
    \item \textbf{Base vs.\ ground-truth}: how well the full-attention base model response matches the reference answer outputted by the base model \textit{equipped with the full long context KV-cache}.
    \item \textbf{\RegAtt{} vs.\ base}: semantic equivalence between the two model outputs, regardless of correctness.
\end{enumerate}
All scores range from 1 (unrelated or contradictory) to 5 (same meaning). Evaluations (1) and (2) use a task-specific rubric (Listing~\ref{lst:eval-gt}) that assesses how much of the ground-truth's meaning is captured in the model response. Evaluation (3) uses a dedicated comparison rubric (Listing~\ref{lst:eval-compare}) that measures whether a reader would come away with the same understanding from both outputs---shared errors count as agreement. The comparison prompt omits the source context so the judge focuses purely on whether the two responses agree, without re-checking factual accuracy.

\paragraph{Judge configuration.}
We query the judge via the OpenAI-compatible API with default decoding parameters. The judge output is parsed as JSON containing a score and brief reasoning.

\begin{lstlisting}[
    float=h!,
    basicstyle=\footnotesize\ttfamily,
    breaklines=true,
    frame=single,
    caption={Ground-truth evaluation prompt. The \texttt{\{rubric\}} placeholder is populated with the task-specific rubric shown below. The judge assesses semantic similarity between the model response and the ground truth rather than absolute correctness.},
    label={lst:eval-gt},
    aboveskip=6pt,belowskip=6pt]
You are an expert evaluator assessing semantic similarity
between a model's response and a ground truth answer.

<context>
{context_chunk}
</context>

<instruction>
{instruction}
</instruction>

<ground_truth>
{ground_truth}
</ground_truth>

<model_response>
{generated}
</model_response>

{rubric}

Follow these evaluation steps:
1. Understand what the ground truth is saying.
2. Understand what the model response is saying.
3. Assess how much of the ground truth's meaning is captured in
   the response.
4. Assign a score from 1 to 5 based on the criteria above.

Output your evaluation as JSON:
  {"score": 1 to 5, "reasoning": "brief explanation"}
Score guide: 5=same meaning as ground truth, 4=mostly same
meaning with minor differences, 3=partially captures the
meaning, 2=mostly different, 1=unrelated or contradictory
\end{lstlisting}

\paragraph{Task-specific rubrics.}
The \texttt{\{rubric\}} placeholder in Listing~\ref{lst:eval-gt} is populated according to the task type:
\begin{itemize}[nosep]
    \item \textbf{QA}: Does the response convey the same information as the ground truth? Score based on semantic match, not wording. Additional elaboration should not reduce the score.
    \item \textbf{Summarization}: Does the response convey the same overall meaning? Different organization, wording, or level of detail is fine. Score based on whether a reader would come away with the same understanding.
    \item \textbf{Quoting}: Does the response identify and convey the same passage? Score based on whether the same content is captured, not exact wording. Minor word variations or boundary differences are acceptable.
    \item \textbf{Paraphrasing}: Does the response convey the same meaning? Different wording is expected and should not reduce the score. Score based on whether a reader would understand the same thing from both.
\end{itemize}

\begin{lstlisting}[
    float=h!,
    basicstyle=\footnotesize\ttfamily,
    breaklines=true,
    frame=single,
    caption={Pairwise comparison prompt (\RegAtt{} vs.\ base model).},
    label={lst:eval-compare},
    aboveskip=6pt,belowskip=6pt]
You are an expert evaluator assessing semantic similarity
between two model responses.

<instruction>
{instruction}
</instruction>
(The instruction refers to a passage from a book. You do not
need to verify factual accuracy -- only compare the two
responses below.)

<response_a>
{base_generated}
</response_a>

<response_b>
{generated}
</response_b>

Would a reader come away with the same understanding from both
responses?
You are NOT judging correctness -- only whether both responses
convey the same meaning.
If both make the same mistake, that is FULL AGREEMENT.
Different phrasing, level of detail, verbosity, or
organization should NOT reduce the score.
One response being shorter or longer is fine.
Only reduce the score when the two responses would leave a
reader with a meaningfully different understanding of the
answer.
Examples -- score 5: 'played football at New Haven' vs 'Tom
was one of the most powerful ends that ever played football at
New Haven and was a national figure in college'.
Score 3: one says 'football' and the other says 'swimming'.
Score 1: completely contradictory or unrelated answers.

Follow these evaluation steps:
1. Read both responses and understand what each one is saying.
2. Ask: would a reader take away the same understanding from
   both?
3. Ignore differences in length, phrasing, or level of detail.
4. Assign a score from 1 to 5 based on semantic similarity.

Output your evaluation as JSON:
  {"score": 1 to 5, "reasoning": "brief explanation"}
Score guide: 5=same meaning, 4=mostly same understanding with
minor differences, 3=partially similar, 2=mostly different,
1=contradictory or unrelated
\end{lstlisting}

\section{Training Data Generation and Verification Prompts}
\label{app:prompts}

This appendix reproduces the prompt templates used in the data-preparation pipeline (Appendix~\ref{app:data}). \emph{Generation prompts} (Listings~\ref{lst:qa}--\ref{lst:paraphrase}) instruct an LLM to produce instruction--response pairs from a text chunk. \emph{Verification prompts} (Listings~\ref{lst:answer}--\ref{lst:consistency}) are used to re-answer each instruction against the full document and to assess consistency between the chunk-based and full-context responses. In every generation prompt the model first populates a \texttt{<scratchpad>} with key facts before emitting the final JSON pairs.

\begin{lstlisting}[
    float=h!,
    basicstyle=\footnotesize\ttfamily,
    breaklines=true,
    frame=single,
    caption={QA generation prompt.},
    label={lst:qa},
    aboveskip=6pt,belowskip=6pt]
Instruction: First, read the context below and identify the key
facts, events, and concepts it covers. List them as bullet points
in a <scratchpad> section. Then, use those bullet points to
generate question-answer pairs.

GOAL: generate {count} question-answer pairs from the chunk of
text provided.

CRITICAL: Each question must reference the specific content it
asks about (e.g., a person, event, concept, or fact mentioned in
the text) so that someone reading the full document -- not just
this chunk -- could locate the relevant information and answer
correctly. Do NOT use phrases like "in the text", "in the
passage", "according to the chunk", or "mentioned above".
Instead, name the specific subject matter directly. Each pair
must cover a DIFFERENT fact or aspect of the context -- no two
pairs should ask about the same information.

Constraints: First output a <scratchpad> with bullet points of
key facts, then output strictly JSON format:
[{"instruction": "...", "response": "..."}].
Instruction+response should be less than 300 words. Answers must
be contained in the chunk.

Context:
{chunk}
\end{lstlisting}

\begin{lstlisting}[
    float=h!,
    basicstyle=\footnotesize\ttfamily,
    breaklines=true,
    frame=single,
    caption={Summarization generation prompt.},
    label={lst:summarization},
    aboveskip=6pt,belowskip=6pt]
Instruction: First, read the context below and identify the
distinct topics, events, arguments, or descriptions it covers.
List them as bullet points in a <scratchpad> section. Then,
generate summarization pairs that each target a DIFFERENT topic
from your list.

GOAL: generate {count} summarization instruction-response pairs
from the chunk of text provided. Each pair should ask for a
summary of a specific topic, event, argument, or description
found in the chunk, and provide that summary.

CRITICAL: Each instruction must describe WHAT to summarize by
referencing the specific content (e.g., a topic, event,
character, argument, process, or concept discussed in the text)
-- NOT by referring to "the text", "the passage", "the chunk",
or "the above". The instruction must be self-contained so that
someone with access to the full document can locate the relevant
content and produce the same summary. Each pair must cover a
DIFFERENT topic or aspect.

Constraints: First output a <scratchpad> with bullet points of
distinct topics, then output strictly JSON format:
[{"instruction": "...", "response": "..."}].
Instruction+response should be less than 300 words.

Context:
{chunk}
\end{lstlisting}

\begin{lstlisting}[
    float=h!,
    basicstyle=\footnotesize\ttfamily,
    breaklines=true,
    frame=single,
    caption={Quoting generation prompt.},
    label={lst:quoting},
    aboveskip=6pt,belowskip=6pt]
Instruction: First, read the context below and identify notable
sentences, definitions, claims, or descriptions worth quoting.
List them as bullet points in a <scratchpad> section. Then,
generate quoting pairs that each target a DIFFERENT passage.

GOAL: generate {count} quoting instruction-response pairs from
the chunk of text provided. Each pair should ask to quote or
extract a specific passage, sentence, or phrase from the text,
and the response should provide the exact quote.

CRITICAL: Each instruction must identify WHAT to quote by
describing the specific content being sought (e.g., a statement
by a particular person, a definition, a key claim) -- NOT by
referring to "the text above", "the passage", or using
positional references like "the first sentence". The instruction
must be self-contained so that someone with access to the full
document can find and extract the same quote.

Constraints: First output a <scratchpad> with bullet points of
notable quotable passages, then output strictly JSON format:
[{"instruction": "...", "response": "..."}].
Instruction+response should be less than 300 words. Responses
must be exact quotes from the chunk text.

Context:
{chunk}
\end{lstlisting}

\begin{lstlisting}[
    float=h!,
    basicstyle=\footnotesize\ttfamily,
    breaklines=true,
    frame=single,
    caption={Paraphrasing generation prompt.},
    label={lst:paraphrase},
    aboveskip=6pt,belowskip=6pt]
Instruction: First, read the context below and identify the key
ideas, arguments, or descriptions that could be meaningfully
paraphrased. List them as bullet points in a <scratchpad>
section. Then, generate paraphrasing pairs that each target a
DIFFERENT idea.

GOAL: generate {count} paraphrasing instruction-response pairs
from the chunk of text provided. Each pair should ask to
paraphrase or rephrase a specific idea, argument, or description
from the text, and the response should provide a faithful
paraphrase.

CRITICAL: Each instruction must describe WHAT to paraphrase by
referencing the specific content -- NOT by referring to "the
text", "the passage", "the chunk", or using positional
references. The instruction must be self-contained so that
someone with access to the full document can locate the relevant
content and produce an equivalent paraphrase.

Constraints: First output a <scratchpad> with bullet points of
key ideas, then output strictly JSON format:
[{"instruction": "...", "response": "..."}].
Instruction+response should be less than 300 words. Responses
must be faithful paraphrases preserving the original meaning.

Context:
{chunk}
\end{lstlisting}

\begin{lstlisting}[
    float=h!,
    basicstyle=\footnotesize\ttfamily,
    breaklines=true,
    frame=single,
    caption={Answer generation prompt (used during verification with
             either the originating chunk or the full document as
             context).},
    label={lst:answer},
    aboveskip=6pt,belowskip=6pt]
Instruction: You are a precise question-answering assistant.
Answer the instruction using ONLY the information provided in
the {context_type} below. If the answer cannot be determined
from the context, respond with "INSUFFICIENT_CONTEXT".

GOAL: Provide a concise, accurate response (max 300 words)
based solely on the given context.

Constraints:
- Use ONLY information from the provided context
- Keep response under 300 words
- Be specific and factual
- If information is insufficient, respond with
  "INSUFFICIENT_CONTEXT"

{Context_Label}:
{context}

Instruction:
{instruction}

Response:
\end{lstlisting}

\begin{lstlisting}[
    float=h!,
    basicstyle=\footnotesize\ttfamily,
    breaklines=true,
    frame=single,
    caption={Consistency check prompt (used during verification).
             The task\_guidance block is swapped per task type;
             see below.},
    label={lst:consistency},
    aboveskip=6pt,belowskip=6pt]
Instruction: You are a semantic consistency validator. Evaluate
whether two responses to the same instruction are semantically
consistent (i.e., they provide the same information, even if
phrased differently).

GOAL: Determine if the responses are consistent and provide a
confidence score and reasoning.

Output Format (JSON):
{
    "is_consistent": true/false,
    "confidence": 0.0-1.0,
    "reasoning": "Brief explanation"
}

Consistency Criteria:
- CONSISTENT: Responses convey the same core information
  (different phrasing is OK)
- INCONSISTENT: Responses contradict each other or provide
  fundamentally different information
- INSUFFICIENT: One or both responses are
  "INSUFFICIENT_CONTEXT"

{task_guidance}

Instruction:
{instruction}

Response from Chunk:
{response_chunk}

Response from Full Context:
{response_full}

Evaluation (JSON):
\end{lstlisting}

\paragraph{Task-specific consistency criteria.}
The \texttt{\{task\_guidance\}} placeholder in Listing~\ref{lst:consistency} is populated according to the task type:
\begin{itemize}
    \item \textbf{QA}: Responses must contain the same core factual information; different phrasing or level of detail is acceptable.
    \item \textbf{Summarization}: Both summaries should cover the same key points; different organizational choices are acceptable.
    \item \textbf{Quoting}: Quoted text must be identical or near-identical; minor differences in quote boundaries are acceptable.
    \item \textbf{Paraphrasing}: Both paraphrases must preserve the same meaning; different stylistic choices are acceptable as long as factual content matches.
\end{itemize}


\applefootnote{\textcolor{textgray}{\sffamily Apple and the Apple logo are trademarks of Apple Inc., registered in the U.S. and other countries and regions.}}

\end{document}